\PassOptionsToPackage{table}{xcolor}
\documentclass[10pt,twocolumn,letterpaper]{article}

\usepackage[pagenumbers]{iccv} 

\definecolor{iccvblue}{rgb}{0.21,0.49,0.74}
\usepackage[pagebackref,breaklinks,colorlinks]{hyperref}

\usepackage{multirow}
\usepackage{subcaption}
\usepackage{caption}
\usepackage{array}
\usepackage{stfloats}
\usepackage{makecell}
\def\paperID{0000} 
\def\confName{ICCV}
\def\confYear{2025}

\title{1000+ FPS 4D Gaussian Splatting for Dynamic Scene Rendering}

\author{
Yuan Yuheng \quad
Qiuhong Shen \quad
Xingyi Yang \quad
Xinchao Wang\\
National University of Singapore \\
\tt\small{\{yuhengyuan,qiuhong.shen,xyang\}@u.nus.edu, xinchao@nus.edu.sg}
}

\begin{document}

\twocolumn[{%
\renewcommand\twocolumn[1][]{#1}%

\maketitle

\begin{minipage}[b]{.7\linewidth}
    \centering
    \hspace{-1cm}
    \includegraphics[scale=.11]{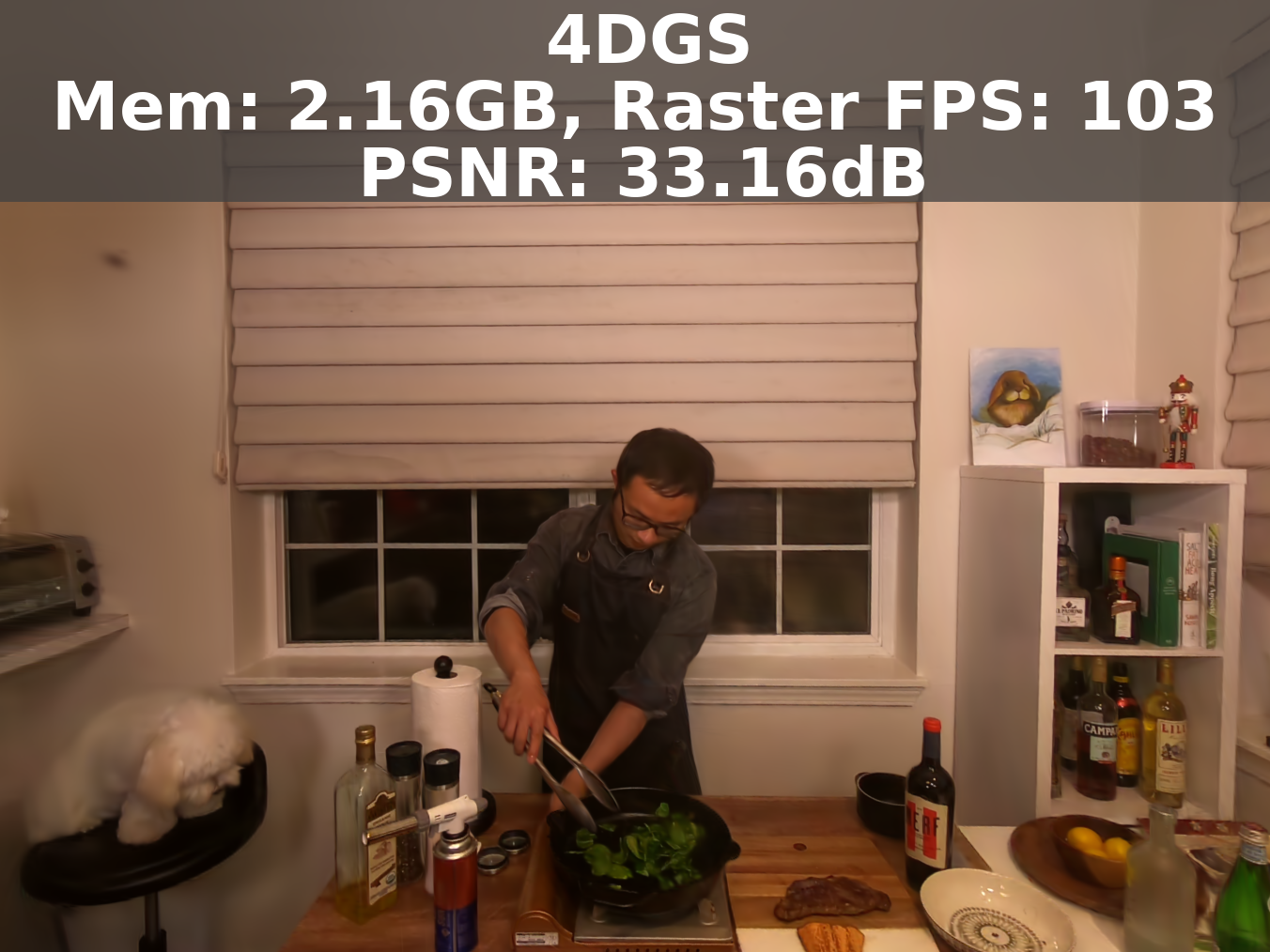}
    \hspace{0.2cm}
    \includegraphics[scale=.11]{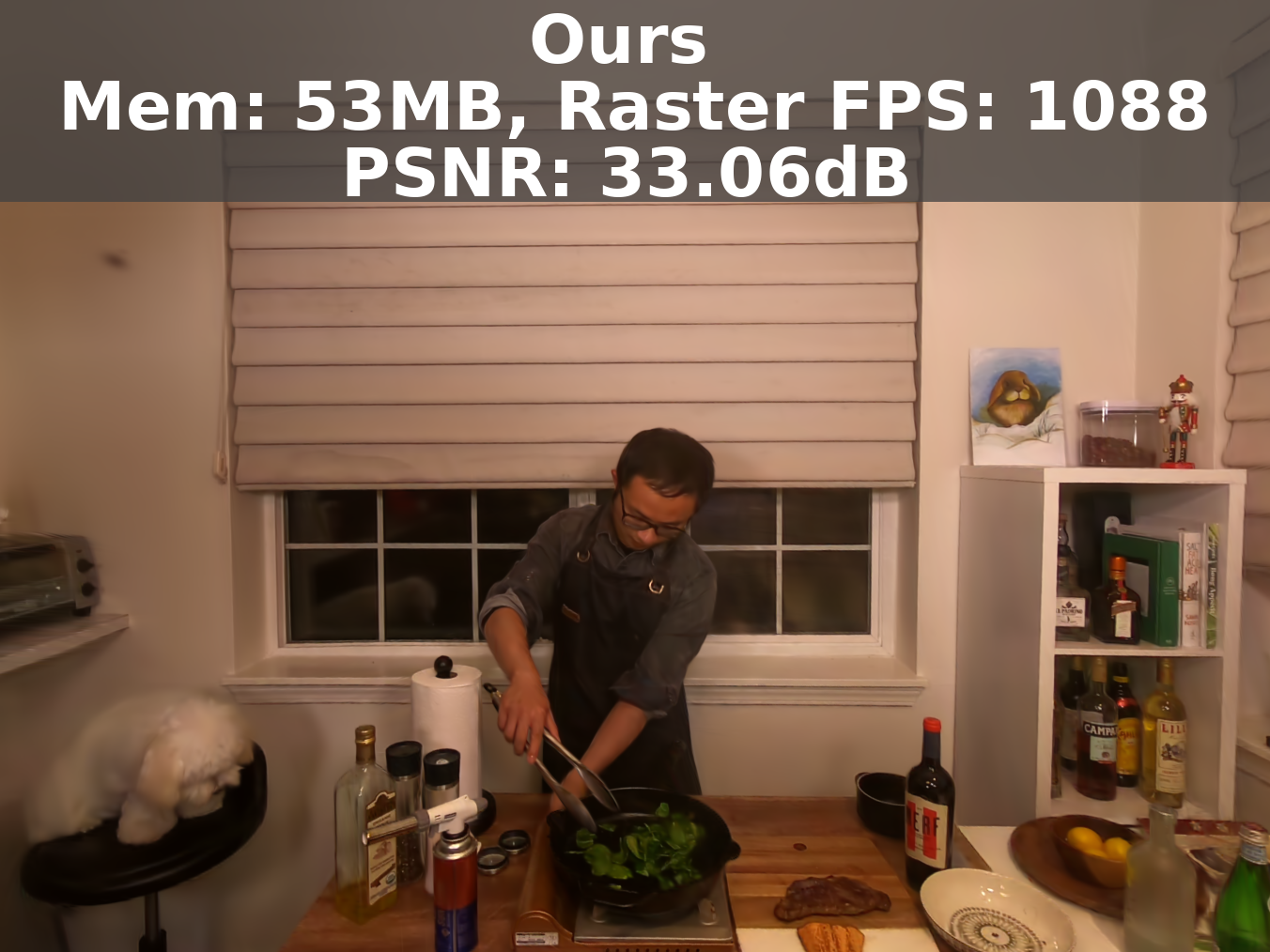}
\end{minipage}
\begin{minipage}[b]{.27\linewidth}
    \hspace{-1.cm}
    \vspace{-.34cm}
    \includegraphics[scale=.36]{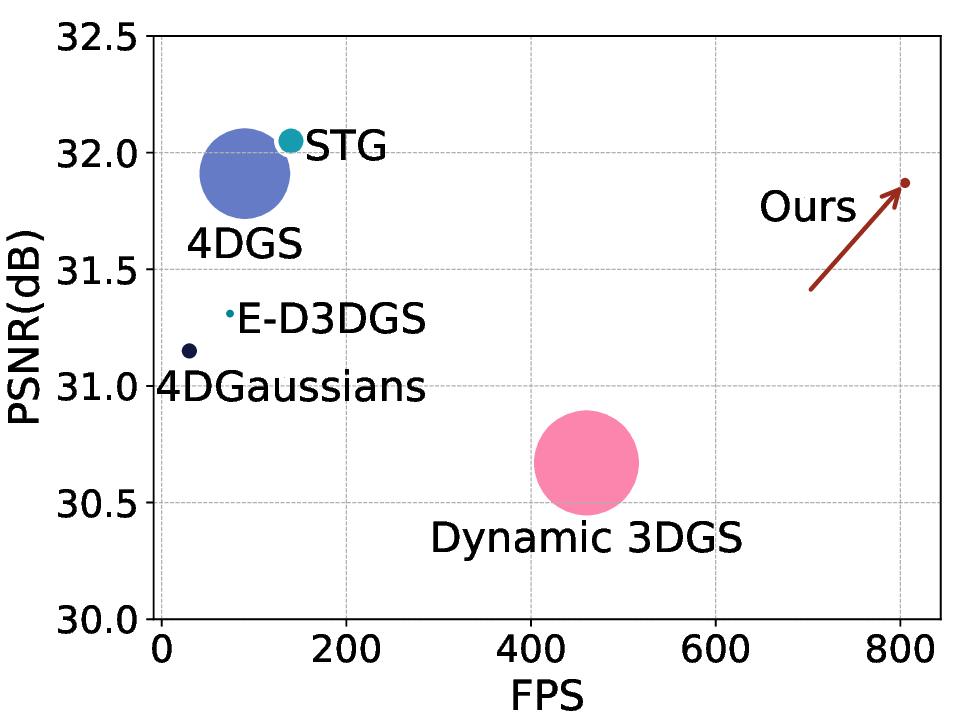}
\end{minipage}

\captionof{figure}{\textbf{Compressibility and Rendering Speed.} We introduce \textbf{4DGS-1K}, a novel compact representation with high rendering speed. In contrast to 4D Gaussian Splatting (4DGS)~\cite{yang2023real}, we can achieve rasterization at 1000+~FPS while maintaining comparable photorealistic quality with only $2\%$ of the original storage size. The right figure is the result tested on the N3V~\cite{li2022neural} datasets, where the radius of the dot corresponds to the storage size.}
\label{fig:teaser}
}]

\begin{abstract}
4D Gaussian Splatting (4DGS) has recently gained considerable attention as a method for reconstructing dynamic scenes. Despite achieving superior quality, 4DGS typically requires substantial storage and suffers from slow rendering speed. In this work, we delve into these issues and identify two key sources of temporal redundancy. 
(Q1) \textbf{Short-Lifespan Gaussians}: 4DGS uses a large portion of Gaussians with short temporal span to represent scene dynamics, leading to an excessive number of Gaussians. 
(Q2) \textbf{Inactive Gaussians}: When rendering, only a small subset of Gaussians contributes to each frame. Despite this, all Gaussians are processed during rasterization, resulting in redundant computation overhead.
To address these redundancies, we present \textbf{4DGS-1K}, which runs at over 1000 FPS on modern GPUs. 
For Q1, we introduce the Spatial-Temporal Variation Score, a new pruning criterion that effectively removes short-lifespan Gaussians while encouraging 4DGS to capture scene dynamics using Gaussians with longer temporal spans.
For Q2, we store a mask for active Gaussians across consecutive frames, significantly reducing redundant computations in rendering. 
Compared to vanilla 4DGS, our method achieves a $41\times$ reduction in storage and $9\times$ faster rasterization speed on complex dynamic scenes, while maintaining comparable visual quality.
Please see our project page at \href{https://4DGS-1K.github.io}{4DGS-1K}.
\end{abstract}
\section{Introduction}
Novel view synthesis for dynamic scenes allows for the creation of realistic representations of 4D environments, which is essential in fields like computer vision, virtual reality, and augmented reality. Traditionally, this area has been led by neural radiance fields (NeRF)~\cite{mildenhall2021nerf,lombardi2019neural, li2022neural, attal2023hyperreel, fridovich2023k}, which model opacity and color over time to depict dynamic scenes. While effective, these NeRF-based methods come with high training and rendering costs, limiting their practicality, especially in real-time applications and on devices with limited resources.

Recently, point-based representations like 4D Gaussian Splatting (4DGS)~\cite{yang2023real} have emerged as strong alternatives. 4DGS models a dynamic scene using a set of 4D Gaussian primitives, each with a 4-dimensional mean and a $4\times4$ covariance matrix. At any given timestamp, a 4D Gaussian is decomposed into a set of conditional 3D Gaussians and a marginal 1D Gaussian, the latter controlling the opacity at that moment. This mechanism allows 4DGS to effectively capture both static and dynamic features of a scene, enabling high-fidelity dynamic scene reconstruction.

However, representing dynamic scenes with 4DGS is both storage-intensive and slow. Specifically, 4DGS often requires millions of Gaussians, leading to significant storage demands (averaging 2GB for each scene on the N3V~\cite{li2022neural} dataset) and suboptimal rendering speed. In comparison, mainstream deformation field methods~\cite{wu20244d} require only about 90MB for the same dataset. Therefore, reducing the storage size of 4DGS~\cite{yang2023real} and improving rendering speed are essential for efficiently representing complex dynamic scenes.

We look into the cause of such an explosive number of Gaussian and place a specific emphasis on two key issues.
\textbf{(Q1)} A large portion of Gaussians exhibit a short temporal span. In empirical experiments, 4DGS tends to favor ``flicking'' Gaussians to fit complex dynamic scenes, which just influence a short portion of the temporal domain. This necessitates that 4DGS relies on a large number of Gaussians to reconstruct a high-fidelity scene. As a result, substantial storage is needed to record the attributes of these Gaussians:
\textbf{(Q2)} Inactive Gaussians lead to redundant computation. During rendering, 4DGS needs to process all Gaussians. However, only a very small portion of Gaussians are active at that moment. Therefore, most of the computation time is spent on inactive Gaussians. This phenomenon greatly hampers the rendering speed.
In this paper, we introduce \textbf{4DGS-1K}, a framework that significantly reduces the number of Gaussians to minimize storage requirements and speedup rendering while maintaining high-quality reconstruction.
To address these issues, 4DGS-1K introduces a two-step pruning approach:

\begin{itemize}
    \item \textbf{Pruning Short-Lifespan Gaussians.} We propose a novel pruning criterion called the \emph{spatial-temporal variation score}, which evaluates the temporal impact of each Gaussian. Gaussians with minimal influence are identified and pruned, resulting in a more compact scene representation with fewer Gaussians with short temporal span.
    \item \textbf{Filtering Inactive Gaussians.} To further reduce redundant computations during rendering, we use a key-frame temporal filter that selects the Gaussians needed for each frame. On top of this, we share the masks for adjacent frames. This is based on our observation that Gaussians active in adjacent frames often overlap significantly. 
\end{itemize}
Besides, the pruning in step 1 enhances the masking process in step 2. 
By pruning Gaussians, we increase the temporal influence of each Gaussian, which allows us to select sparser key frames and further reduce storage requirements. 

We have extensively tested our proposed model on various dynamic scene datasets including real and synthetic scenes. As shown in~\cref{fig:teaser}, 4DGS-1K reduces storage costs by 41$\times$ on the Neural 3D Video datasets~\cite{li2022neural} while maintaining equivalent scene representation quality. Crucially, it enables real-time rasterization speeds exceeding 1,000 FPS. These advancements collectively position 4DGS-1K as a practical solution for high-fidelity dynamic scene modeling without compromising efficiency.

In summary, our contributions are three-fold:
\begin{itemize}
    \item We delve into the temporal redundancy of 4D Gaussian Splatting, and explain the main reason for the storage pressure and suboptimal rendering speed.
    \item We introduce \textbf{4DGS-1K}, a compact and memory-efficient framework to address these issues. It consists of two key components, a spatial-temporal variation score-based pruning strategy and a temporal filter.
    \item Extensive experiments demonstrate that 4DGS-1K not only achieves a substantial storage reduction of approximately 41$\times$ but also accelerates rasterization to 1000+ FPS while maintaining high-quality reconstruction.
\end{itemize}

\section{Related Work}

\subsection{Novel view synthesis for static scenes}

Recently, neural radiance fields(NeRF)~\cite{mildenhall2021nerf} have achieved encouraging results in novel view synthesis. NeRF~\cite{mildenhall2021nerf} represents the scene by mapping 3D coordinates and view dependency to color and opacity. Since NeRF~\cite{mildenhall2021nerf} requires sampling each ray by querying the MLP for hundreds of points, this significantly limits the training and rendering speed. 
Subsequent studies~\cite{chen2022tensorf, schwarz2022voxgraf, sun2022direct, wang2022fourier, fridovich2022plenoxels, muller2022instant, reiser2021kilonerf, wang2022r2l} have attempted to speed up the rendering by introducing specialized designs. However, these designs also constrain the widespread application of these models.
In contrast, 3D Gaussian Splatting(3DGS)~\cite{kerbl20233d} has gained significant attention, which utilizes anisotropic 3D Gaussians to represent scenes. 
It achieves high-quality results with intricate details, while maintaining real-time rendering performance. 

\subsection{Novel view synthesis for dynamic scenes}
Dynamic NVS poses new challenges due to the temporal variations in the input images. 
Previous NeRF-based dynamic scene representation methods~\cite{lombardi2019neural, mildenhall2019local, li2022neural, li2022streaming, attal2023hyperreel, fridovich2023k, pumarola2021d, cao2023hexplane, song2023nerfplayer, wang2023mixed} handle dynamic scenes by learning a mapping from spatiotemporal coordinates to color and density. 
Unfortunately, these NeRF-based models are constrained in their applications due to low rendering speeds.
Recently, 3D Gaussians Splatting~\cite{kerbl20233d} has emerged as a novel explicit representation, with many studies~\cite{yang2024deformable, wu20244d,  guo2024motion, lu20243d, bae2024per, das2024neural} attempting to model the dynamic scenes based on it. 
4D Gaussian Splatting(4DGS)~\cite{yang2023real} is one of the representatives. It utilizes a set of 4D Gaussian primitives.
However, 4DGS often requires a huge redundant number of Gaussians for dynamic scenes. These Gaussians lead to tremendous storage and suboptimal rendering speed. 
To this end, we focus on analyzing the temporal redundancy of 4DGS~\cite{yang2023real} in hopes of developing a novel framework to achieve lower storage requirements and higher rendering speeds.

\subsection{Gaussian Splatting Compression}
\label{subsection:comperssion}
3D Gaussian-based large-scale scene reconstruction typically requires millions of Gaussians, resulting in the requirement of up to several gigabytes of storage. Therefore, subsequent studies have attempted to tackle these issues. Specifically, 
Compgs~\cite{navaneet2024compgs} and Compact3D~\cite{lee2024compact} employ vector quantization to store Gaussians within codebooks. Concurrently, inspired by model pruning, some studies ~\cite{fan2023lightgaussian, fang2024mini, niemeyer2024radsplat, ali2024trimming, papantonakis2024reducing, liu2024efficientgs} have proposed criterion to prune Gaussians by a specified ratio. 
However, compared to 3DGS~\cite{kerbl20233d}, 4DGS~\cite{yang2023real} introduces an extra temporal dimension to enable dynamic representation. Previous 3DGS-based methods may therefore be unsuitable for 4DGS. Consequently, we first identify a key limitation leading to this problem, referred as \emph{temporal redundancy}. Furthermore, we propose a novel pruning criterion leveraging spatial-temporal variation, and a temporal filter to achieve more efficient storage requirements and higher rendering speed.    
\section{Preliminary of 4D Gaussian Splatting}
\label{subsection:Preliminary}
Our framework builds on 4D Gaussian Splatting (4DGS)~\cite{yang2023real}, which reconstructs dynamic scenes by optimizing a collection of \emph{anisotropic 4D Gaussian primitives}.
For each Gaussian, it is characterized by a 4D mean $\mu=(\mu_x,\mu_y,\mu_z,\mu_t) \in \mathbb{R}^4$ coupled with a covariance matrix $\Sigma \in \mathbb{R}^{4\times4}$.

By treating time and space dimensions equally, the 4D covariance matrix $\Sigma$ can be decomposed into a scaling matrix $S_{4D}=(s_x,s_y,s_z,s_t)\in\mathbb{R}^4$ and a rotation matrix $R_{4D} \in \mathbb{R}^{4\times4}$. $R_{4D}$ is represented by a pair of left quaternion $q_{l} \in \mathbb{R}^4$ and right quaternion $q_{r} \in \mathbb{R}^4$.

During rendering, each 4D Gaussian is decomposed into a conditional 3D Gaussian and a 1D Gaussian at a specific time $t$. Moreover, the conditional 3D Gaussian can be derived from the properties of the multivariate Gaussian with:
\begin{equation}
	\begin{split}
	\mu_{xyz|t} &= \mu_{1:3} + \Sigma_{1:3,4}\Sigma^{-1}_{4,4}(t-\mu_t)\\ 
	\Sigma_{xyz|t}&= \Sigma_{1:3,1:3} - \Sigma_{1:3,4}\Sigma^{-1}_{4,4}\Sigma_{4,1:3}
	\end{split}
\label{eq:4dgs}
\end{equation}
Here, $\mu_{1:3} \in \mathbb{R}^{3}$ and $\Sigma_{1:3,1:3} \in \mathbb{R}^{3\times3}$ denote the spatial mean and covariance, while $\mu_{t}$ and $\Sigma_{4,4}$ are scalars representing the temporal components.
To perform rasterization, given a pixel under view $\mathcal{I}$ and timestamp $t$, its color $\mathcal{I}(u,v,t)$ can be computed by blending visible Gaussians that are sorted by their depth:
\begin{equation}
    \mathcal{I}(u,v,t)=\sum_i^Nc_i(d)\alpha_i\prod_{j=1}^{i-1}(1-\alpha_j )
\label{eq:blending}
\end{equation}
with
\begin{equation}
\begin{split}
    \alpha_i &= p_i(t)p_t(u,v|t)\sigma_i \\
    p_i(t) &\sim \mathcal{N}(t;\mu_t,\Sigma_{4,4})
\end{split} 
\label{eq:temalpha}
\end{equation}
where $c_i(d)$ is the color of each Gaussian, and $\alpha_i$ is given by evaluating a 2D Gaussian with covariance $\Sigma_{2D}$ multiplied with a learned per-point opacity $\sigma_i$ and temporal Gaussian distribution $p_i(t)$. In the following discussion, we denote $\Sigma_{4,4}$ as $\Sigma_t$ for simplicity.


\noindent\textbf{Temporal Redundancy.} 
Despite achieving high quality, 
4DGS requires a huge number of Gaussians to model dynamic scenes. We identify a key limitation leading to this problem: 4DGS represents scenes through temporally independent Gaussians that lack explicit correlation across time. This means that, even static objects are redundantly represented by hundreds of Gaussians, which inconsistently appear or vanish across timesteps. We refer to this phenomenon as \emph{temporal redundancy}. As a result, scenes end up needing more Gaussians than they should, leading to excessive storage demands and suboptimal rendering speeds. In \cref{sec:method}, we analyze the root causes of this issue and propose a set of solutions to reduce the count of Gaussians.



\section{Methodology}
\label{sec:method}

Our goal is to compress 4DGS by reducing the number of Gaussians while preserving rendering quality. To achieve this, we first analyze the redundancies present in 4DGS, as detailed in~\cref{subsection:Delving}. Building on this analysis, we introduce 4DGS-1K in~\cref{subsection:4dgs1k}, which incorporates a set of compression techniques designed for 4DGS. 4DGS-1K enables rendering speeds of over 1,000 FPS on modern GPUs.



\begin{figure*}[tp] \centering
    \begin{subfigure}{0.3\linewidth}
    \includegraphics[width=.95\linewidth]{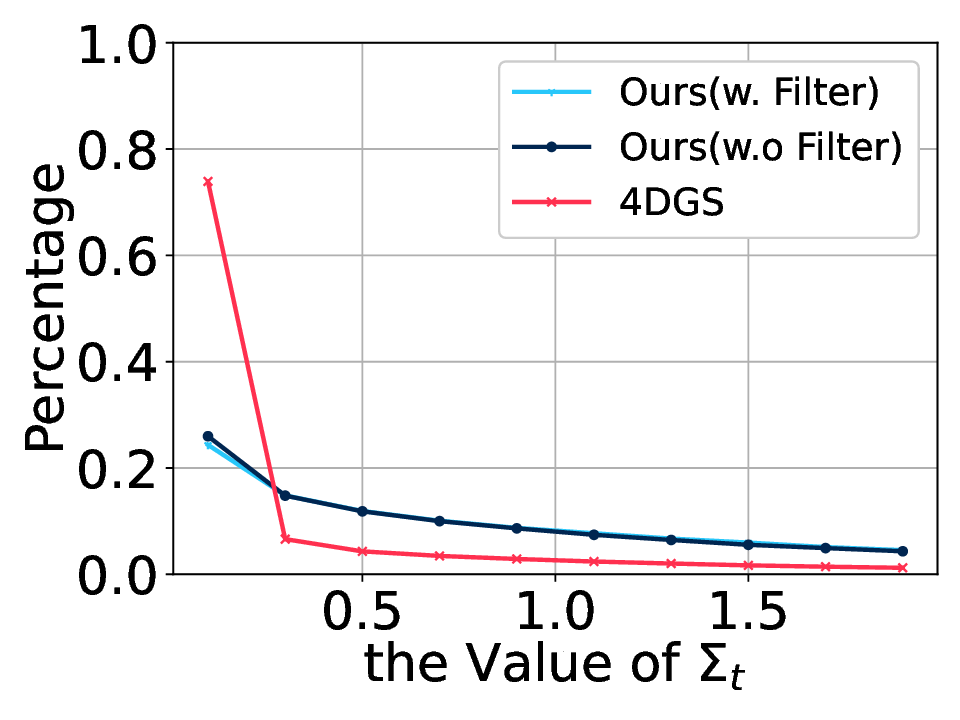}
    \caption{}
    \label{fig:distribution}
  \end{subfigure}
    \begin{subfigure}{0.3\linewidth}
    \includegraphics[width=.95\linewidth]{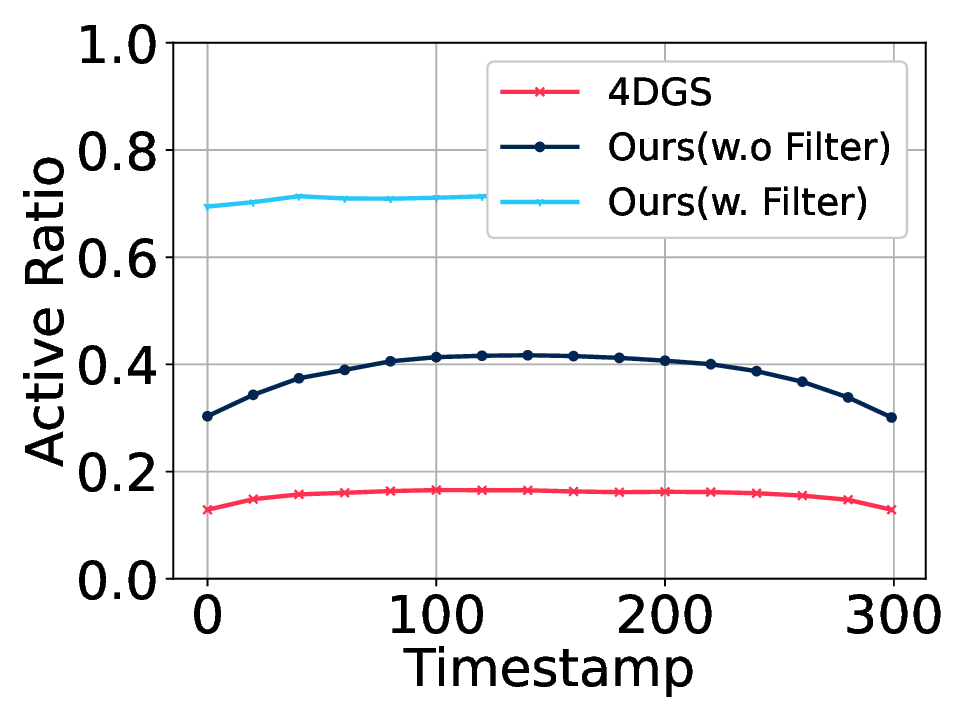}
    \caption{}
    \label{fig:ratio}
  \end{subfigure}
  \begin{subfigure}{0.3\linewidth}
    \includegraphics[width=.95\linewidth]{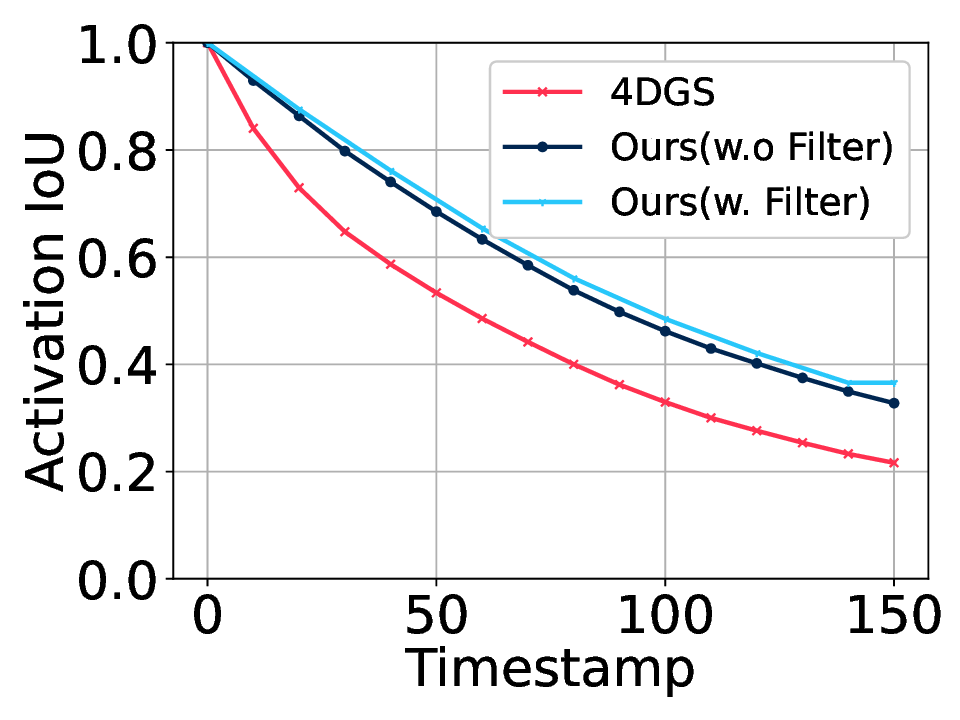}
    \caption{}
    \label{fig:iou}
  \end{subfigure}
\caption{\textbf{Temporal redundancy Study.} (a) The $\Sigma_t$ distribution of 4DGS. The red line shows the result of vanilla 4DGS. The other two lines represent our model has effectively reduced the number of transient Gaussians with small $\Sigma_t$. (b) The active ratio during rendering at different timestamps. It demonstrates that most of the computation time is spent on inactive Gaussians in vanilla 4DGS. However, 4DGS-1K can significantly reduce the occurrence of inactive Gaussians during rendering to avoid unnecessary computations. (c) This figure shows the IoU between the set of active Gaussians in the first frame and frame t. It proves that active Gaussians tend to overlap significantly across adjacent frames.
} \label{fig:redundancy}
\end{figure*}

\begin{figure}[thb] \centering
    \includegraphics[width=0.23\textwidth]{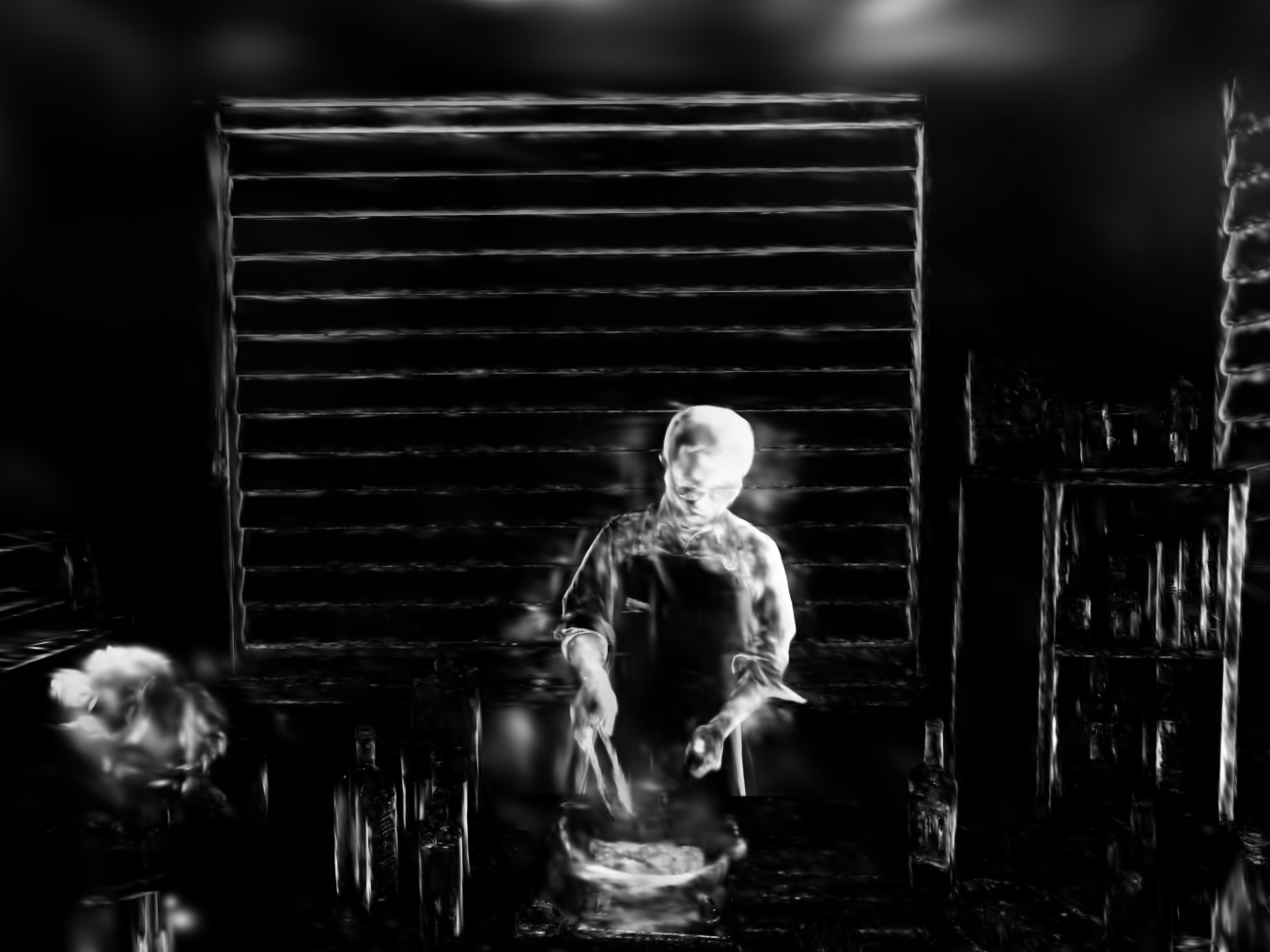}
    \includegraphics[width=0.23\textwidth]{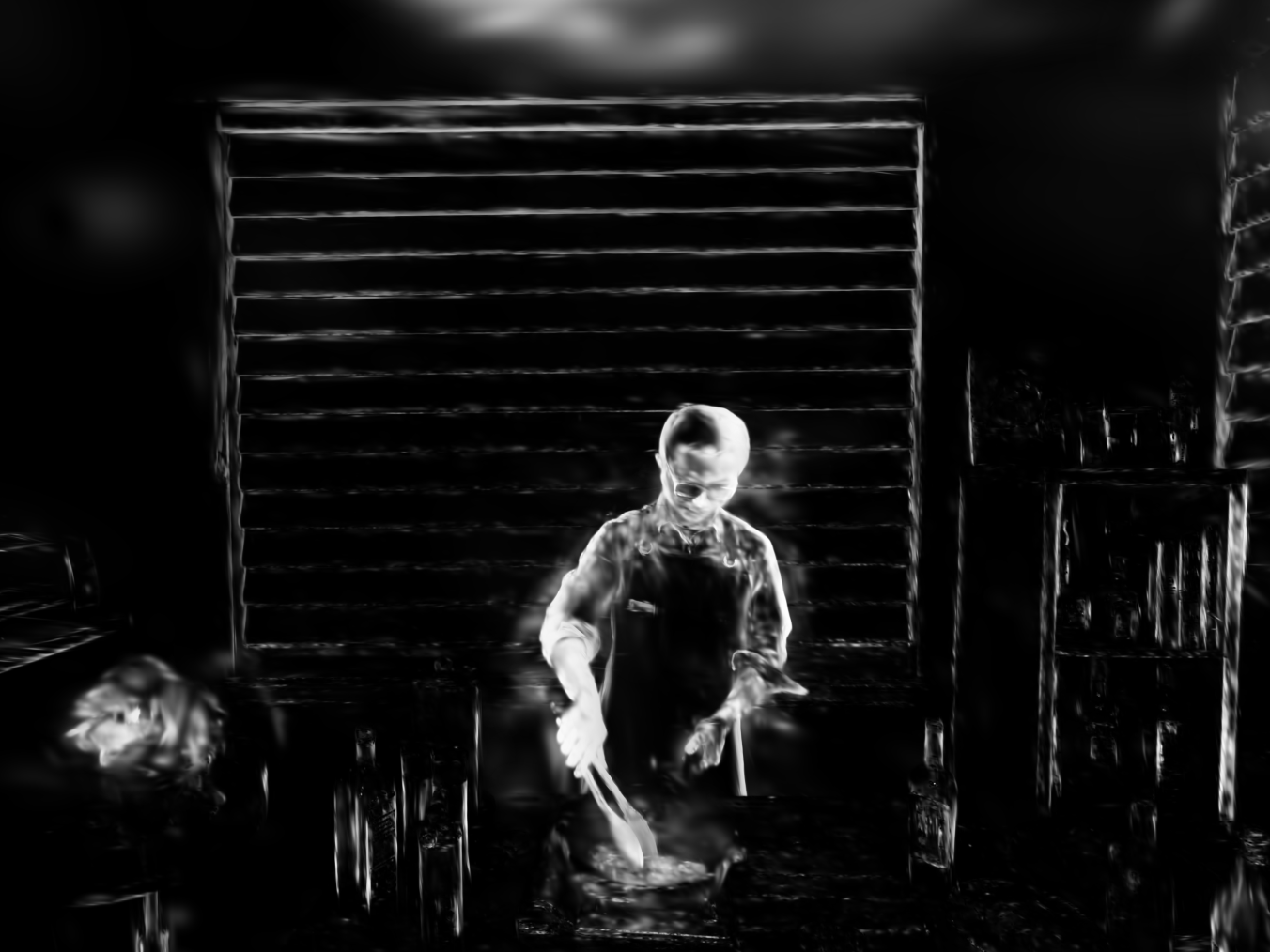} 
    \caption{\textbf{Visualizations of Distribution of $\Sigma_t$.} Most of these Gaussians are concentrated along the edges of moving objects.} \label{fig:temporal}
\end{figure}
\subsection{Understanding Redundancy in 4DGS}
\label{subsection:Delving}

This section investigates why 4DGS requires an excessive number of Gaussians to represent dynamic scenes. In particular, we identify two key factors. First, 4DGS models object motion using a large number of transient Gaussians that inconsistently appear and disappear across timesteps, leading to redundant temporal representations. Second, for each frame, only a small fraction of Gaussians actually contribute to the rendering. We discuss those problems below.

\noindent\textbf{Massive Short-Lifespan Gaussians}. We observe that 4DGS tends to store numerous Gaussians that flicker in time. We refer to these as \emph{Short-Lifespan Gaussians}. To investigate this property, we analyze the Gaussians' opacity, which controls visibility. Intuitively, Short-Lifespan Gaussians exhibit an opacity pattern that rapidly increases and then suddenly decreases. In 4DGS, this behavior is typically reflected in the time variance parameter $\Sigma_{t}$—small $\Sigma_{t}$ values indicate a short lifespan.

\noindent\underline{Observations.} Specifically, we plot the distribution of $\Sigma_t$ for all Gaussians in the \textit{Sear Steak} scene. As shown in ~\cref{fig:distribution}, most of Gaussians has small $\Sigma_t$ values~(e.g. 70\% have $\Sigma_t< 0.25$). Moreover, as shown in~\cref{fig:temporal}, we visualize the spatial distribution of $\Sigma_t$ values. We take the reciprocal of $\Sigma_t$ and then normalize it. Therefore, brighter regions in the image indicate smaller $\Sigma_t$. Most of these Gaussians are concentrated along the edges of moving objects.

Therefore, in 4DGS, nearly all Gaussians have a short lifespan, especially around the fast-moving objects. 
This property leads to high storage needs and slower rendering.




\noindent\textbf{Inactive Gaussians.} Another finding is that, during the forward rendering, actually, only a small fraction of Gaussians are contributing. Interestingly, active ones tend to overlap significantly across adjacent frames. To quantify this, we introduce two metrics: (1) \emph{Active ratio}. This ratio is defined as the proportion of the total number of active Gaussians across all views at any moment relative to the total number of Gaussians. (2) \emph{Activation Intersection-over-Union (IoU)}. This is computed as IoU between the set of active Gaussians in the first frame and in frame $t$.

\noindent\underline{Observations.} Again, we plot the two metrics from \textit{Sear Steak} scene. As shown in ~\cref{fig:ratio}, nearly $85\%$ of Gaussians are inactive at each frame, even though all Gaussians are processed during rendering. Moreover, \cref{fig:iou} demonstrates that the active Gaussians remain quite consistent over time, with an IoU above 80\% over a 20-frame window.

The inactive gaussians bring a significant issue in 4DGS, because each 4D Gaussian must be decomposed into a 3D Gaussian and a 1D Gaussian before rasterization (see \cref{eq:4dgs}). Therefore, a large portion of computational resources is wasted on inactive Gaussians.

 In summary, redundancy in 4DGS comes from massive Short-Lifespan Gaussians and inactive Gaussians. These insights motivate our compression strategy to eliminate redundant computations while preserving rendering quality.

\subsection{4DGS-1K for Fast Dynamic Scene Rendering}
\label{subsection:4dgs1k}
\begin{figure*}[tp] \centering
    \includegraphics[width=.9\linewidth]{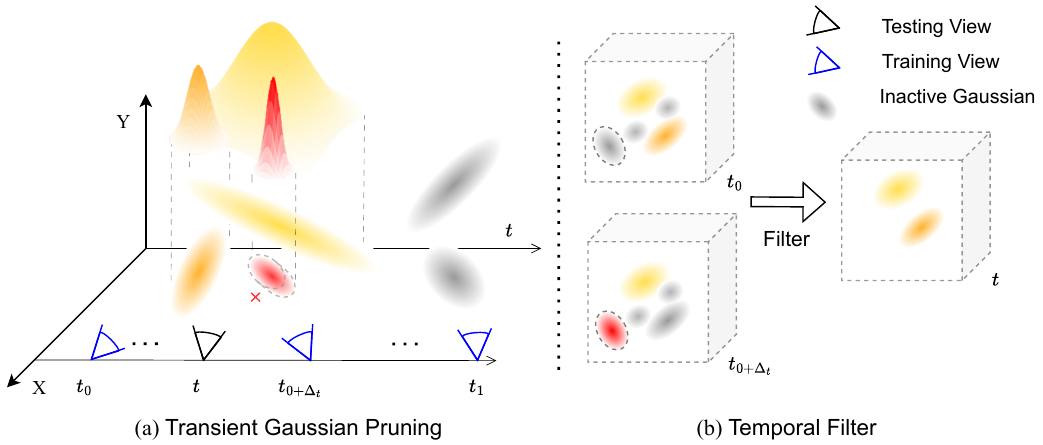}
    \caption{
    \textbf{Overview of 4DGS-1K.} (a) We first calculate the spatial-temporal variation score for each 4D Gaussian on training views, to prune Gaussians with short lifespan (The Red Gaussian). (b) The temporal filter is introduced to filter out inactive Gaussians before the rendering process to alleviate suboptimal rendering speed. At a given timestamp $t$, the set of Gaussians participating in rendering is derived from the two adjacent key-frames, $t_0$ and $t_{0+\Delta_t}$.
    } 
    \label{fig:pipeline}
\end{figure*}
Building on the analysis above, we introduce 4DGS-1K, a suite of compression techniques specifically designed for 4DGS to eliminate redundant Gaussians. As shown in \cref{fig:pipeline}, this process involves two key steps. First, we identify and globally prune unimportant Gaussians with low Spatial-Temporal Variation Score in~\cref{subsection:pruning}. Second, we apply local pruning using a temporal filter to inactive Gaussians that are not needed at each timestep in~\cref{subsection:filter}.

\subsubsection{Pruning with Spatial-Temporal Variation Score}
\label{subsection:pruning}


We first prune unimportant 4D Gaussians to improve efficiency.
Like 3DGS, we remove those that have a low impact on rendered pixels. Besides, we additionally remove short-lifespan Gaussians—those that persist only briefly over time. To achieve this, we introduce a novel spatial-temporal variation score as the pruning criterion for 4DGS. It is composed of two parts, \emph{spatial score} that measures the Gaussians contributions to the pixels in rendering, and \emph{temporal score} considering the lifespan of Gaussians.


\noindent\textbf{Spatial score.} Inspired by the previous method~\cite{fan2023lightgaussian, fang2024mini} and $\alpha$-blending in 3DGS~\cite{kerbl20233d}, we define the spatial score by aggregating the ray contribution of Gaussian $g_i$ along all rays $r$ across all input images at a given timestamp. It can accurately capture the contribution of each Gaussian to one pixel. Consequently, the spatial contribution score $\mathcal{S}^{S}$ is obtained by traversing all pixels:
\begin{equation}
    \mathcal{S}^S_{i} = \sum^{NHW}_{k=1}\alpha_i\prod_{j=1}^{i-1}(1-\alpha_j )
\end{equation}
where $\alpha_i\prod_{j=1}^{i-1}(1-\alpha_j )$ reflects the contribution of $i^{th}$ Gaussian to the final color of all pixels according to the alpha composition in \cref{eq:blending}.


\noindent\textbf{Temporal score.} It is expected to assign a higher temporal score to Gaussians with a longer lifespan. To quantify this, we compute the \emph{second derivative of} temporal opacity function $p_i(t)$ defined in \cref{eq:temalpha}. The second derivative $p^{(2)}_{i}(t)$ is computed as 
\begin{equation}
    p^{(2)}_{i}(t) = (\frac{(t-\mu_t)^2}{\Sigma_t^2} -\frac{1}{\Sigma_t})p_i(t)
\end{equation} 
Intuitively, large second derivative magnitude corresponds to unstable, short-lived Gaussians, while low second derivative indicates smooth, persistent ones.

Moreover, since the second derivative spans the real number domain $\mathbb{R}$, we apply $\tanh$ function to map it to the interval~$\left(0,1\right)$. Consequently, the score for opacity variation, $\mathcal{S}^{TV}_i$, of each Gaussian $g_{i,t}$ is expressed as:
\begin{equation}
    \mathcal{S}^{TV}_i = \sum^{T}_{t=0}\frac{1}{0.5\cdot\tanh(
    \left|p^{(2)}_i(t)\right|)+0.5}.
\end{equation}
In addition to the opacity range rate, the volume of 4D Gaussians is necessary to be considered, as described in \cref{eq:4dgs}. The volume should be normalized following the method in \cite{fan2023lightgaussian}, denoted as $\gamma(\mathcal{S}^{4D})=Norm(V(\mathcal{S}^{4D}))$. Therefore, the final temporal score $\mathcal{S}_i^T = \mathcal{S}_i^{TV}\gamma(S^{4D}_i)$

Finally, by aggregating both spatial and temporal score, the spatial-temporal variation score $\mathcal{S}_i$ can be written as:
\begin{equation}
\mathcal{S}_i =\sum^{T}_{t=0}\mathcal{S}_i^T\mathcal{S}_i^S
\end{equation}
\textbf{Pruning.} All 4D Gaussians are ranked based on their spatial-temporal variation score $\mathcal{S}_i$, and Gaussians with lower scores are pruned to reduce the storage burden of 4DGS~\cite{yang2023real}. The remaining Gaussians are optimized over a set number of iterations to compensate for minor losses resulting from pruning.

\subsubsection{Fast rendering with temporal filtering}
\label{subsection:filter}
Our analysis reveals that inactive Gaussians induces unnecessary computations in 4DGS, significantly slowing down rendering. To address this issue, we introduce a temporal filter that dynamically selects active Gaussians. We observed that active Gaussians in adjacent frames overlap considerably (as detailed in \cref{subsection:Delving}), which allows us to share their corresponding masks across a window of frames.

\noindent\textbf{Key-frame based Temporal Filtering.}
Based on this observation, we design a key-frame based temporal filtering for active Gaussians. We select sparse key-frames at even intervals and share their masks with surrounding frames. 

Specifically, we select a list of key-frame timestamps $ \{t_i\}_{i=0}^{T}$, where $T$ depends on the chosen interval $\Delta_t$. For each $t_i$, we render the images from all training views at current timestamp and calculate the visibility list $ \{m_{i,j}\}_{j=1}^{N}$, where $m_{i,j}$ is the visibility mask obtained by \cref{eq:blending} from the $j^{th}$ training viewpoint at timestamp $t_i$ and $N$ is the number of training views at current timestamp. The final set of active Gaussian masks is given by $\left\{ \bigcup_{j=1}^Nm_{i,j} \right\}_{i=0}^T$.

\noindent\textbf{Filter based Rendering.}
To render the images from any viewpoint at a given timestamp $t_{test}$, we consider its two nearest key-frames, denoted as $t_l$ and $t_r$. Then, we perform rasterization while only considering the Gaussians marked by mask $ \left\{ \bigcup_{j=1}^Nm_{i,j} \right\}_{i=l,r}$. This method explicitly filters out inactive Gaussians to speed up rendering.

Note that using long intervals may overlook some Gaussians, reducing rendering quality. Therefore, we fine-tune Gaussians recorded by the masks to compensate for losses.




\section{Experiment}

\begin{table*}[tp]
\centering
\caption{\textbf{Quantitative comparisons on the Neural 3D Video Dataset.}}
\begin{minipage}{\linewidth}
\centering
\footnotesize
\renewcommand{\arraystretch}{0.9}
\resizebox{0.9\linewidth}{!}{
\begin{tabular}{ccccccccc}
\hline
Method                                    & PSNR$\uparrow$  & SSIM$\uparrow$  & LPIPS$\downarrow$ & Storage(MB)$\downarrow$ &  FPS$\uparrow$ & Raster FPS$\uparrow$ & \#Gauss$\downarrow$ \\ \hline
Neural Volume$^1$\cite{lombardi2019neural}& 22.80   & -      & 0.295  & -          & -      &     -      &   -    \\
DyNeRF$^1$\cite{li2022neural}             & 29.58   & -      & 0.083  & \cellcolor{yellow!25}28         & 0.015  &     -      &   -    \\
StreamRF\cite{li2022streaming}            & 28.26   & -      & -      & 5310       & 10.90  &     -      &   -     \\
HyperReel\cite{attal2023hyperreel}        & 31.10   & 0.927  & 0.096  & 360        & 2.00   &     -      &   -     \\
K-Planes\cite{fridovich2023k}             & 31.63   & -      & 0.018  & 311        & 0.30   &     -      &   -    \\ \hline
Dynamic 3DGS\cite{luiten2023dynamic}      & 30.67   & 0.930  & 0.099  & 2764       & \cellcolor{yellow!25}460    &     -      &   -     \\
4DGaussian\cite{wu20244d}                 & 31.15   & 0.940  & \cellcolor{yellow!25}0.049 & 90         & 30     &     -      &   -      \\
E-D3DGS\cite{bae2024per}                  & 31.31   & 0.945  & \cellcolor{red!25}0.037 & 35         & 74     &     -      &   -     \\ \hline
STG\cite{li2024spacetime}                 & \cellcolor{red!25}32.05  & \cellcolor{yellow!25}0.946  & \cellcolor{orange!25}0.044 & 200        & 140    &     -      &   -     \\
4D-RotorGS\cite{duan20244d}               & 31.62   & 0.940  & 0.140  & -          & 277    &     -      &   -     \\ \hline
MEGA\cite{zhang2024mega}                  & 31.49   & -      & 0.056  & \cellcolor{orange!25}25  & 77     &     -      &   -     \\
Compact3D\cite{lee2024compact}            & 31.69   & 0.945  & 0.054  & \cellcolor{red!25}15     & 186    &     -      &   -     \\ \hline
4DGS\cite{yang2023real}                   & \cellcolor{orange!25}32.01   & -      & 0.055  & -          & 114    &     -      &   -     \\
4DGS$^2$\cite{yang2023real}               & \cellcolor{yellow!25}31.91   & \cellcolor{orange!25}0.946  & 0.052  & 2085       & 90     & 118        & 3333160 \\ 
Ours                                     & 31.88   & \cellcolor{red!25}0.946  & 0.052  & 418        & \cellcolor{red!25}\textbf{805}  & \cellcolor{red!25}\textbf{1092} & 666632 \\  
Ours-PP                                  & 31.87   & 0.944  & 0.053  & \textbf{50}             & \cellcolor{orange!25}805     & 1092       & 666632 \\ \hline
\end{tabular}}
\end{minipage}

\vspace{1ex}
\begin{minipage}{0.9\linewidth}
\raggedright
\footnotesize
\text{$^1$ The metrics of the model are tested without “coffee martini” and the resolution is set to 1024 × 768.}\\[0.5ex]
\text{$^2$ The retrained model from the official implementation.}
\end{minipage}
\label{tab:n3v}
\end{table*}

\begin{figure*}[t] \centering
    \newcommand{\hwidth}{1pt}
    \newcommand{\imgwidth}{0.2\textwidth}
    \newcommand{\patchwidth}{0.1\textwidth}
    \makebox[\imgwidth]{\small Ground Truth}
    \makebox[\imgwidth]{\small 4DGS}
    \makebox[\imgwidth]{\small Ours} 
    \makebox[\imgwidth]{\small Ours-PP} 
    \\
    \includegraphics[width=\imgwidth]{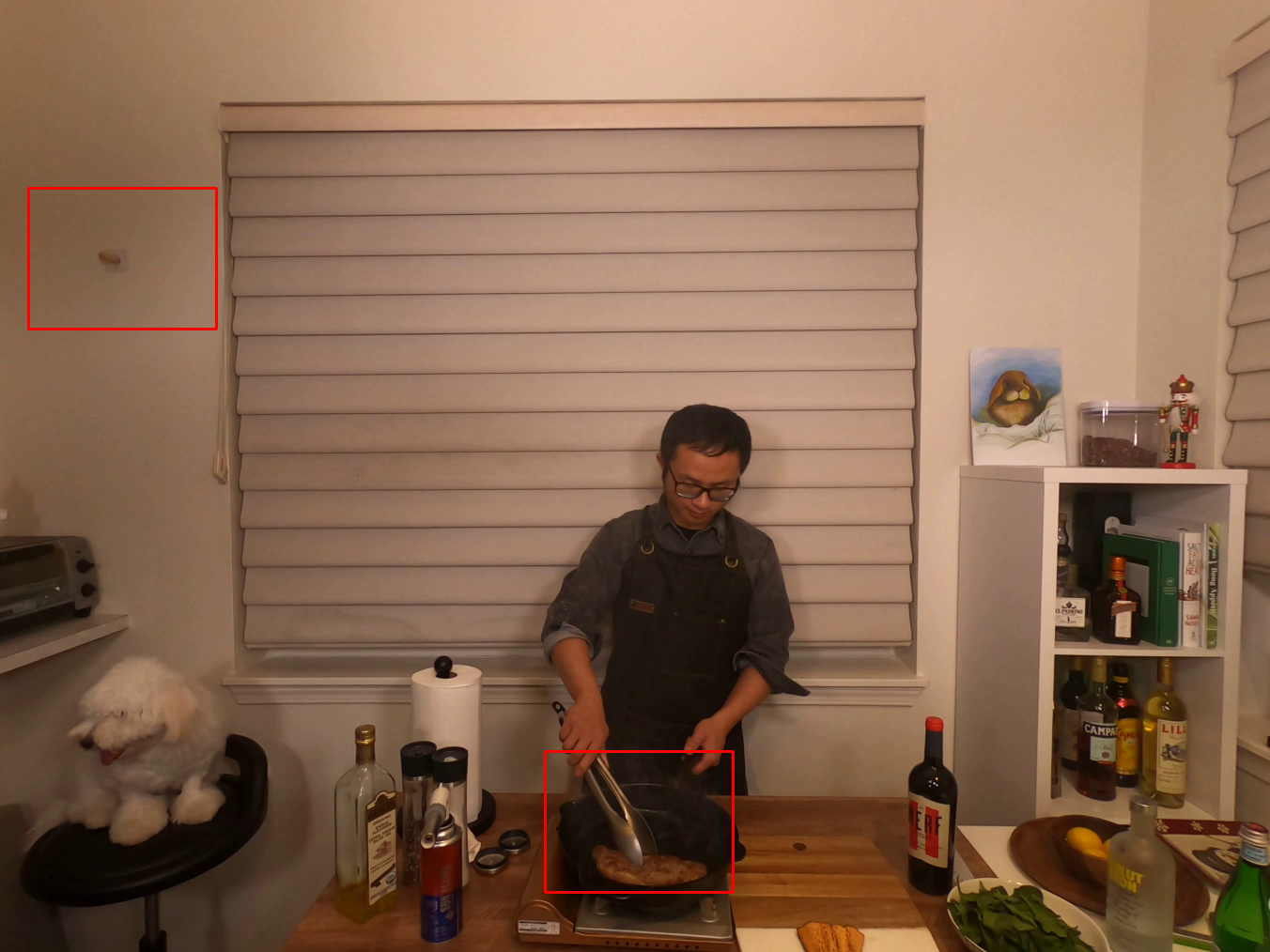}
    \includegraphics[width=\imgwidth]{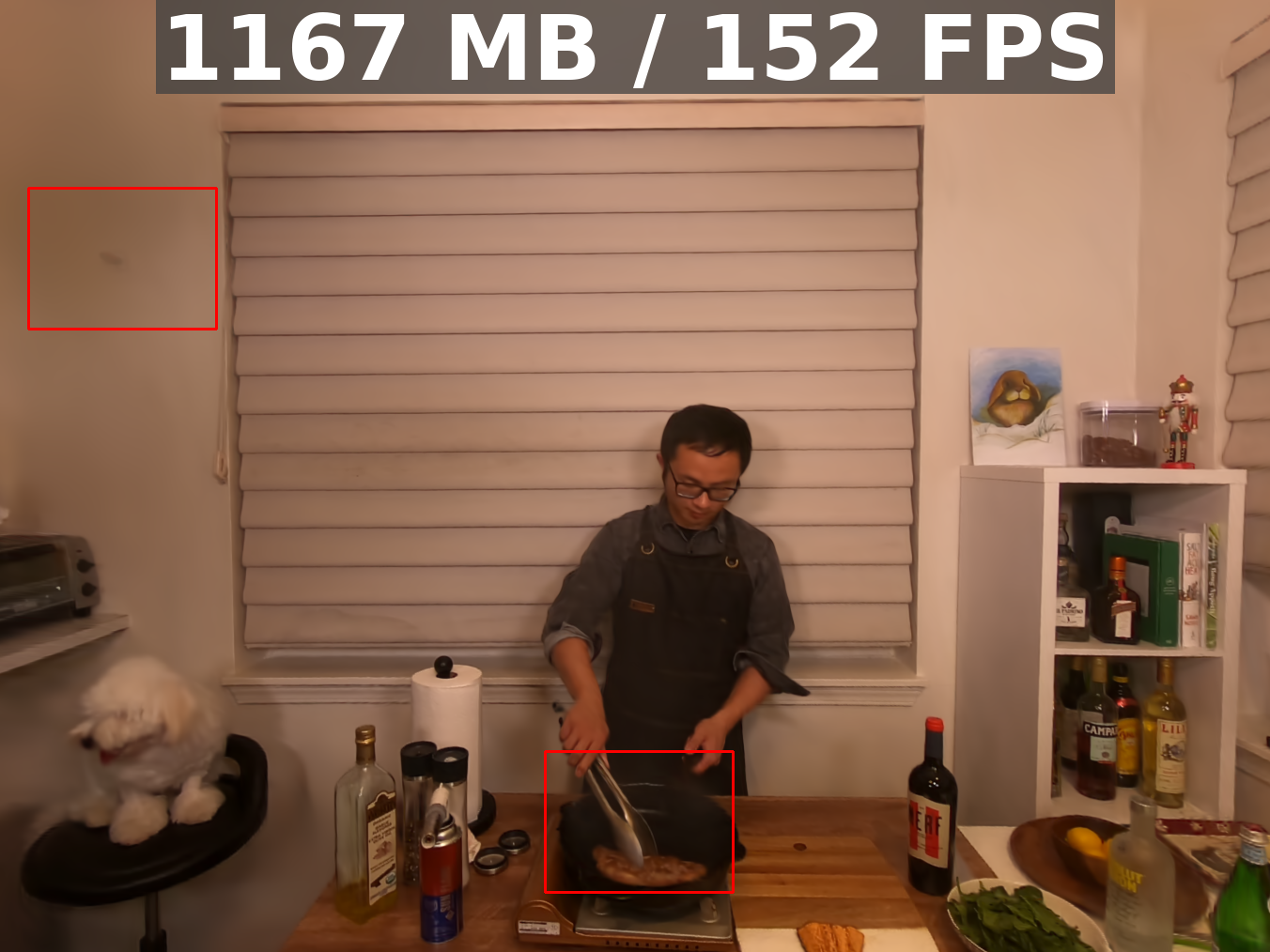}
    \includegraphics[width=\imgwidth]{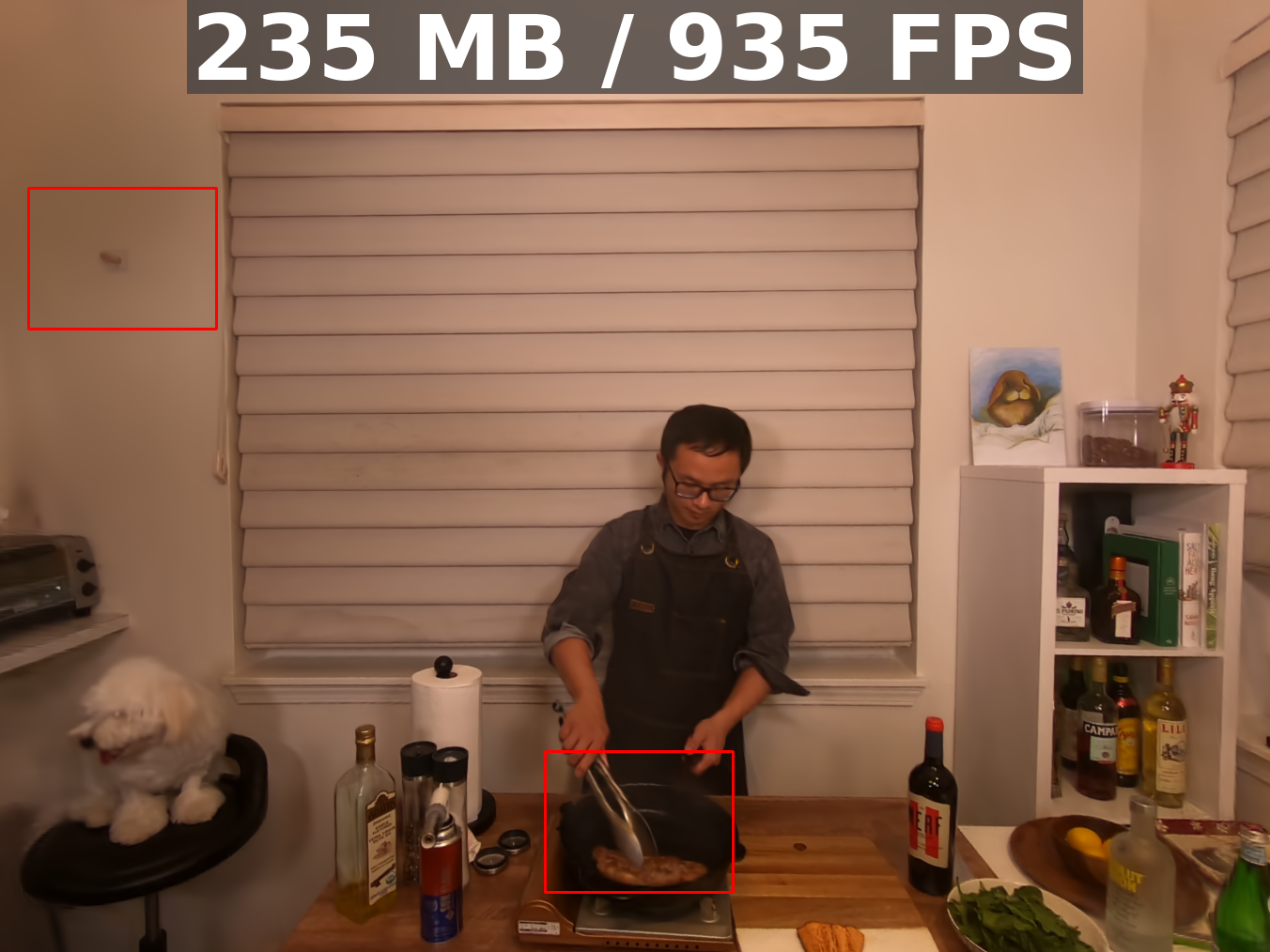}
    \includegraphics[width=\imgwidth]{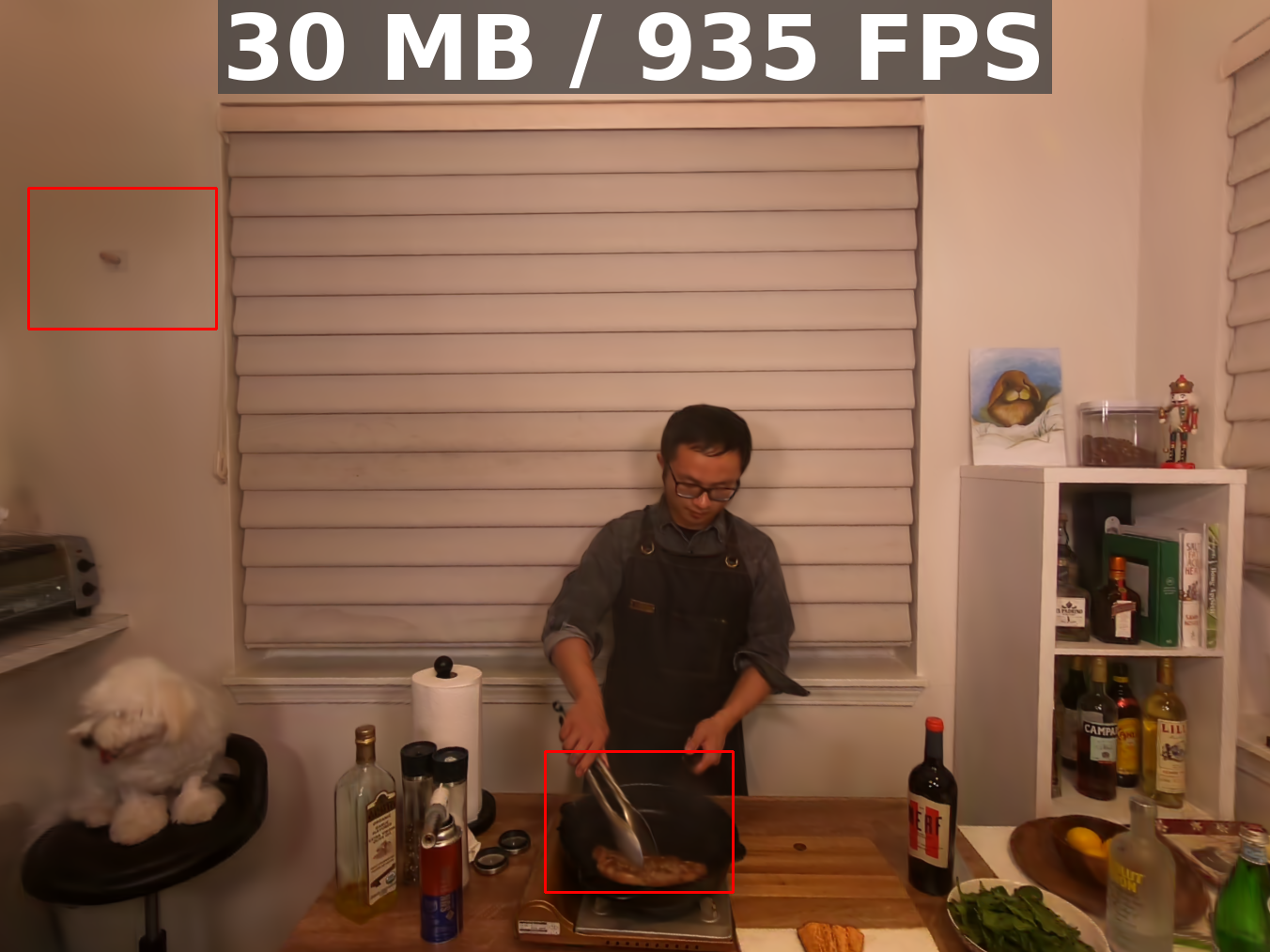}
    \\[2pt]
    \makebox[\imgwidth]{%
        \includegraphics[width=\patchwidth]{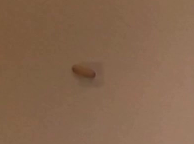}\hspace{\hwidth}
        \includegraphics[width=\patchwidth]{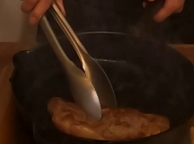}} 
    \makebox[\imgwidth]{%
        \includegraphics[width=\patchwidth]{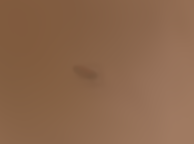}\hspace{\hwidth}
        \includegraphics[width=\patchwidth]{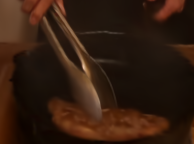}}
    \makebox[\imgwidth]{%
        \includegraphics[width=\patchwidth]{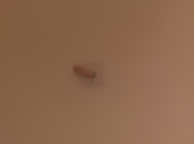}\hspace{\hwidth}
        \includegraphics[width=\patchwidth]{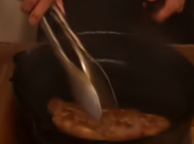}}
    \makebox[\imgwidth]{%
        \includegraphics[width=\patchwidth]{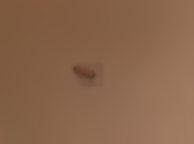}\hspace{\hwidth}
        \includegraphics[width=\patchwidth]{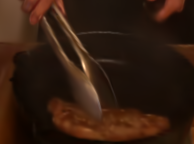}}
    \\[-0.1em]
    \makebox[\textwidth]{\small (a) Results on Sear Steak Scene.} 
    \\[0.5em]
    \makebox[\imgwidth]{\small Ground Truth}
    \makebox[\imgwidth]{\small 4DGS}
    \makebox[\imgwidth]{\small Ours} 
    \makebox[\imgwidth]{\small Ours-PP} 
    \includegraphics[width=\imgwidth]{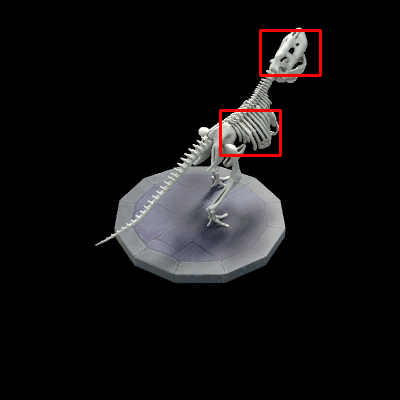}
    \includegraphics[width=\imgwidth]{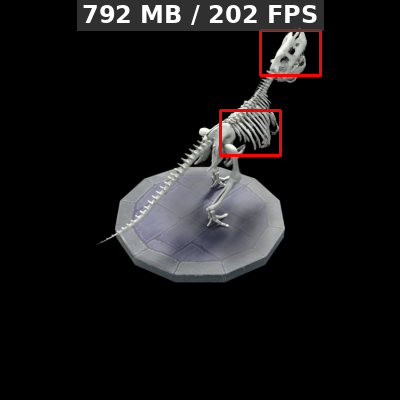}
    \includegraphics[width=\imgwidth]{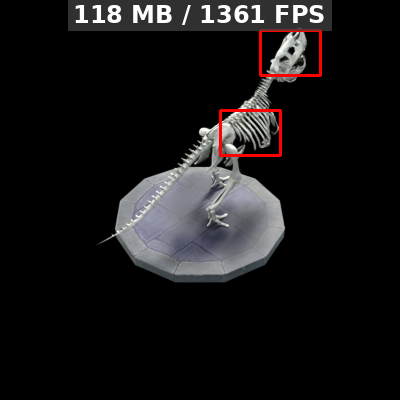}
    \includegraphics[width=\imgwidth]{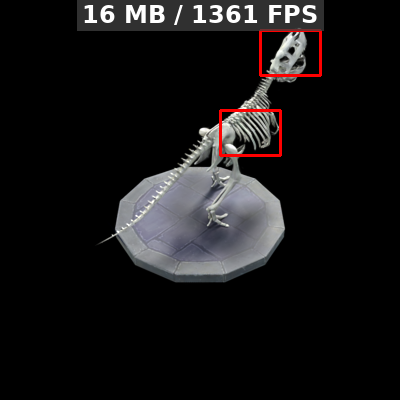}
    \\[2pt]
    
    \makebox[\imgwidth]{%
        \includegraphics[width=\patchwidth]{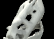}\hspace{\hwidth}
        \includegraphics[width=\patchwidth]{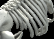}} 
    \makebox[\imgwidth]{%
        \includegraphics[width=\patchwidth]{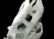}\hspace{\hwidth}
        \includegraphics[width=\patchwidth]{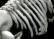}}
    \makebox[\imgwidth]{%
        \includegraphics[width=\patchwidth]{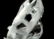}\hspace{\hwidth}
        \includegraphics[width=\patchwidth]{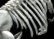}}
    \makebox[\imgwidth]{%
        \includegraphics[width=\patchwidth]{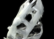}\hspace{\hwidth}
        \includegraphics[width=\patchwidth]{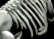}}
    \\[-0.1em]
    \makebox[\textwidth]{\small (b) Results on Trex Scene.} 
    \\[0.5em]
    \caption{\textbf{Qualitative comparisons of 4DGS and our method.}} 
    \label{fig:qualitative}
\end{figure*}

\subsection{Experimental Settings}
\begin{table*}[tp]
\centering
\caption{\textbf{Quantitative comparisons on the D-NeRF Dataset.}}
\begin{minipage}{\linewidth}
\centering
\footnotesize
\renewcommand{\arraystretch}{0.9}
\resizebox{0.9\linewidth}{!}{
\begin{tabular}{ccccccccc}
\hline
Method                                    & PSNR$\uparrow$  & SSIM$\uparrow$  & LPIPS$\downarrow$ & Storage(MB)$\downarrow$ &  FPS$\uparrow$ & Raster FPS$\uparrow$ & \#Gauss$\downarrow$ \\ \hline
DNeRF\cite{pumarola2021d}                & 29.67  & 0.95   & 0.08 & -     & 0.1  & -      & -    \\
TiNeuVox\cite{fang2022fast}              & 32.67  & 0.97   & 0.04 & -     & 1.6  & -      & -     \\
K-Planes\cite{fridovich2023k}            & 31.07  & 0.97   & \cellcolor{yellow!25}0.02 & -  & 1.2  & -      & -    \\ \hline
4DGaussian\cite{wu20244d}               & 32.99  & 0.97   & 0.05 & \cellcolor{orange!25}18   & 104  & -      & -      \\
Deformable3DGS\cite{yang2024deformable}  & \cellcolor{red!25}40.43  & \cellcolor{red!25}0.99   & \cellcolor{red!25}0.01 & \cellcolor{yellow!25}27 & 70   & -      & 131428     \\ \hline
4D-RotorGS\cite{duan20244d}             & \cellcolor{orange!25}34.26  & 0.97   & 0.03 & 112   & \cellcolor{yellow!25}1257 & -      & -     \\
4DGS\cite{yang2023real}                 & \cellcolor{yellow!25}34.09  & \cellcolor{orange!25}0.98   & \cellcolor{orange!25}0.02 & -  & -    & -        & -        \\
4DGS$^1$\cite{yang2023real}             & 32.99  & 0.97   & 0.03 & 278   & 376  & 1232   & 445076     \\
Ours                                    & 33.34  & \cellcolor{yellow!25}0.97   & 0.03 & 42    & \cellcolor{red!25}\textbf{1462}  & \cellcolor{red!25}\textbf{2482}  & 66460   \\
Ours-PP                                 & 33.37  & 0.97   & 0.03 & \cellcolor{red!25}\textbf{7}  & \cellcolor{orange!25}1462    & 2482      & 66460 \\ \hline
\end{tabular}}
\end{minipage}

\vspace{1ex}
\begin{minipage}{0.9\linewidth}
\raggedright
\footnotesize
\text{$^1$ The retrained model from the official implementation.}
\end{minipage}
\label{tab:dnerf}
\vspace{-4mm}
\end{table*}

\begin{table*}[tp]
\centering
\caption{\textbf{Ablation study of per-component contribution.}}
\label{tab:ablation}
\begin{minipage}{\linewidth}
\centering
\small
\renewcommand{\arraystretch}{0.9}
\resizebox{0.9\linewidth}{!}{
\begin{tabular}{ccccccccccc}
\hline
ID & \multicolumn{3}{c}{Method\textbackslash{}Dataset} & \multirow{2}{*}{PSNR$\uparrow$} & \multirow{2}{*}{SSIM$\uparrow$} & \multirow{2}{*}{LPIPS$\downarrow$} & \multirow{2}{*}{Storage(MB)$\downarrow$} & \multirow{2}{*}{FPS$\uparrow$} & \multirow{2}{*}{Raster FPS$\uparrow$} & \multirow{2}{*}{\#Gauss$\downarrow$} \\ \cline{1-4}
   & Filter & Pruning & PP & & & & & & & \\ \hline
a & \multicolumn{3}{c}{vanilla 4DGS$^{1}$}      & \cellcolor{orange!25}31.91   & \cellcolor{orange!25}0.9458 & \cellcolor{orange!25}0.0518 & 2085 & 90  & 118   & 3333160 \\
b & \checkmark$^{1,2}$       &         &      & 31.51   & 0.9446 & 0.0539 & 2091 & 242   & 561   & 3333160 \\
c & \checkmark$^{2}$         &         &      & 29.56   & 0.9354 & 0.0605 & 2091 & 300   & 561   & 3333160 \\
d &                          & \checkmark &      & \cellcolor{red!25}31.92  & \cellcolor{red!25}0.9462 & \cellcolor{red!25}0.0513 & \cellcolor{orange!25}417  & 312   & 600   & 666632 \\
e & \checkmark               & \checkmark &      & \cellcolor{yellow!25}31.88   & \cellcolor{yellow!25}0.9457 & \cellcolor{yellow!25}0.0524 & \cellcolor{yellow!25}418  & \cellcolor{orange!25}805   & \cellcolor{orange!25}1092  & 666632 \\
f & \checkmark$^{2}$         & \checkmark &      & 31.63   & 0.9452 & 0.0524 & 418  & \cellcolor{yellow!25}789   & \cellcolor{yellow!25}1080  & 666632 \\
g & \checkmark               & \checkmark & \checkmark & 31.87  & 0.9444 & 0.0532 & \cellcolor{red!25}50  & \cellcolor{red!25}805   & \cellcolor{red!25}1092  & 666632 \\ \hline
\end{tabular}
}
\end{minipage}

\vspace{1ex}
\begin{minipage}{0.9\linewidth}
\raggedright
\footnotesize
\text{$^1$ The result with environment map. $^2$ The result without finetuning.}
\end{minipage}
\end{table*}

\textbf{Datasets.} We utilize two dynamic scene datasets to demonstrate the effectiveness of our method: (1) \textbf{Neural 3D Video Dataset (N3V)}~\cite{li2022neural}. This dataset consists of six dynamic scenes, and the resolution is $2704\times2028$. For a fair comparison, we align with previous work~\cite{yang2023real,li2024spacetime} by conducting evaluations at a half-resolution of 300 frames. (2) \textbf{D-NeRF Dataset}~\cite{pumarola2021d}. This dataset is a monocular video dataset comprising eight videos of synthetic scenes. We choose standard test views that originate from novel camera positions not encountered during the training process.

\noindent\textbf{Evaluation Metrics.} To evaluate the quality of rendering dynamic scenes, we employ several commonly used image quality assessment metrics: Peak Signal-to-Noise Ratio (PSNR), Structural Similarity Index Measure (SSIM), and Learned Perceptual Image Patch Similarity (LPIPS)~\cite{zhang2018unreasonable}. Following the previous work, LPIPS~\cite{zhang2018unreasonable} is computed using AlexNet~\cite{krizhevsky2012imagenet} and VGGNet~\cite{simonyan2014very} on the N3V dataset and the D-NeRF dataset, respectively. Moreover, we report the number of Gaussians and storage. To demonstrate the improvement in rendering speed, we report two types of FPS: (1) \textbf{FPS.} It considers the entire rendering function. Due to interference from other operations, it can't effectively demonstrate the acceleration achieved by our method. (2) \textbf{Raster FPS.} It only considers the rasterization, the most computationally intensive component during rendering.

\noindent\textbf{Baselines.} Our primary baseline for comparison is 4DGS~\cite{yang2023real}, which serves as the foundation of our model. Moreover, we compare 4DGS-1K with two
concurrent works on 4D compression, MEGA~\cite{zhang2024mega} and Compact3D~\cite{lee2024compact}. Certainly, we conduct comparisons with 4D-RotorGS~\cite{duan20244d} which is another form of representation for 4D Gaussian Splatting with the capability for real-time rendering speed and high-fidelity rendering results. In addition, we also compare our work against NeRF-based methods, like Neural Volume~\cite{lombardi2019neural}, DyNeRF~\cite{li2022neural}, StreamRF~\cite{li2022streaming}, HyperReel~\cite{attal2023hyperreel}, DNeRF~\cite{pumarola2021d} and K-Planes~\cite{fridovich2023k}. Furthermore, other recent competitive Gaussian-based methods are also considered in our comparison, including Dynamic 3DGS~\cite{luiten2023dynamic}, STG~\cite{li2024spacetime}, 4DGaussian~\cite{wu20244d}, and E-D3DGS~\cite{bae2024per}. 

\noindent\textbf{Implementation Details.} Our method is tested in a single RTX 3090 GPU. We train our model following the experiment setting in 4DGS~\cite{yang2023real}. After training, we perform the pruning and filtering strategy. Then, we fine-tune 4DGS-1K for 5,000 iterations while disabling additional clone/split operations. For pruning strategy, the pruning ratio is set to $80\%$ on the N3V Dataset, and $85\%$ on the D-NeRF Dataset. For the temporal filtering, we set the interval $\Delta_t$ between key-frames to $20$ frames on the N3V Dataset. Considering the varying capture speeds on the D-NeRF dataset, we select $6$ key-frames rather than a specific frame interval. Additionally, to further compress the storage of 4DGS~\cite{yang2023real}, we implement post-processing techniques in our model, denoted as Ours-PP. It includes vector quantization~\cite{navaneet2024compgs} on SH of Gaussians and compressing the mask of filter into bits. 

Note that we don't apply environment maps implemented by 4DGS on Coffee Martini and Flame Salmon scenes, which significantly affects the rendering speed. Subsequent results indicate that removing it for 4DGS-1K does not significantly degrade the rendering quality. 

\subsection{Results and Comparisons}

\textbf{Comparisons on real-world dataset.} \cref{tab:n3v} presents a quantitative evaluation on the N3V dataset. 4DGS-1K achieves rendering quality comparable to the current baseline. Compared to 4DGS~\cite{yang2023real}, we achieve a $41\times$ compression and $9\times$ faster in rendering speed at the cost of a $0.04dB$ reduction in PSNR. In addition, compared to MEGA~\cite{zhang2024mega} and Compact3D~\cite{lee2024compact}, two
concurrent works on 4D compression, the rendering speeds are 10$\times$ and 4$\times$ faster respectively while maintaining a comparable storage requirement and high quality reconstruction. Moreover, the FPS of 4DGS-1K far exceeds the current state-of-the-art levels. It is nearly twice as fast as the current fastest model, Dynamic 3DGS~\cite{luiten2023dynamic} while requiring only $\textbf{1\%}$ of the storage size. Additionally, 4DGS-1K achieves better visual quality than that of Dynamic 3DGS~\cite{luiten2023dynamic}, with an increase of about $1.2dB$ in PSNR. Compared to the storage-efficient model, E-D3DGS~\cite{bae2024per} and DyNeRF~\cite{li2022neural} we achieve an increase of over $0.5dB$ in PSNR and fast rendering speed. \cref{fig:qualitative} offers qualitative comparisons for the Sear Steak, demonstrating that our results contain more vivid details.

\noindent\textbf{Comparisons on synthetic dataset.} In our experiments, we benchmarked 4DGS-1K against several baselines using the monocular synthetic dataset introduced by D-NeRF~\cite{pumarola2021d}. The result is shown in \cref{tab:dnerf}. Compared to 4DGS~\cite{yang2023real}, our method achieves up to $40\times$ compression and $4\times$ faster rendering speed. It is worth noting that the rendering quality of our model even surpasses that of the original 4DGS, with an increase of about $0.38dB$ in PSNR. Furthermore, our approach exhibits higher rendering quality and smaller storage overhead compared to most Gaussian-based methods. We provide qualitative results in~\cref{fig:qualitative} for a more visual assessment.

\subsection{Ablation Study}
To evaluate the contribution of each component and the effectiveness of the pruning strategy for temporal filtering, we conducted ablation experiments on the N3V dataset~\cite{li2022neural}. More ablations are provided in the supplement(See~\cref{sec:add_abl}).

\begin{table}[h]
\caption{\textbf{Ablation study of Spatial-Temporal Variation Score.} We compare our Spatial-Temporal Variation Score with other variants, and report the PSNR score of each scene.}
\centering
\footnotesize
\renewcommand{\arraystretch}{0.9}
\begin{tabular}{cccc}
\hline
ID&Model                                        & Sear Steak & Flame Salmon \\ \hline
a&4DGS w/o Prune                                        &      33.60 &    29.10          \\
b&$\mathcal{S}^S_{i}$ Only                     &      33.62 &    28.75     \\
c&$\mathcal{S}^T_{i}$ Only                     &      33.59 &    28.79    \\ \hline
d&$\mathcal{S}_{i}$ (w. $p^{(1)}_{i}(t)$)      &      33.67 &    28.81     \\
e&$\mathcal{S}_{i}$ (w. $\Sigma_t$)            &      33.47 &    28.71     \\ \hline
f&Ours                                         &      33.76 &    28.90     \\ \hline
\end{tabular}

\label{tab:ablation_score}
\end{table}

\noindent\textbf{Pruning.}
As shown in~\cref{tab:ablation}, our pruning strategy reduces the number of Gaussians by $80\%$, and achieves $5\times$ compression ratio and $5\times$ faster rasterization speed while slightly improving rendering quality. As shown in \cref{fig:distribution}, our pruning strategy also reduces the presence of Gaussians with short lifespan. As such, 4DGS-1k processes far fewer unnecessary Gaussians~(See \cref{fig:ratio}) during rendering. 

Furthermore, we compare our Spatial-Temporal Variation Score with serveral variants. Specific settings are described in~\cref{sec:moreablation}. As shown in~\cref{tab:ablation_score}, using spatial and temporal scores separately reduce the PSNR.  This occurs because separate scores can amplify extreme Gaussians. For instance, using only the spatial score (b) may retain Gaussians that cover just a single frame but occupy a large spatial volume. Our combined score balances these factors. For variant d, using the first derivative may cause some small Gaussians to have large $\mathcal{S}^T_{i}$ compared to ours. Moreover, since most Gaussians have small $\Sigma_t$, it is difficult to distinguish them by using $\Sigma_t$ along~(See e). 
Moreover, as shown in~\cref{fig:iou}, the pruning process expands the range of adjacent frames. It allows larger intervals for the temporal filter. We will discuss it in the next part.

\noindent\textbf{Temporal Filtering.} As illustrated in \cref{tab:ablation}, the results of b and c are obtained by directly applying the filter to 4DGS without fine-tuning. It proves that this component can enhance the rendering speed of 4DGS. However, as mentioned in \cref{subsection:Delving}, the 4DGS contains a huge number of short lifespan Gaussians. It results in some Gaussians being overlooked in the filter, causing a slight decrease in rendering quality. However, through pruning, most Gaussians are ensured to have long lifespan, making them visible even at large intervals. Therefore, it alleviates the issue of Gaussians being overlooked~(See f). Furthermore, appropriate fine-tuning allows the Gaussians in the active Gaussians list to relearn the scene features to compensate for the loss incurred by the temporal filter~(See e and f).

\section{Conclusion}

In this paper, we present \textbf{4DGS-1K}, a compact and memory-efficient dynamic scene representation capable of running at over 1000 FPS on modern GPUs. We introduce a novel pruning criterion called the spatial-temporal variation score, which eliminates a significant number of redundant Gaussian points in 4DGS, drastically reducing storage requirements. Additionally, we propose a temporal filter that selectively activates only a subset of Gaussians during each frame's rendering. This approach enables our rendering speed to far surpass that of existing baselines. Compared to vanilla 4DGS, 4DGS-1K achieves a $41 \times$ reduction in storage and $9\times$ faster rasterization speed while maintaining high-quality reconstruction.


\clearpage
\maketitlesupplementary

\noindent The Supplementary material is organized as follows:
\begin{itemize}
    \item \cref{sec:perscene}: provides additional visualization results and quantitative results. Furthermore, it also shows the resource consumption which reveals the potential of 4DGS-1K for deployment on low-performance hardware. 
    \item \cref{sec:add_abl}: provides additional ablation study. It firstly provides the variant settings in the main text, then it presents more additional ablation study to illustrate that our parameter selection is the result of a trade-off between rendering quality and storage size.
    \item \cref{sec:discussion}: discusses the reason of improved performance for 4DGS-1K. Furthermore, we introduce the limitations and potential future directions of 4DGS-1K.
\end{itemize}

\section{Experimental Results}
\label{sec:perscene}
\subsection{Per scene result} 
\begin{figure*}[b] \centering
    \newcommand{\hwidth}{1pt}
    \newcommand{\imgwidth}{0.24\textwidth}
    \newcommand{\patchwidth}{0.115\textwidth}
    \makebox[\imgwidth]{\small Ground Truth}
    \makebox[\imgwidth]{\small 4DGS}
    \makebox[\imgwidth]{\small Ours} 
    \makebox[\imgwidth]{\small Ours-PP} 
    \includegraphics[width=\imgwidth]{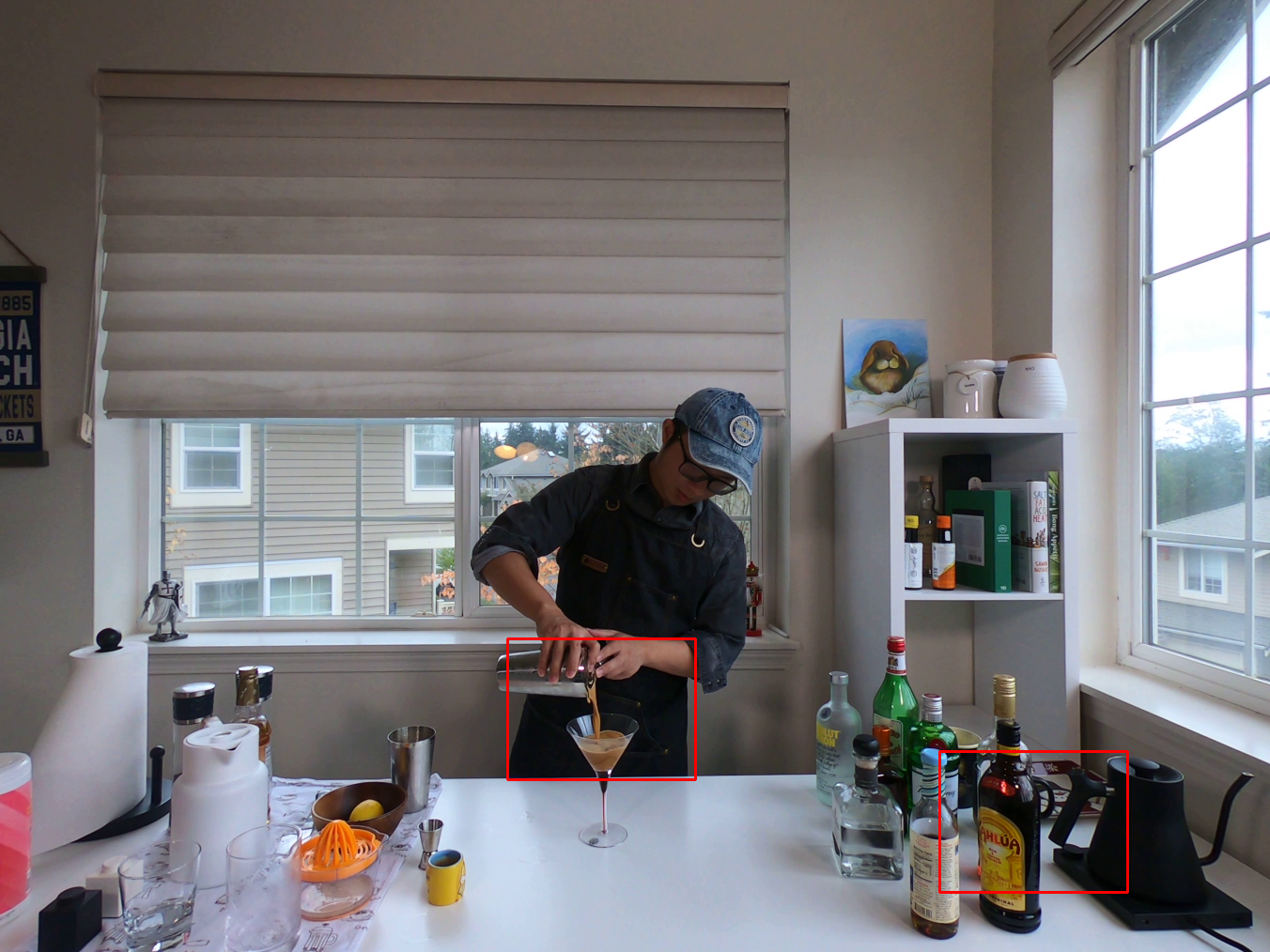}
    \includegraphics[width=\imgwidth]{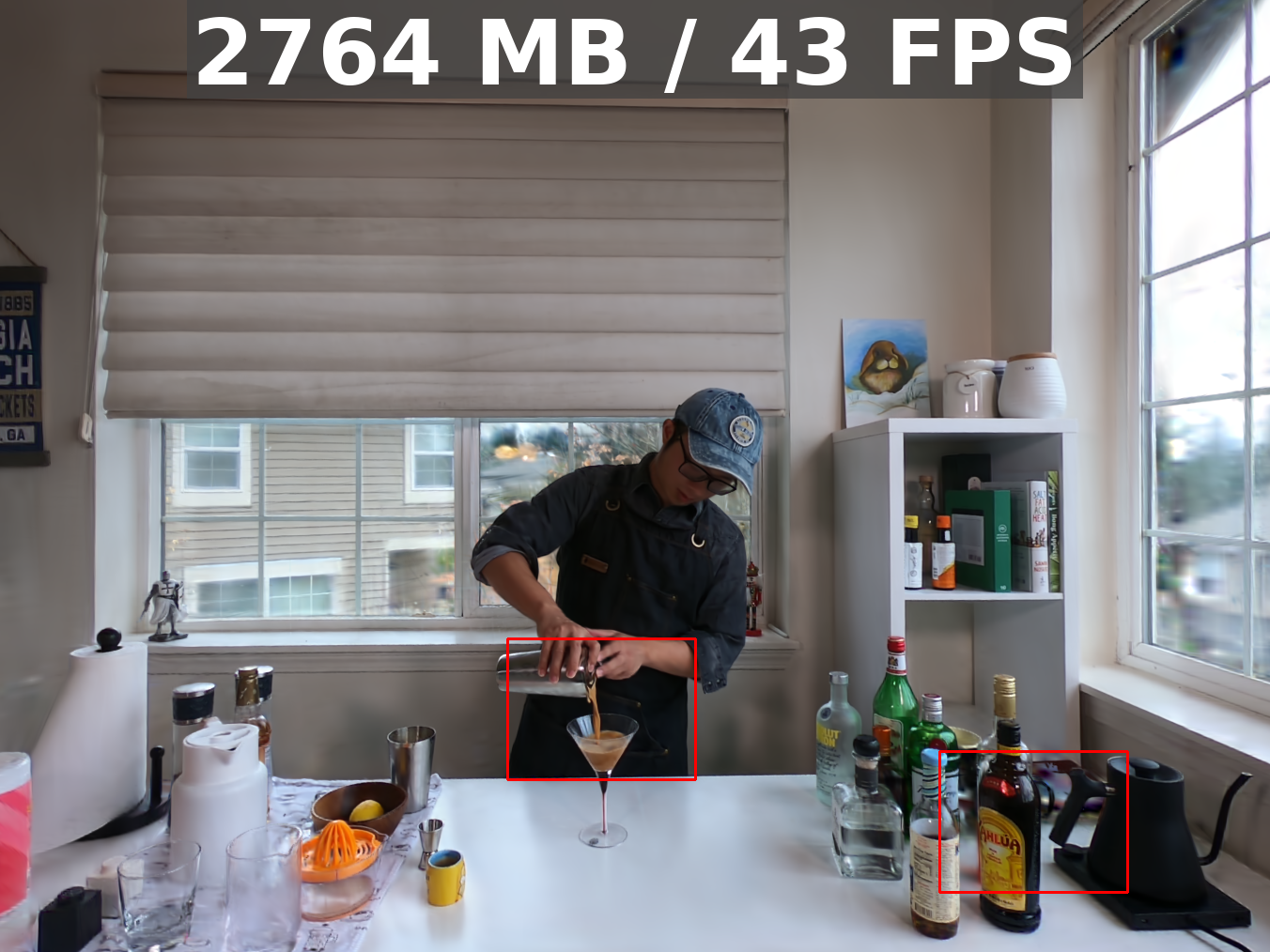}
    \includegraphics[width=\imgwidth]{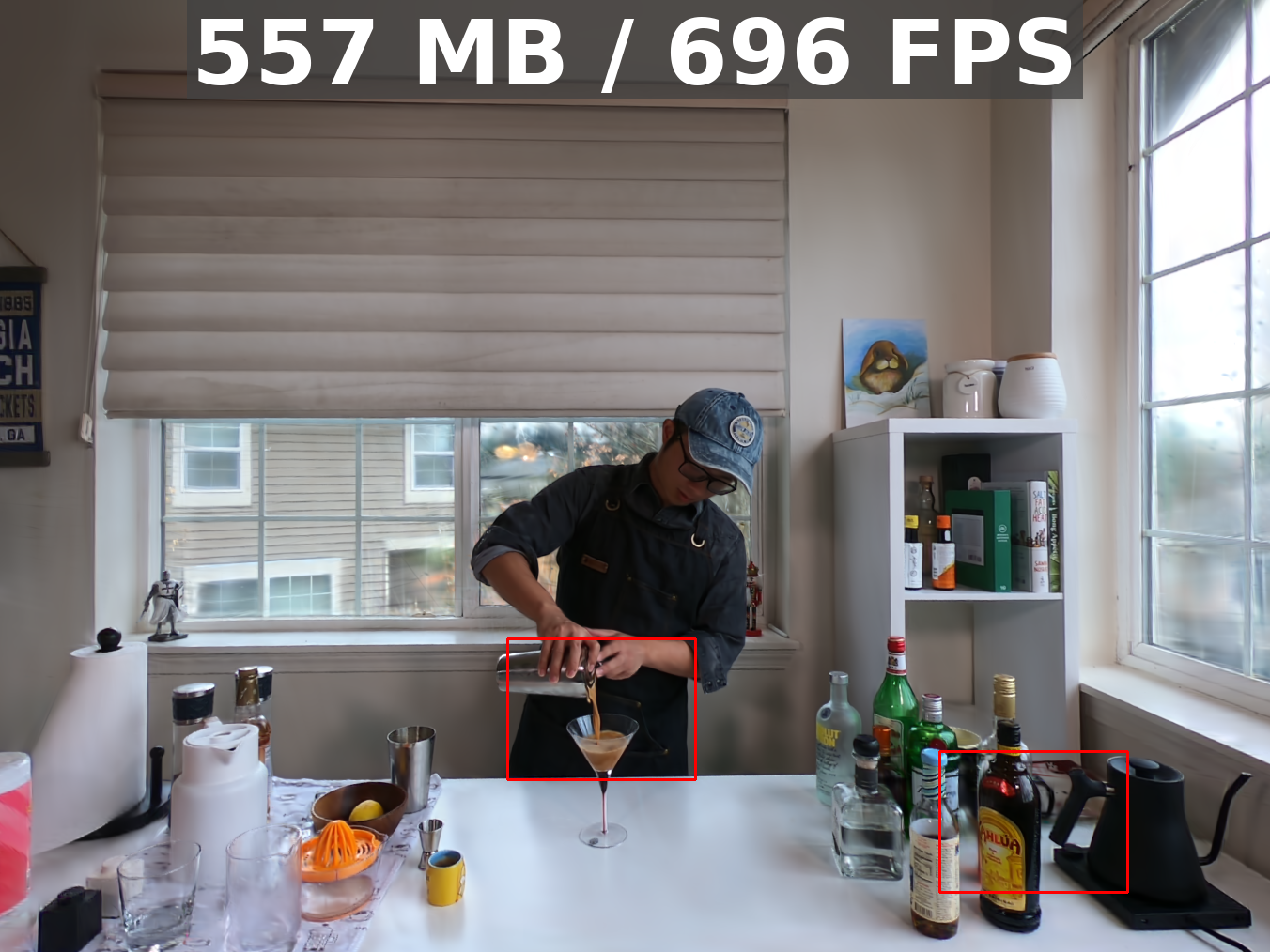}
    \includegraphics[width=\imgwidth]{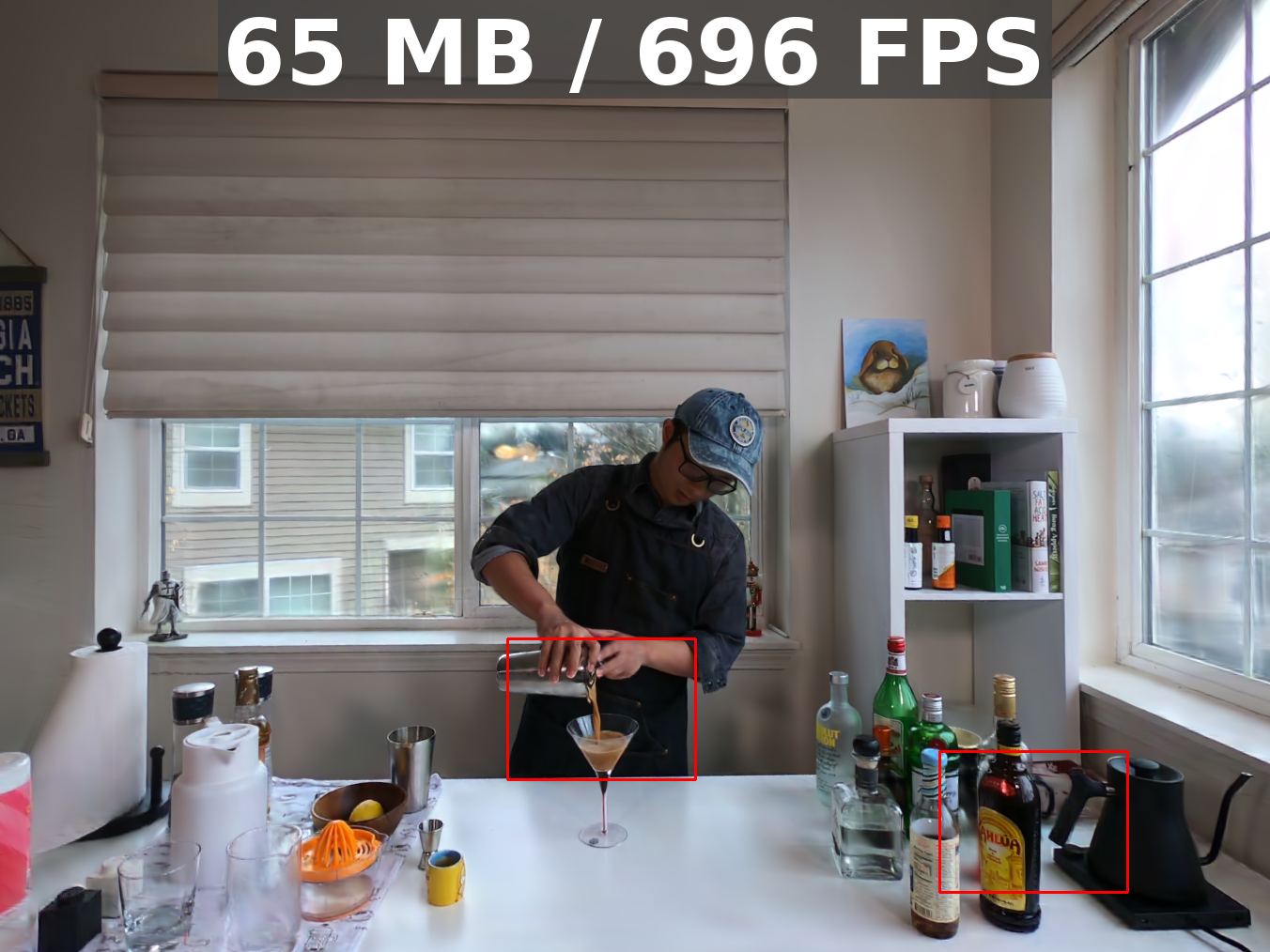}
    \\[2pt]
    
    \makebox[\imgwidth]{%
        \includegraphics[width=\patchwidth]{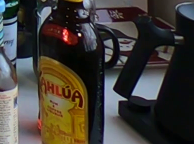}\hspace{\hwidth}
        \includegraphics[width=\patchwidth]{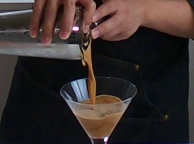}} 
    \makebox[\imgwidth]{%
        \includegraphics[width=\patchwidth]{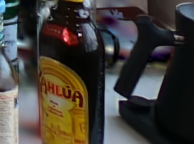}\hspace{\hwidth}
        \includegraphics[width=\patchwidth]{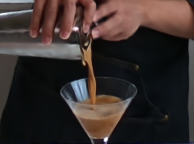}}
    \makebox[\imgwidth]{%
        \includegraphics[width=\patchwidth]{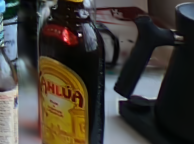}\hspace{\hwidth}
        \includegraphics[width=\patchwidth]{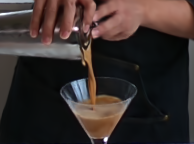}}
    \makebox[\imgwidth]{%
        \includegraphics[width=\patchwidth]{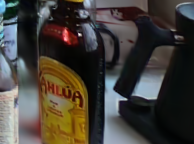}\hspace{\hwidth}
        \includegraphics[width=\patchwidth]{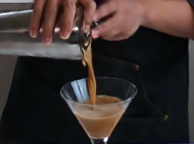}}
    \\[-0.1em]
    \makebox[\textwidth]{\small (a) Results on Coffee Martini Scene.} 
    \\[0.5em]
    \includegraphics[width=\imgwidth]{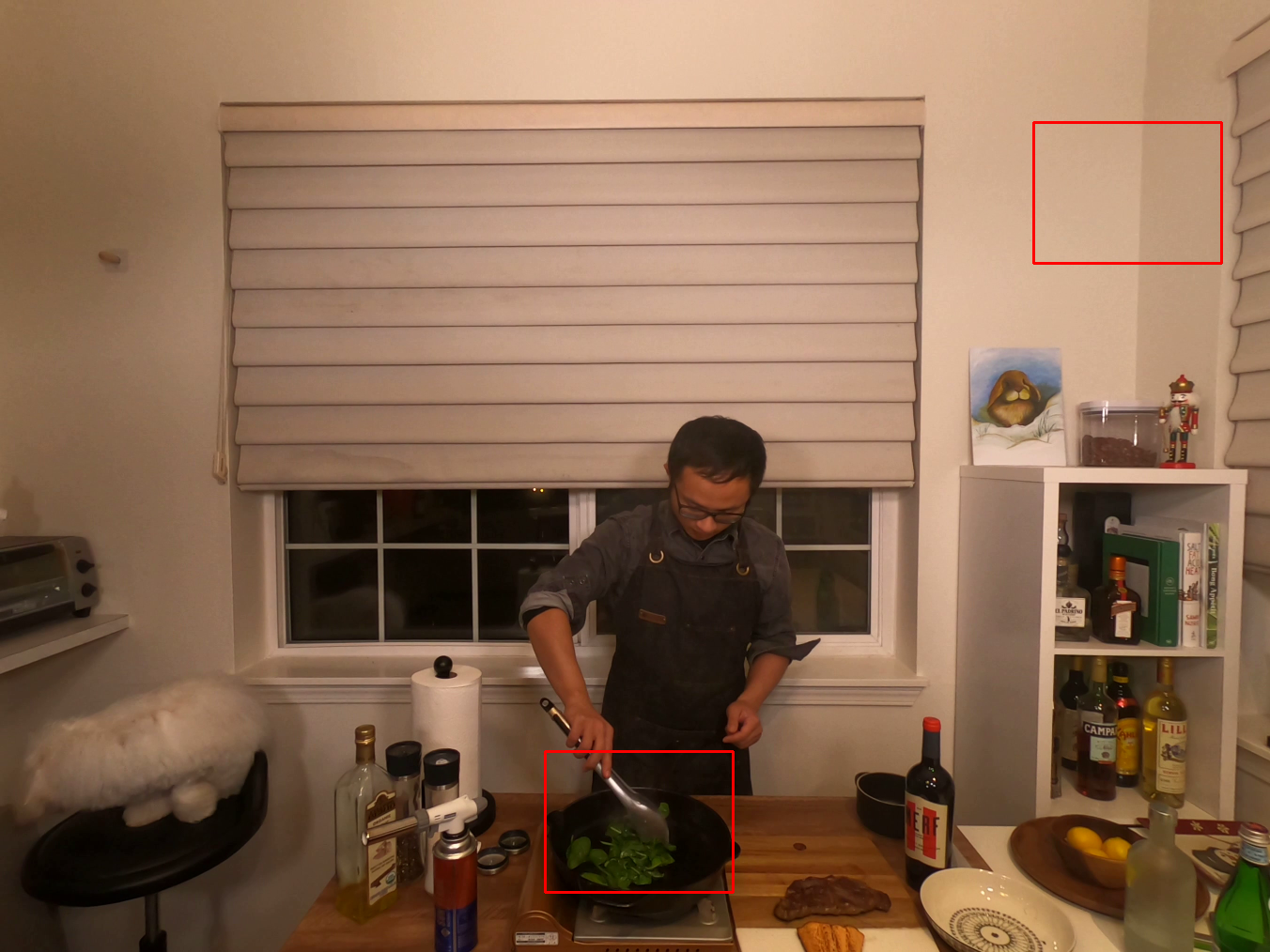}
    \includegraphics[width=\imgwidth]{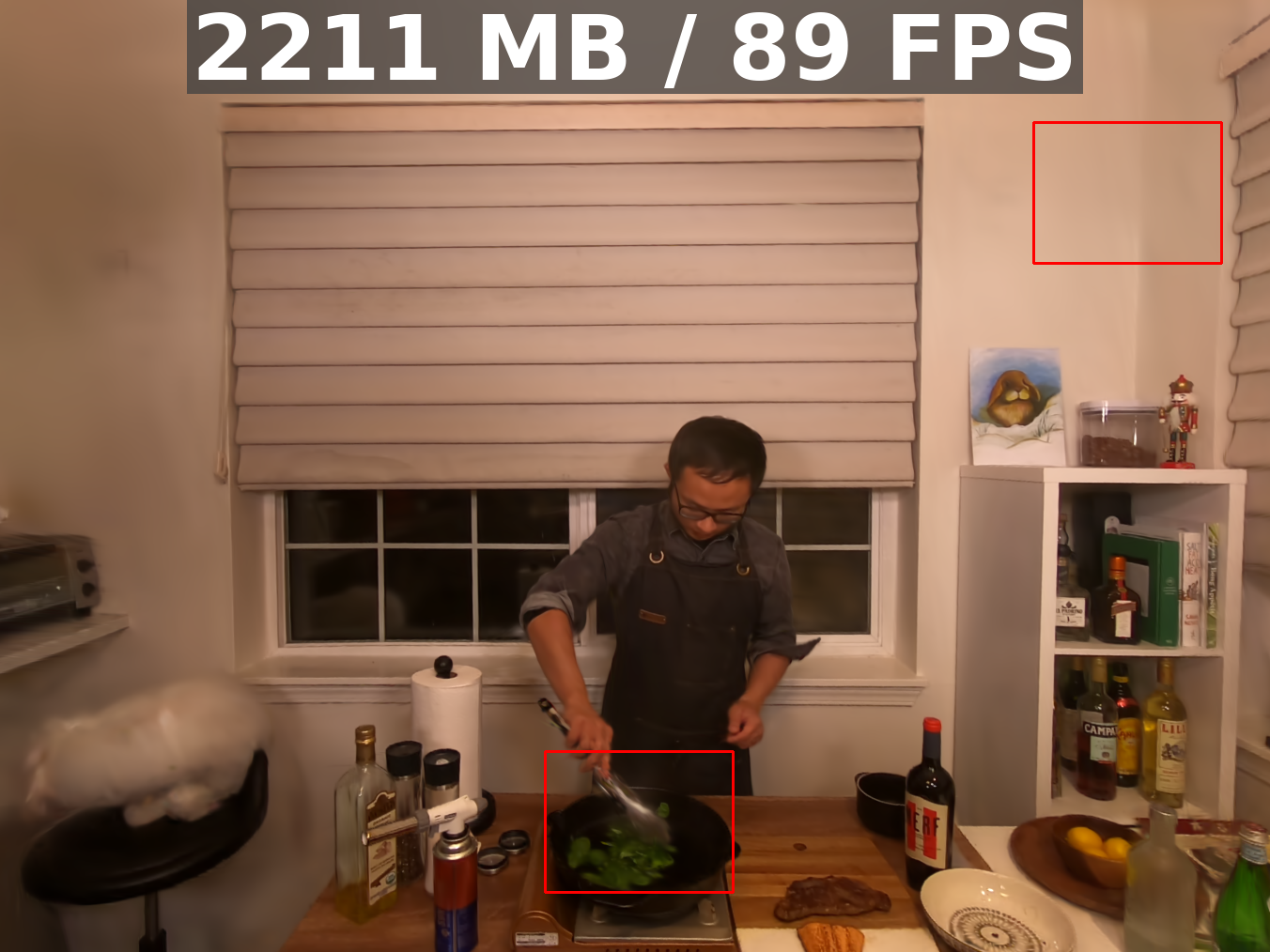}
    \includegraphics[width=\imgwidth]{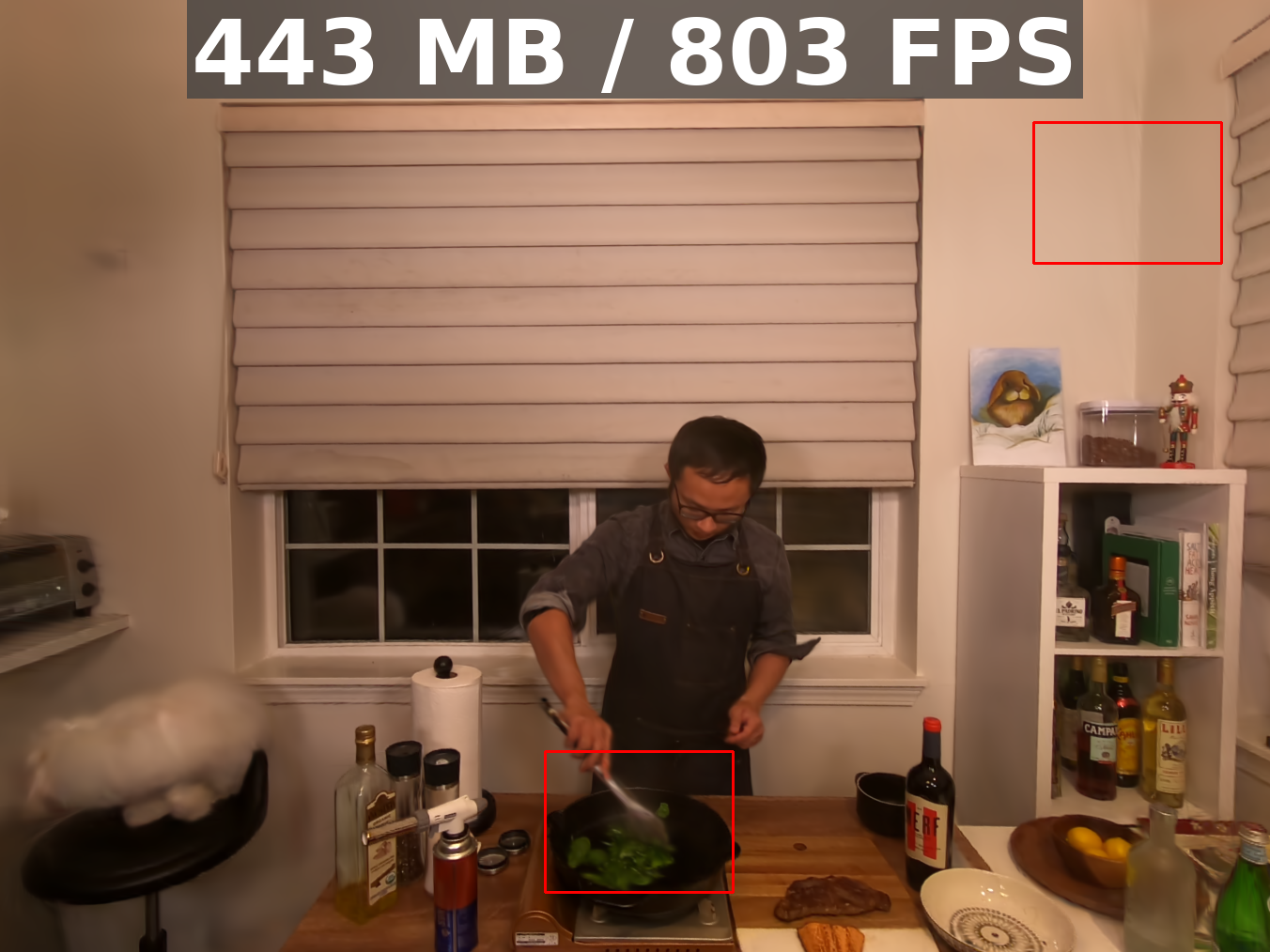}
    \includegraphics[width=\imgwidth]{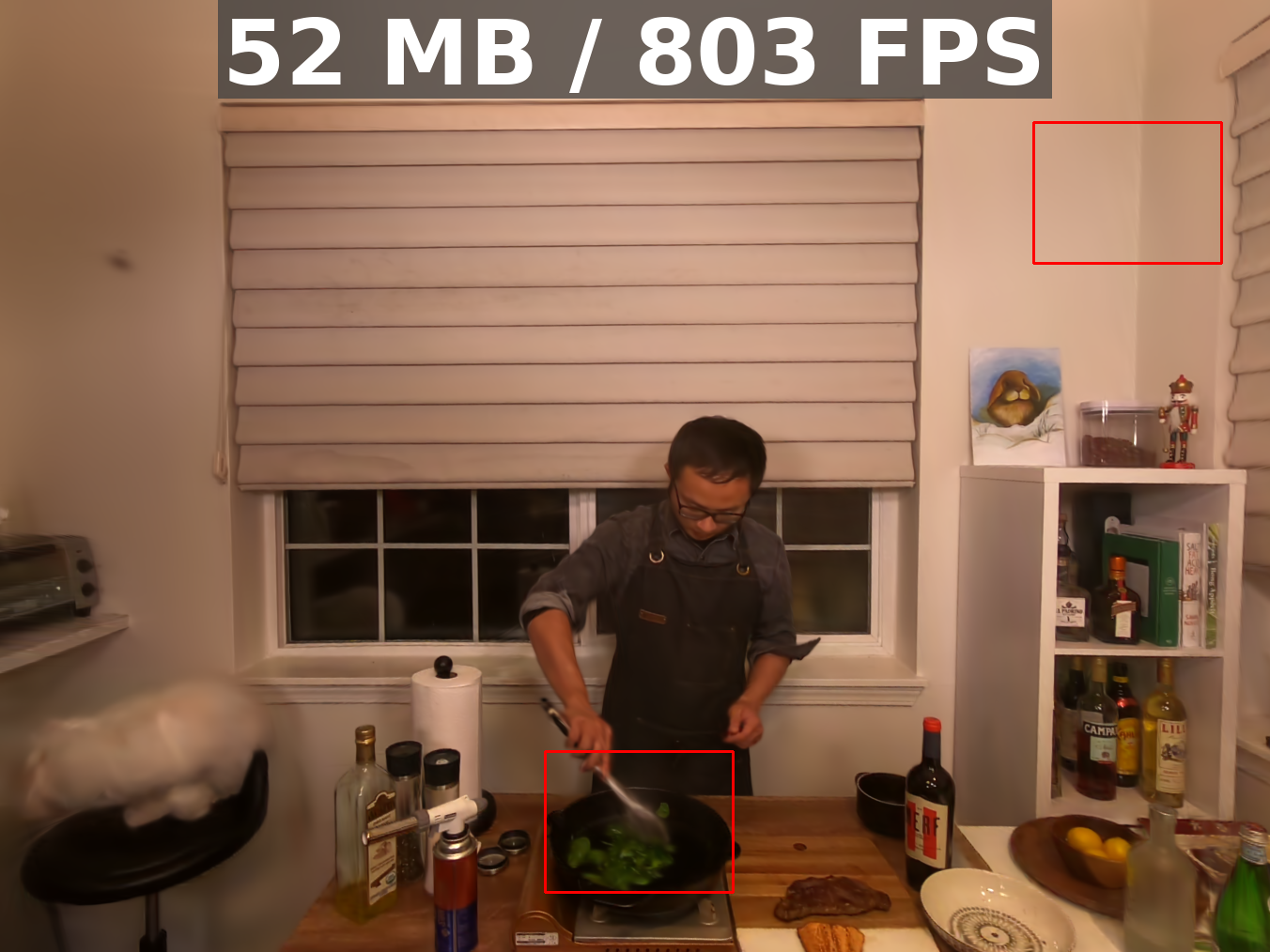}
    \\[2pt]
    
    \makebox[\imgwidth]{%
        \includegraphics[width=\patchwidth]{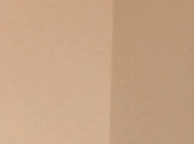}\hspace{\hwidth}
        \includegraphics[width=\patchwidth]{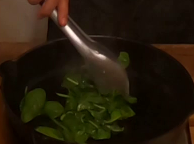}} 
    \makebox[\imgwidth]{%
        \includegraphics[width=\patchwidth]{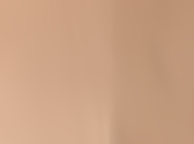}\hspace{\hwidth}
        \includegraphics[width=\patchwidth]{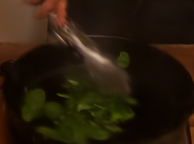}}
    \makebox[\imgwidth]{%
        \includegraphics[width=\patchwidth]{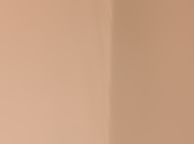}\hspace{\hwidth}
        \includegraphics[width=\patchwidth]{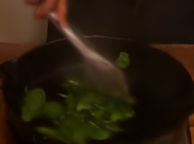}}
    \makebox[\imgwidth]{%
        \includegraphics[width=\patchwidth]{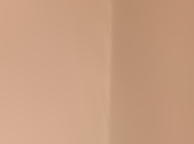}\hspace{\hwidth}
        \includegraphics[width=\patchwidth]{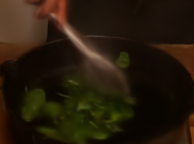}}
    \\[-0.1em]
    \makebox[\textwidth]{\small (b) Results on Cook Spinach Scene.} 
    \\[0.5em]
    \includegraphics[width=\imgwidth]{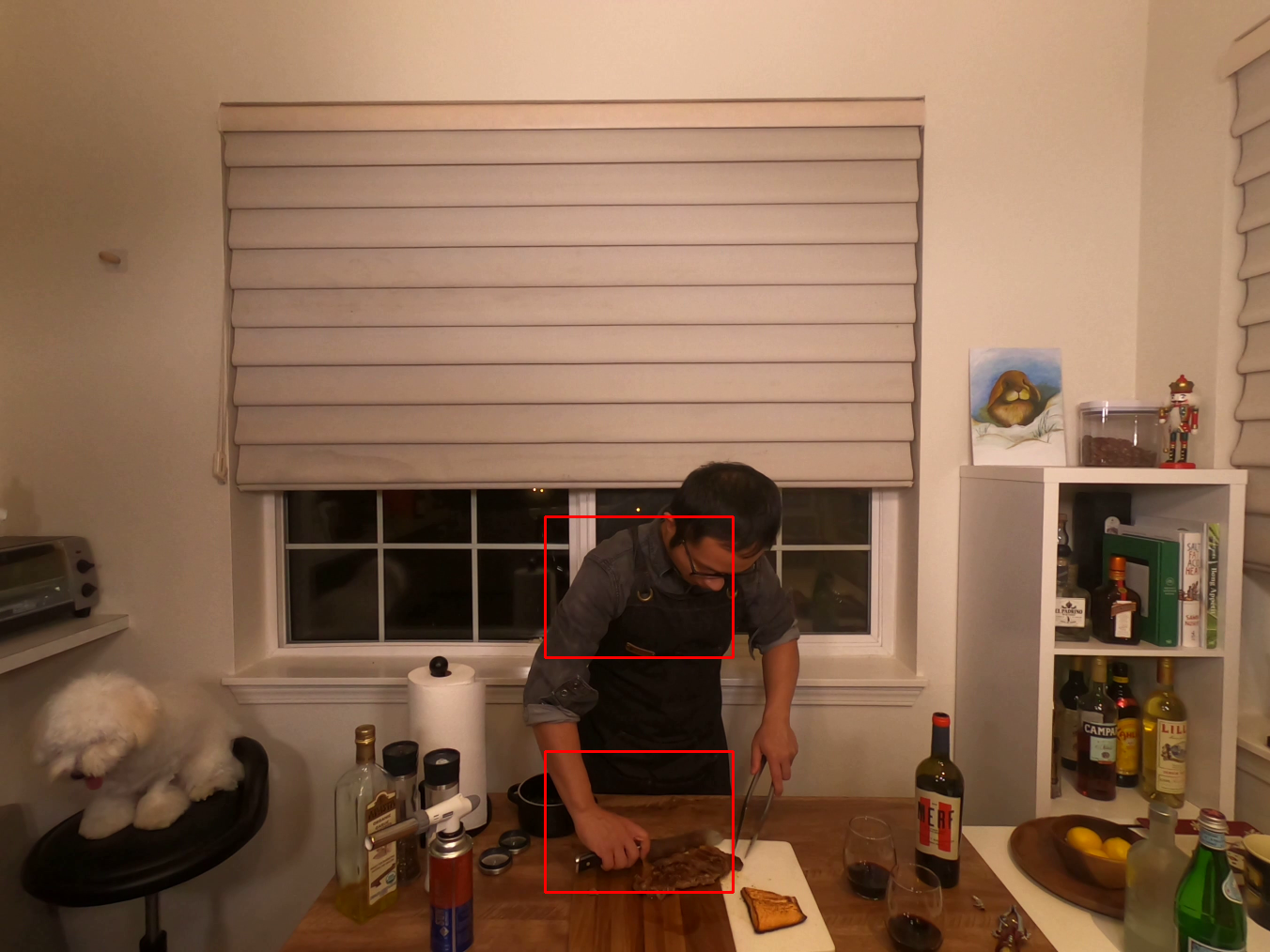}
    \includegraphics[width=\imgwidth]{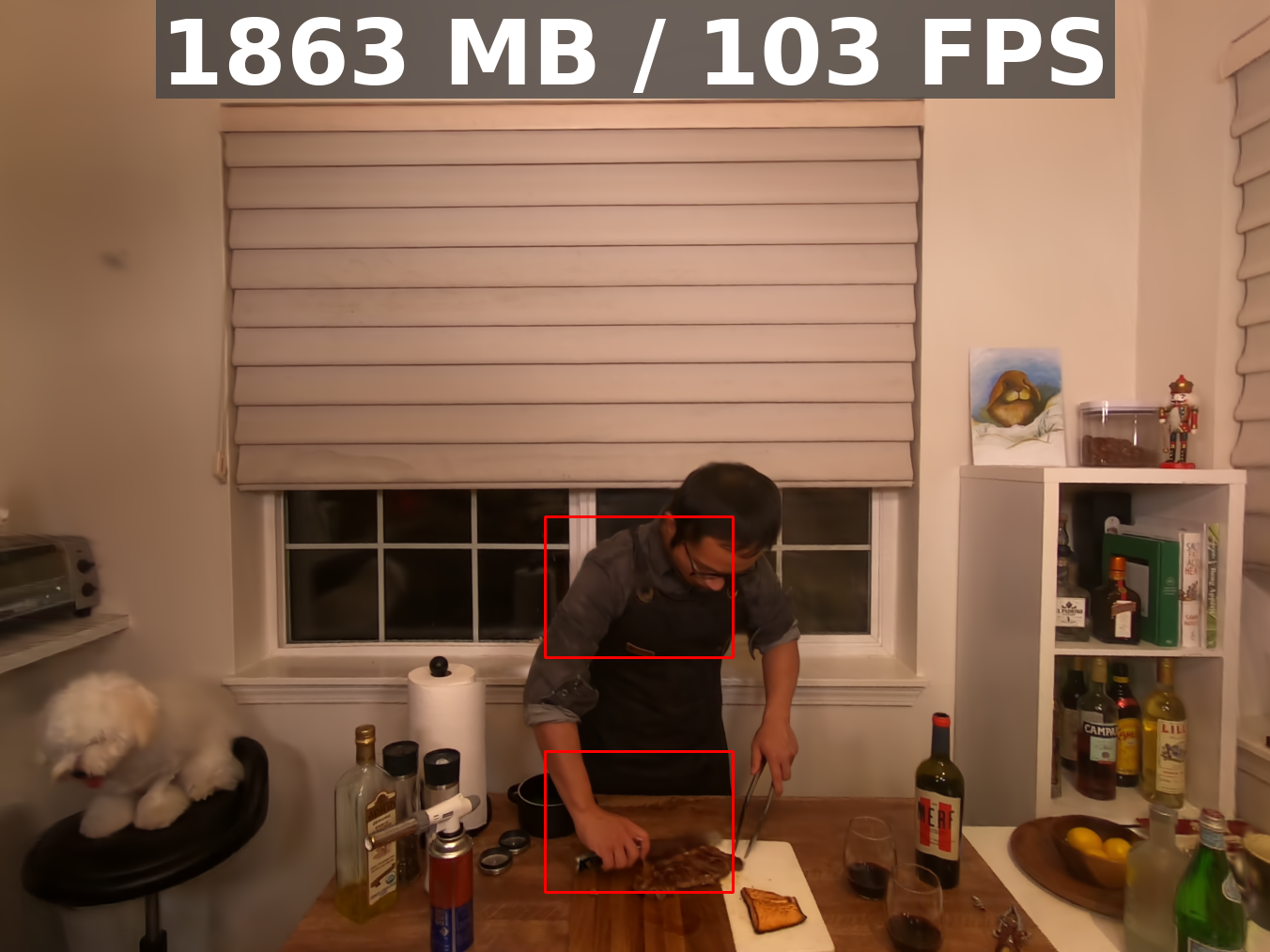}
    \includegraphics[width=\imgwidth]{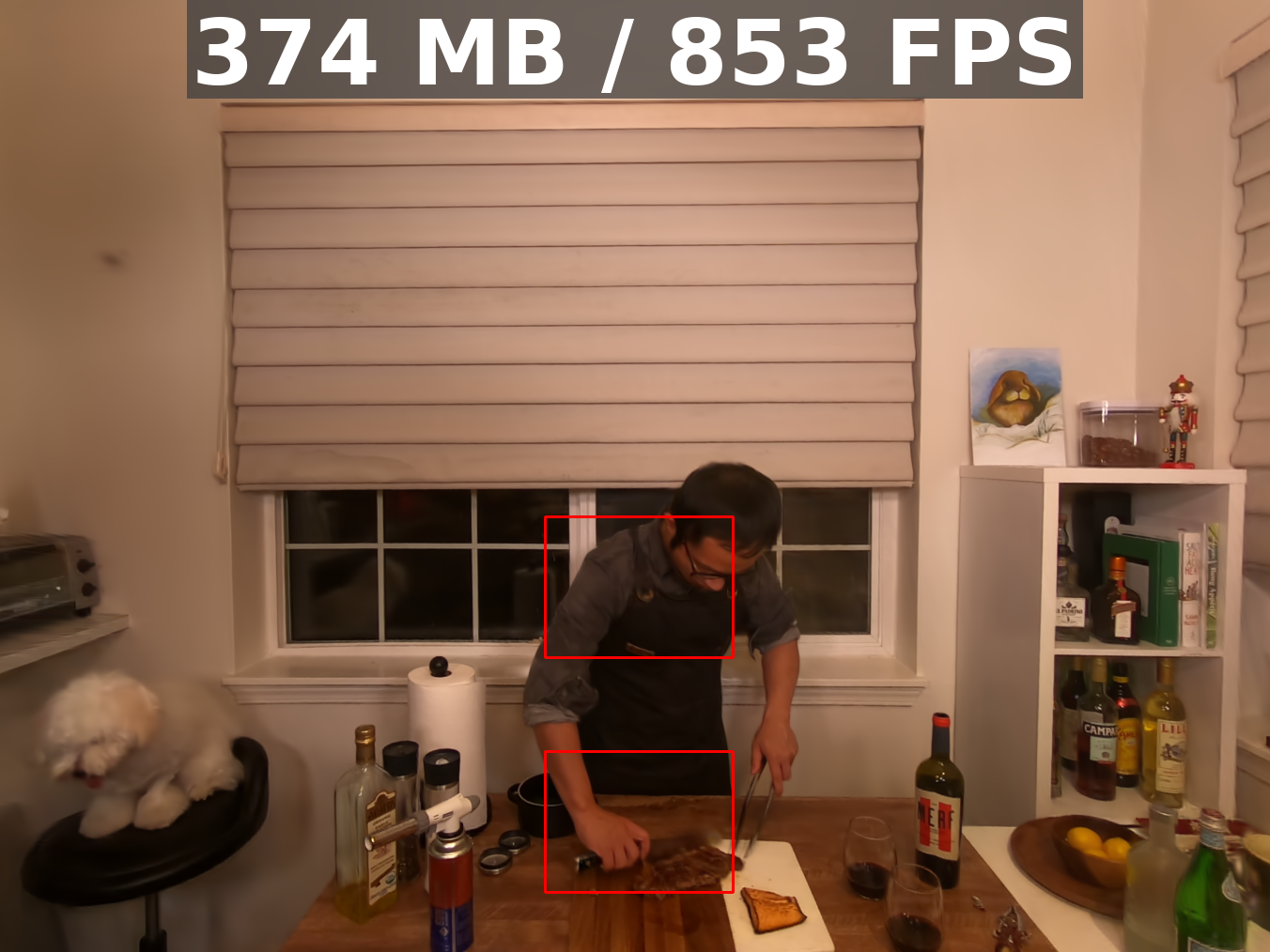}
    \includegraphics[width=\imgwidth]{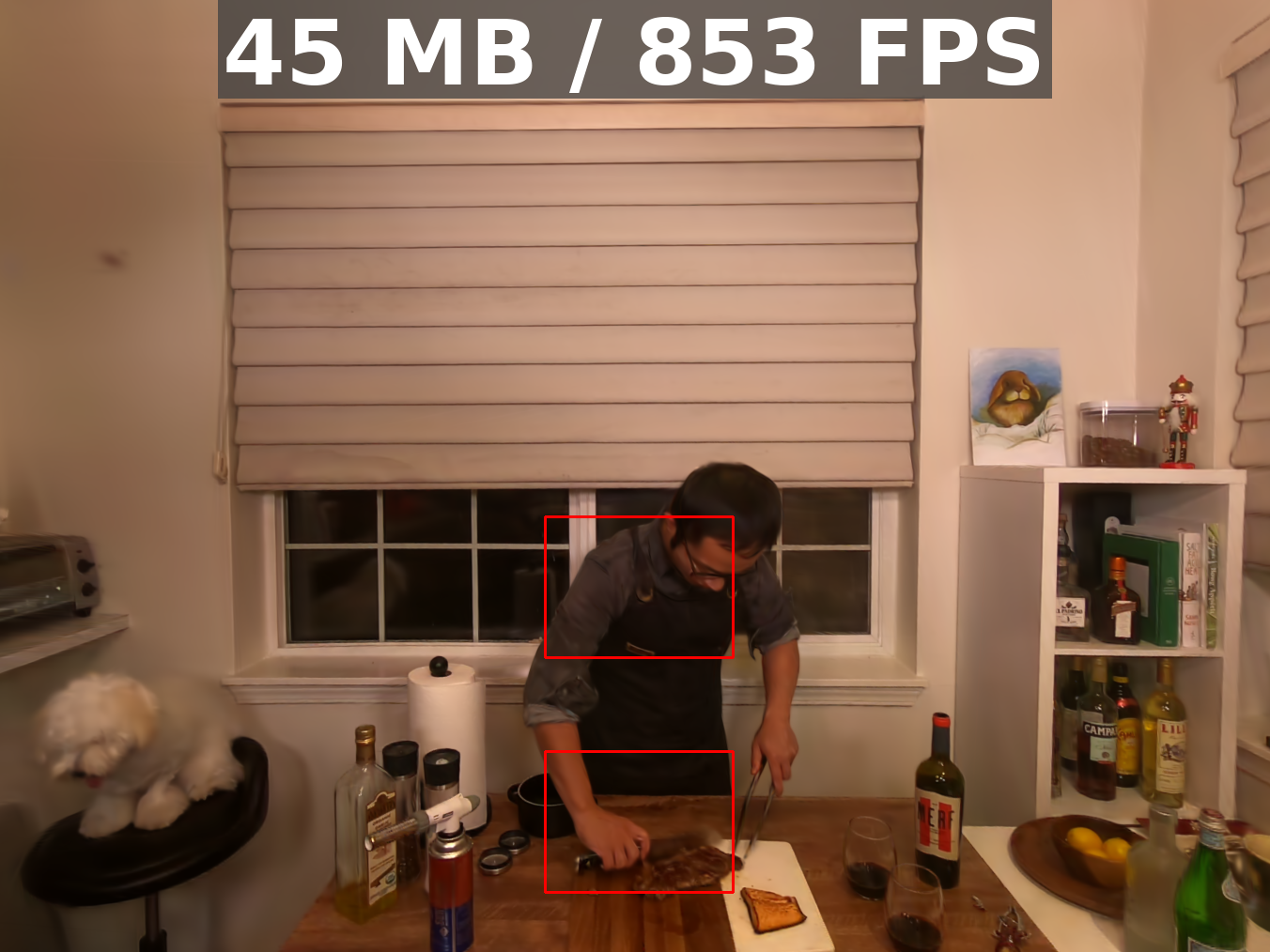}
    \\[2pt]
    
    \makebox[\imgwidth]{%
        \includegraphics[width=\patchwidth]{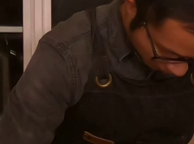}\hspace{\hwidth}
        \includegraphics[width=\patchwidth]{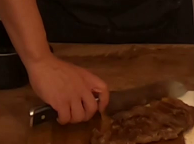}} 
    \makebox[\imgwidth]{%
        \includegraphics[width=\patchwidth]{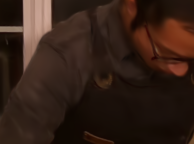}\hspace{\hwidth}
        \includegraphics[width=\patchwidth]{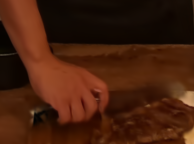}}
    \makebox[\imgwidth]{%
        \includegraphics[width=\patchwidth]{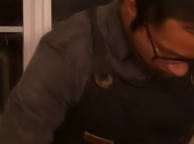}\hspace{\hwidth}
        \includegraphics[width=\patchwidth]{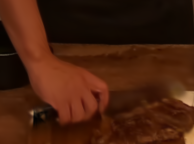}}
    \makebox[\imgwidth]{%
        \includegraphics[width=\patchwidth]{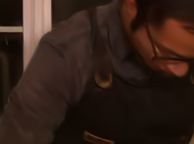}\hspace{\hwidth}
        \includegraphics[width=\patchwidth]{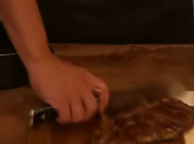}}
    \\[-0.1em]
    \makebox[\textwidth]{\small (c) Results on Cut Roasted Beef Scene.} 
    \\[0.5em]
    
    \caption{\textbf{Qualitative comparisons of 4DGS and our method on the N3V dataset.} To be continued in the next page.} 
    \label{fig:pern3v1}
\end{figure*}

\begin{figure*}[t] \centering
    \newcommand{\hwidth}{1pt}
    \newcommand{\imgwidth}{0.24\textwidth}
    \newcommand{\patchwidth}{0.115\textwidth}
    \makebox[\imgwidth]{\small Ground Truth}
    \makebox[\imgwidth]{\small 4DGS}
    \makebox[\imgwidth]{\small Ours} 
    \makebox[\imgwidth]{\small Ours-PP} 
    \includegraphics[width=\imgwidth]{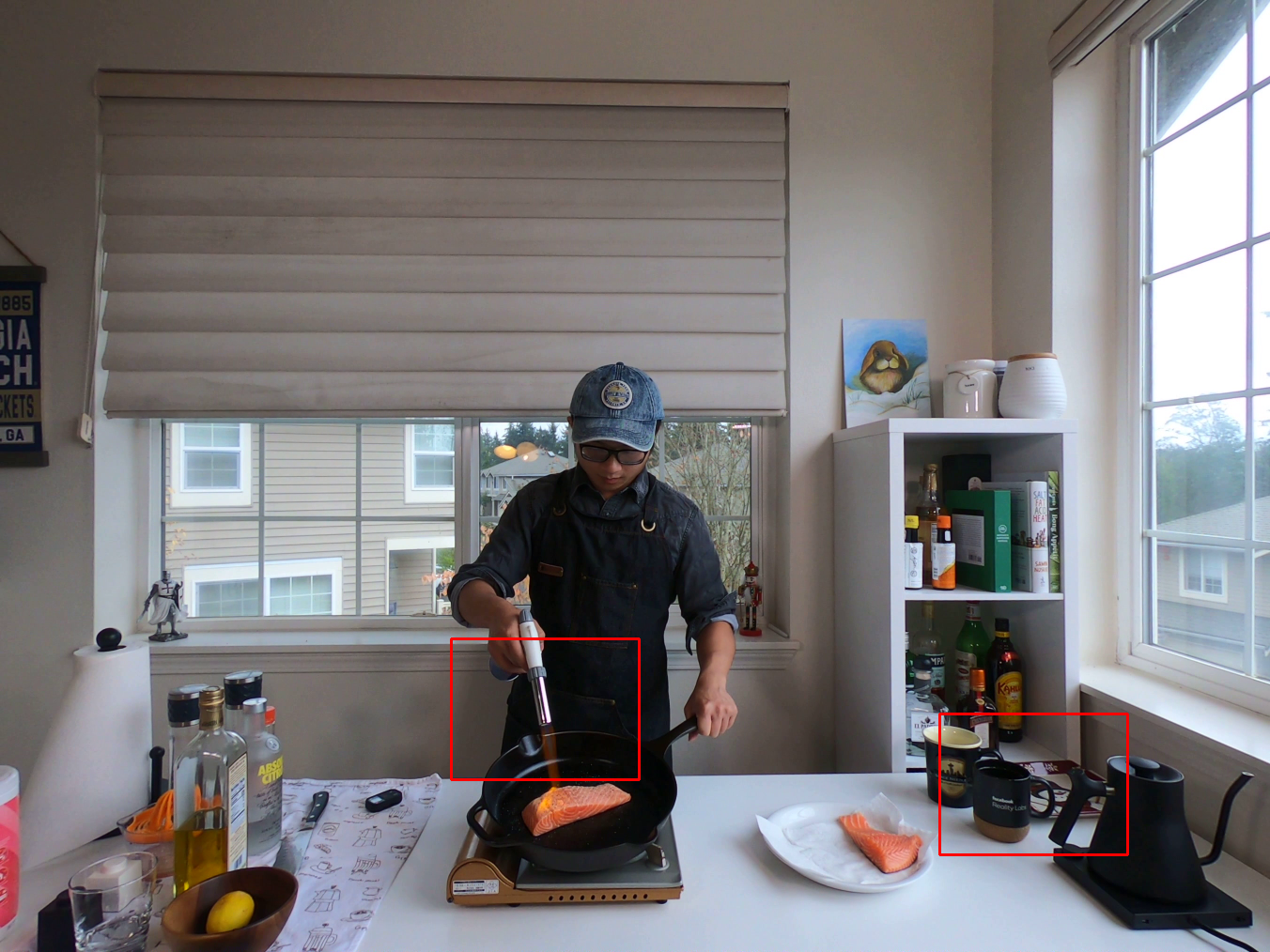}
    \includegraphics[width=\imgwidth]{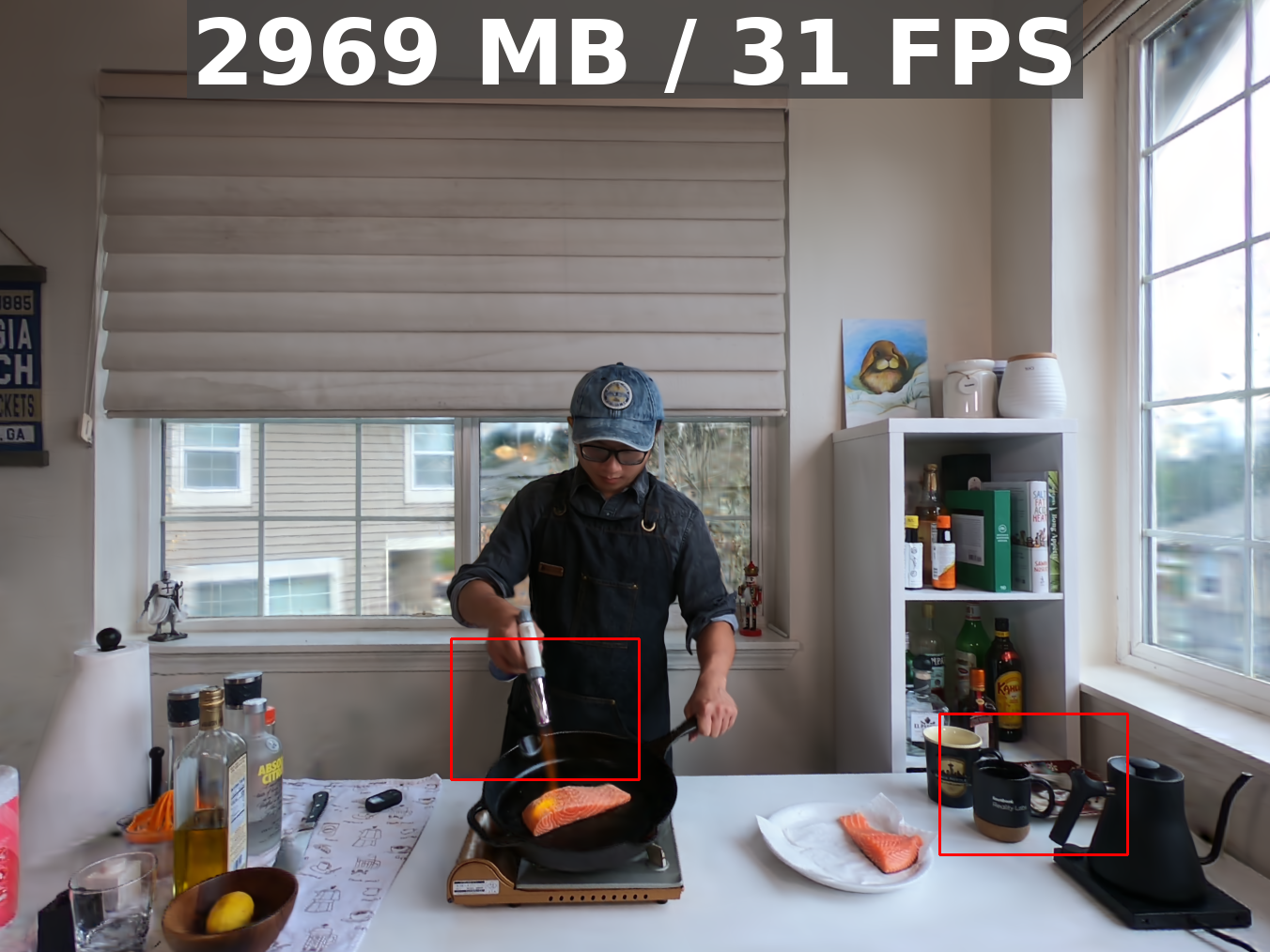}
    \includegraphics[width=\imgwidth]{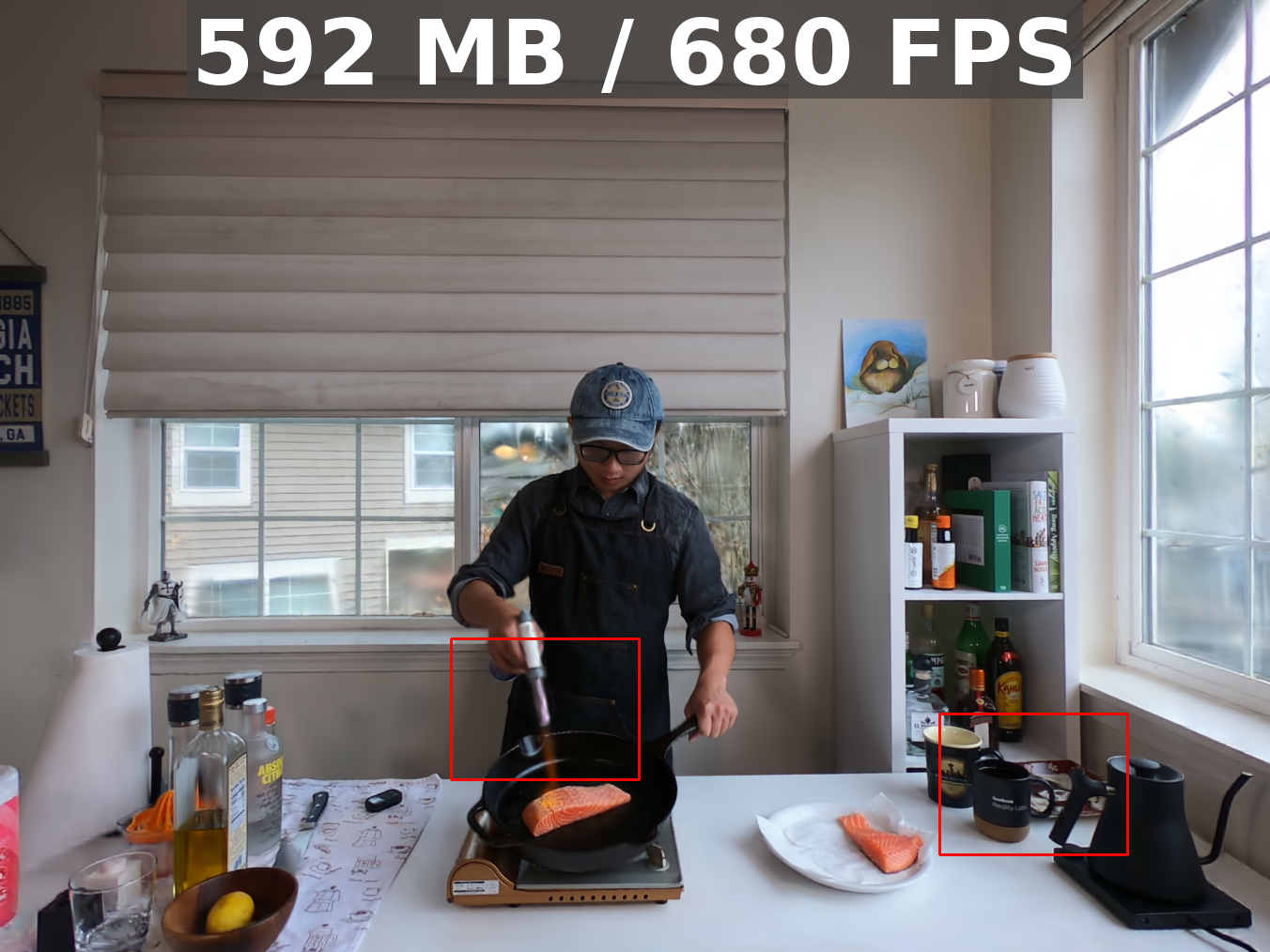}
    \includegraphics[width=\imgwidth]{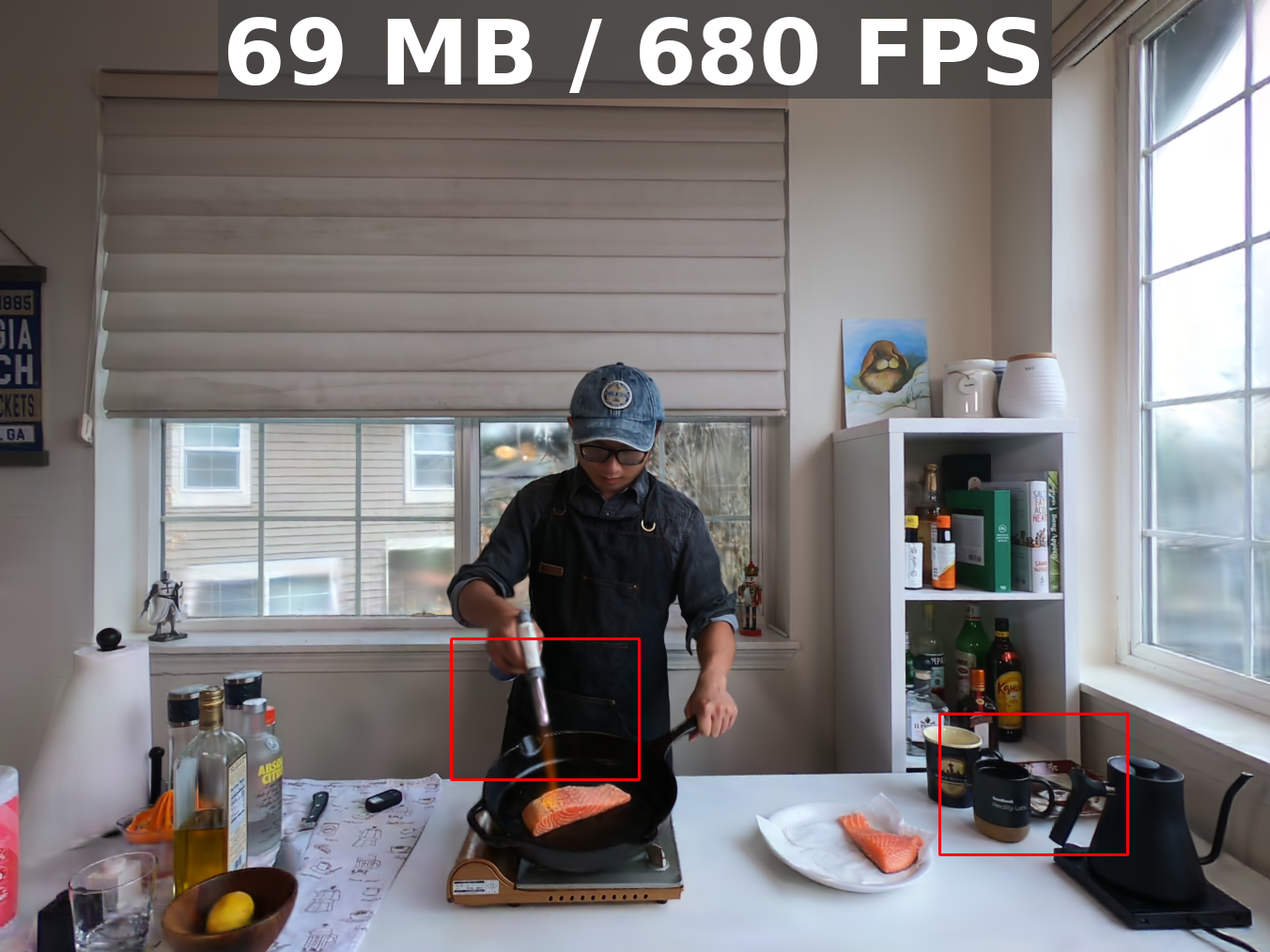}
    \\[2pt]
    
    \makebox[\imgwidth]{%
        \includegraphics[width=\patchwidth]{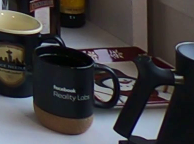}\hspace{\hwidth}
        \includegraphics[width=\patchwidth]{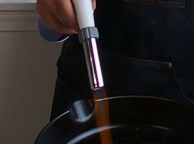}} 
    \makebox[\imgwidth]{%
        \includegraphics[width=\patchwidth]{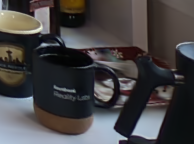}\hspace{\hwidth}
        \includegraphics[width=\patchwidth]{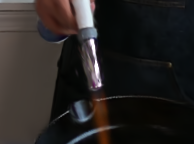}}
    \makebox[\imgwidth]{%
        \includegraphics[width=\patchwidth]{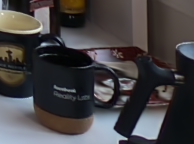}\hspace{\hwidth}
        \includegraphics[width=\patchwidth]{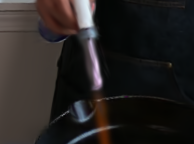}}
    \makebox[\imgwidth]{%
        \includegraphics[width=\patchwidth]{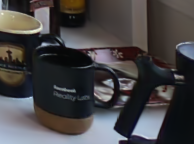}\hspace{\hwidth}
        \includegraphics[width=\patchwidth]{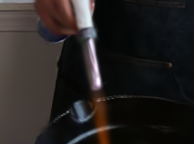}}
    \\[-0.1em]
    \makebox[\textwidth]{\small (a) Results on Flame Salmon Scene.} 
    \\[0.5em]
    \includegraphics[width=\imgwidth]{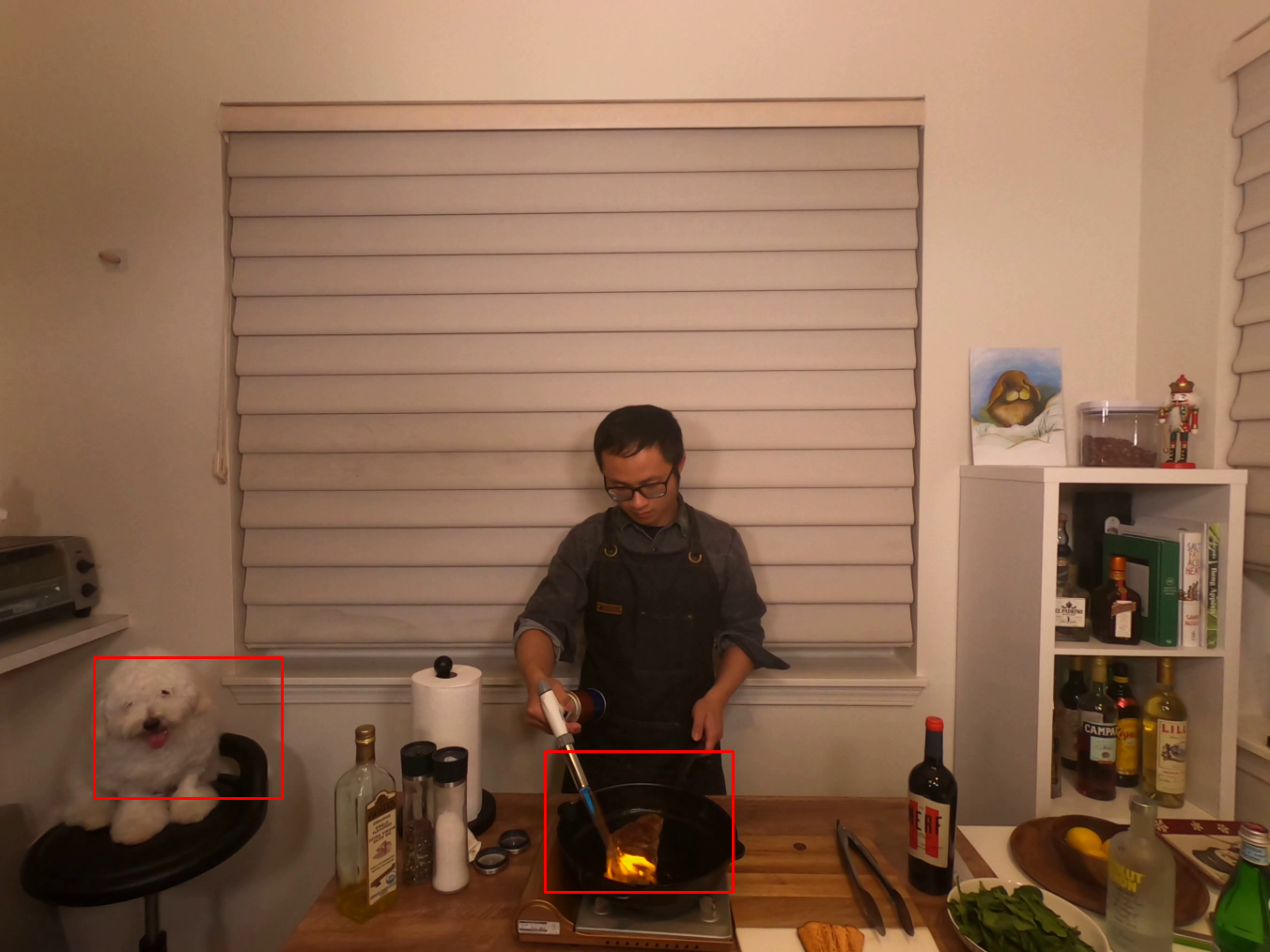}
    \includegraphics[width=\imgwidth]{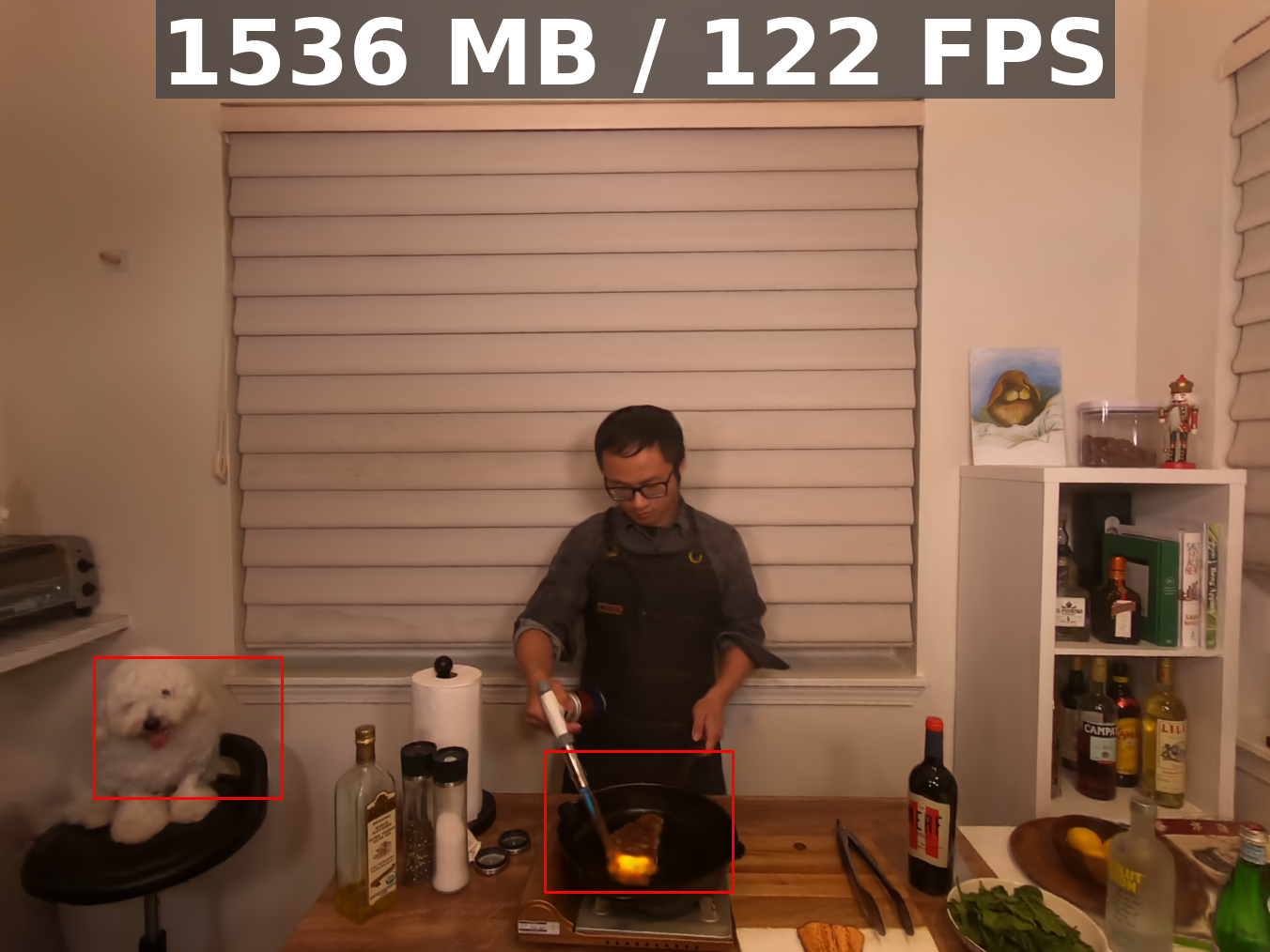}
    \includegraphics[width=\imgwidth]{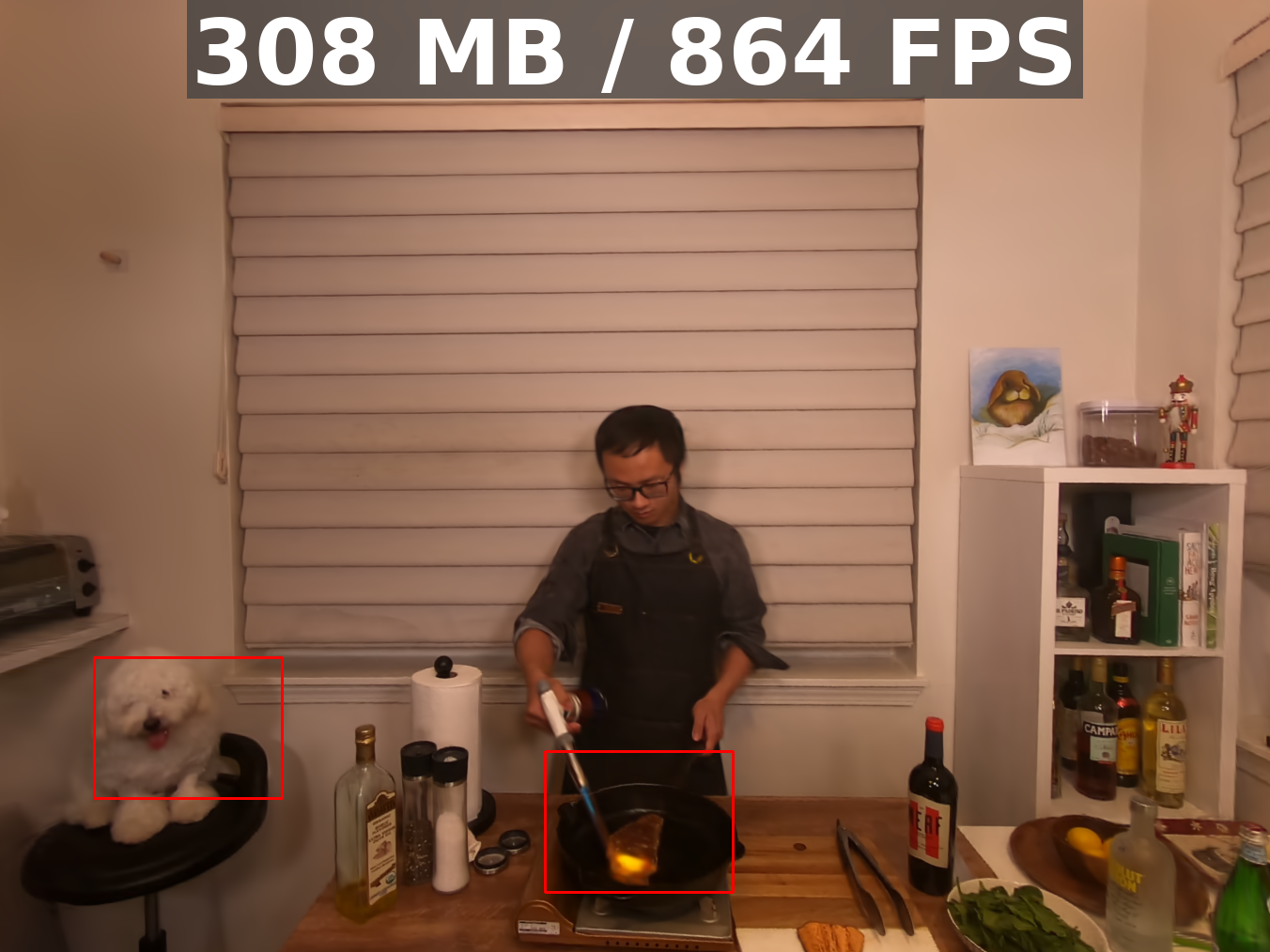}
    \includegraphics[width=\imgwidth]{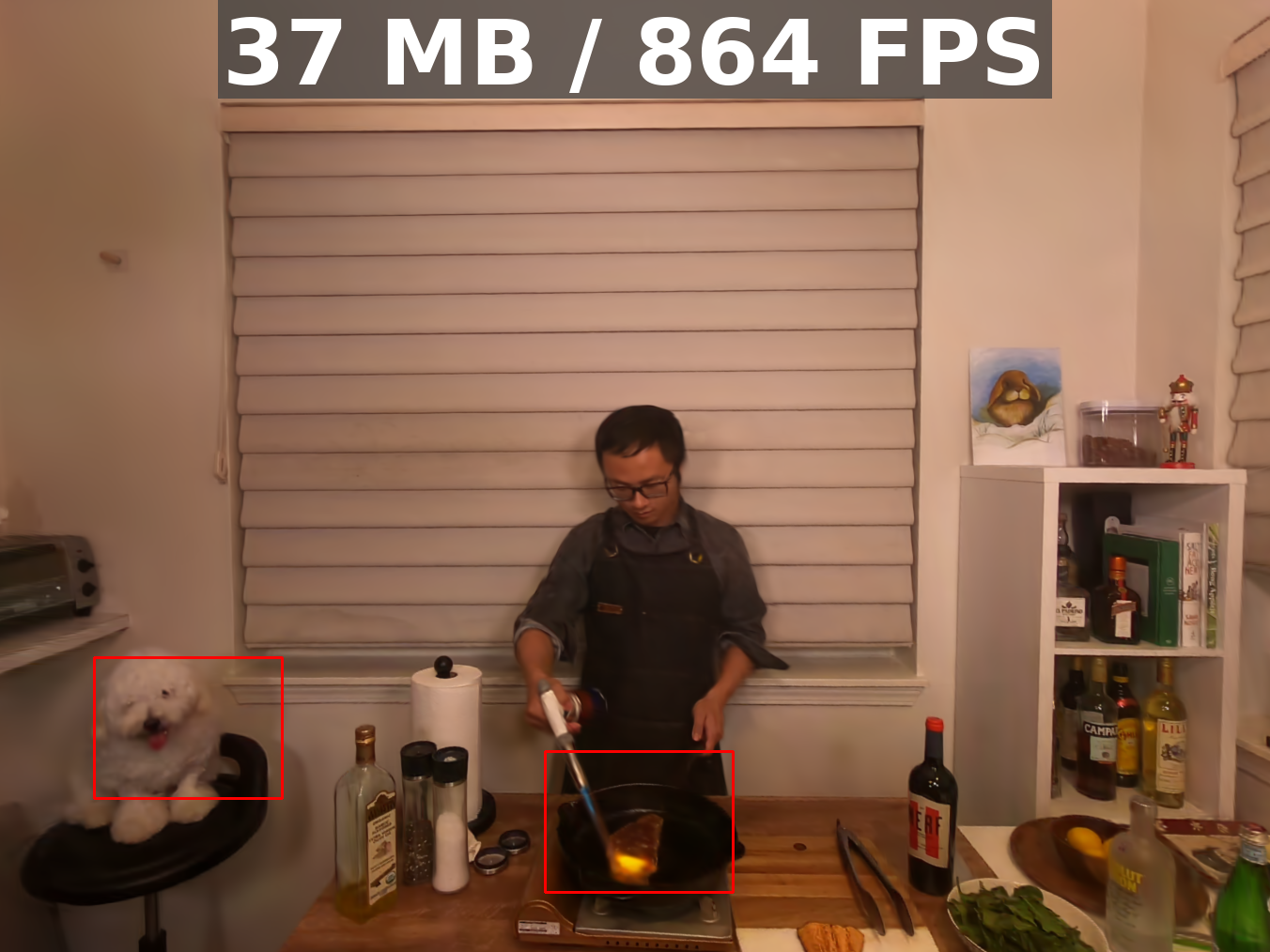}
    \\[2pt]
    
    \makebox[\imgwidth]{%
        \includegraphics[width=\patchwidth]{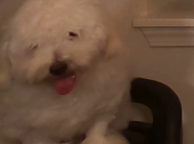}\hspace{\hwidth}
        \includegraphics[width=\patchwidth]{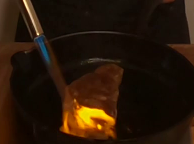}} 
    \makebox[\imgwidth]{%
        \includegraphics[width=\patchwidth]{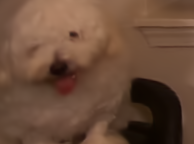}\hspace{\hwidth}
        \includegraphics[width=\patchwidth]{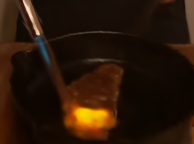}}
    \makebox[\imgwidth]{%
        \includegraphics[width=\patchwidth]{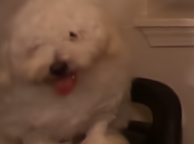}\hspace{\hwidth}
        \includegraphics[width=\patchwidth]{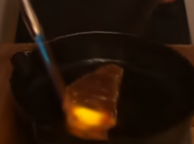}}
    \makebox[\imgwidth]{%
        \includegraphics[width=\patchwidth]{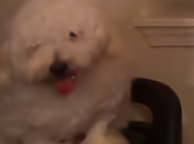}\hspace{\hwidth}
        \includegraphics[width=\patchwidth]{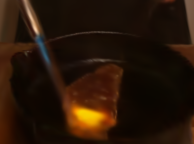}}
    \\[-0.1em]
    \makebox[\textwidth]{\small (b) Results on Flame Steak Scene.} 
    \\[0.5em]
    \includegraphics[width=\imgwidth]{Fig/Qualitative_result/sear_steak/boxed_image_gt.png}
    \includegraphics[width=\imgwidth]{Fig/Qualitative_result/sear_steak/boxed_image_vanilla_label.png}
    \includegraphics[width=\imgwidth]{Fig/Qualitative_result/sear_steak/boxed_image_filter_label.png}
    \includegraphics[width=\imgwidth]{Fig/Qualitative_result/sear_steak/boxed_image_kmeans_label.png}
    \\[2pt]
    
    \makebox[\imgwidth]{%
        \includegraphics[width=\patchwidth]{Fig/Qualitative_result/sear_steak/cropped_image_gt_1.png}\hspace{\hwidth}
        \includegraphics[width=\patchwidth]{Fig/Qualitative_result/sear_steak/cropped_image_gt_2.png}} 
    \makebox[\imgwidth]{%
        \includegraphics[width=\patchwidth]{Fig/Qualitative_result/sear_steak/cropped_image_vanilla_1.png}\hspace{\hwidth}
        \includegraphics[width=\patchwidth]{Fig/Qualitative_result/sear_steak/cropped_image_vanilla_2.png}}
    \makebox[\imgwidth]{%
        \includegraphics[width=\patchwidth]{Fig/Qualitative_result/sear_steak/cropped_image_filter_1.png}\hspace{\hwidth}
        \includegraphics[width=\patchwidth]{Fig/Qualitative_result/sear_steak/cropped_image_filter_2.png}}
    \makebox[\imgwidth]{%
        \includegraphics[width=\patchwidth]{Fig/Qualitative_result/sear_steak/cropped_image_kmeans_1.png}\hspace{\hwidth}
        \includegraphics[width=\patchwidth]{Fig/Qualitative_result/sear_steak/cropped_image_kmeans_2.png}}
    \\[-0.1em]
    \makebox[\textwidth]{\small (c) Results on Sear Steak Scene.} 
    \\[0.5em]
    
    \caption{\textbf{Qualitative comparisons of 4DGS and our method on the N3V dataset.} } 
    \label{fig:pern3v2}
\end{figure*}

\begin{figure*}[t] \centering
    \newcommand{\hwidth}{1pt}
    \newcommand{\imgwidth}{0.24\textwidth}
    \newcommand{\patchwidth}{0.115\textwidth}
    \makebox[\imgwidth]{\small Ground Truth}
    \makebox[\imgwidth]{\small 4DGS}
    \makebox[\imgwidth]{\small Ours} 
    \makebox[\imgwidth]{\small Ours-PP} 
    \includegraphics[width=\imgwidth]{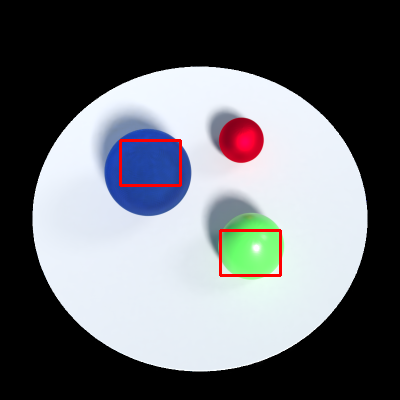}
    \includegraphics[width=\imgwidth]{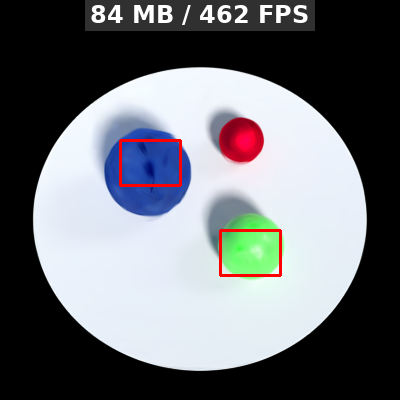}
    \includegraphics[width=\imgwidth]{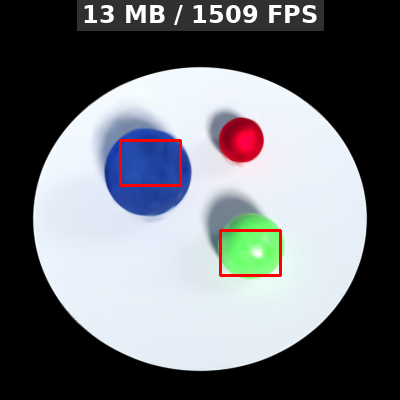}
    \includegraphics[width=\imgwidth]{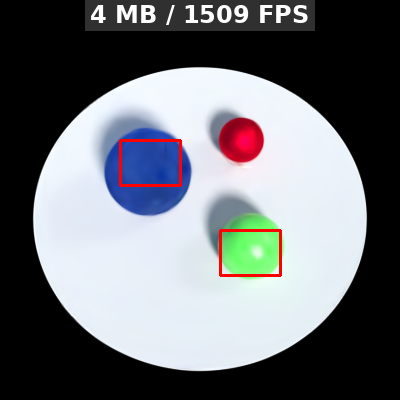}
    \\[2pt]
    
    \makebox[\imgwidth]{%
        \includegraphics[width=\patchwidth]{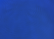}\hspace{\hwidth}
        \includegraphics[width=\patchwidth]{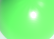}} 
    \makebox[\imgwidth]{%
        \includegraphics[width=\patchwidth]{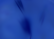}\hspace{\hwidth}
        \includegraphics[width=\patchwidth]{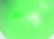}}
    \makebox[\imgwidth]{%
        \includegraphics[width=\patchwidth]{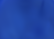}\hspace{\hwidth}
        \includegraphics[width=\patchwidth]{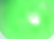}}
    \makebox[\imgwidth]{%
        \includegraphics[width=\patchwidth]{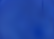}\hspace{\hwidth}
        \includegraphics[width=\patchwidth]{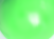}}
    \\[-0.1em]
    \makebox[\textwidth]{\small (a) Results on Bouncingballs Scene.} 
    \\[0.5em]
    \includegraphics[width=\imgwidth]{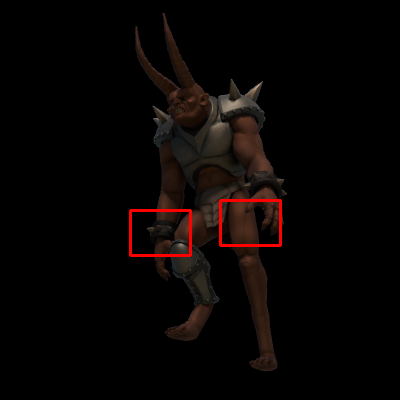}
    \includegraphics[width=\imgwidth]{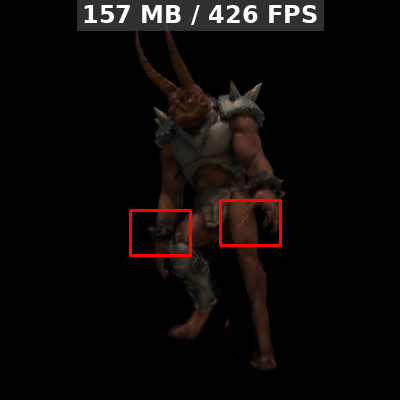}
    \includegraphics[width=\imgwidth]{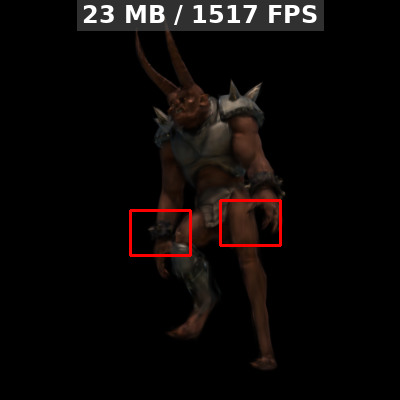}
    \includegraphics[width=\imgwidth]{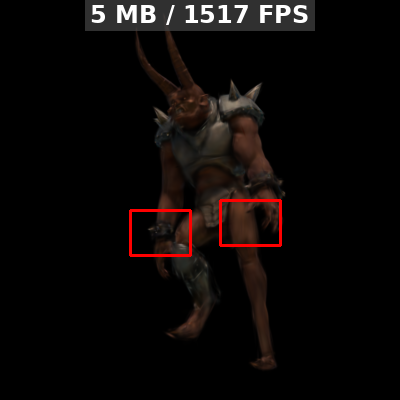}
    \\[2pt]
    
    \makebox[\imgwidth]{%
        \includegraphics[width=\patchwidth]{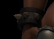}\hspace{\hwidth}
        \includegraphics[width=\patchwidth]{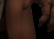}} 
    \makebox[\imgwidth]{%
        \includegraphics[width=\patchwidth]{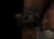}\hspace{\hwidth}
        \includegraphics[width=\patchwidth]{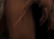}}
    \makebox[\imgwidth]{%
        \includegraphics[width=\patchwidth]{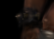}\hspace{\hwidth}
        \includegraphics[width=\patchwidth]{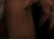}}
    \makebox[\imgwidth]{%
        \includegraphics[width=\patchwidth]{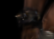}\hspace{\hwidth}
        \includegraphics[width=\patchwidth]{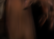}}
    \\[-0.1em]
    \makebox[\textwidth]{\small (b) Results on Hellwarrior Scene.} 
    \\[0.5em]
    \includegraphics[width=\imgwidth]{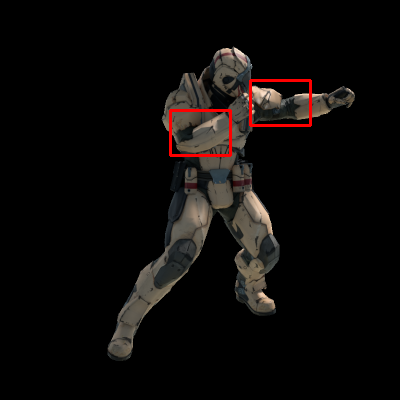}
    \includegraphics[width=\imgwidth]{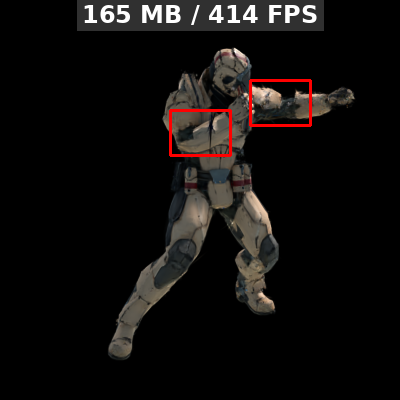}
    \includegraphics[width=\imgwidth]{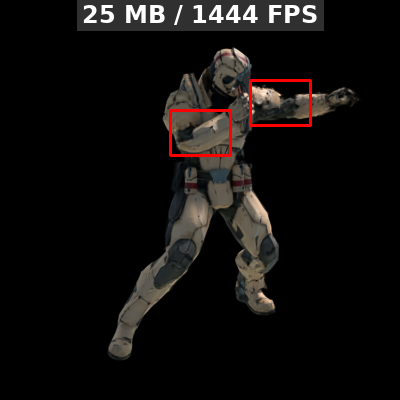}
    \includegraphics[width=\imgwidth]{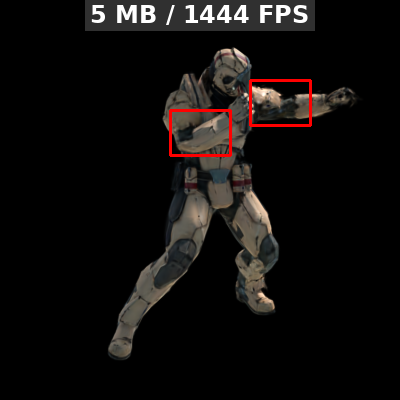}
    \\[2pt]
    
    \makebox[\imgwidth]{%
        \includegraphics[width=\patchwidth]{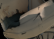}\hspace{\hwidth}
        \includegraphics[width=\patchwidth]{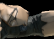}} 
    \makebox[\imgwidth]{%
        \includegraphics[width=\patchwidth]{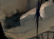}\hspace{\hwidth}
        \includegraphics[width=\patchwidth]{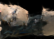}}
    \makebox[\imgwidth]{%
        \includegraphics[width=\patchwidth]{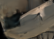}\hspace{\hwidth}
        \includegraphics[width=\patchwidth]{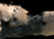}}
    \makebox[\imgwidth]{%
        \includegraphics[width=\patchwidth]{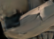}\hspace{\hwidth}
        \includegraphics[width=\patchwidth]{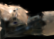}}
    \\[-0.1em]
    \makebox[\textwidth]{\small (c) Results on Hook Scene.} 
    \\[0.5em]
    
    \caption{\textbf{Qualitative comparisons of 4DGS and our method on the D-nerf dataset.} To be continued in the next page.} 
    \label{fig:perdnerf1}
\end{figure*}

\begin{figure*}[t] \centering
    \newcommand{\hwidth}{1pt}
    \newcommand{\imgwidth}{0.24\textwidth}
    \newcommand{\patchwidth}{0.115\textwidth}
    \makebox[\imgwidth]{\small Ground Truth}
    \makebox[\imgwidth]{\small 4DGS}
    \makebox[\imgwidth]{\small Ours} 
    \makebox[\imgwidth]{\small Ours-PP} 
    \includegraphics[width=\imgwidth]{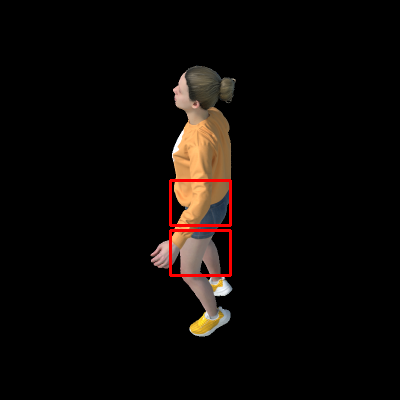}
    \includegraphics[width=\imgwidth]{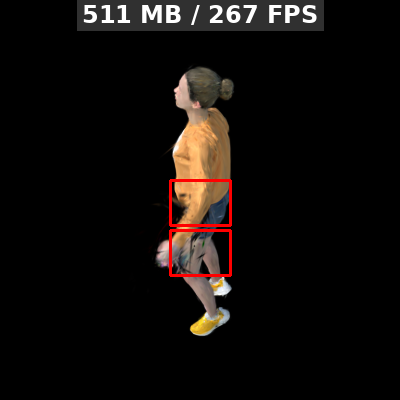}
    \includegraphics[width=\imgwidth]{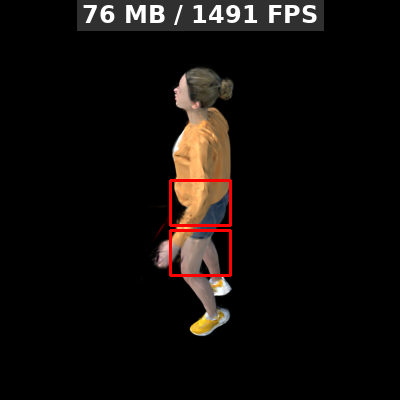}
    \includegraphics[width=\imgwidth]{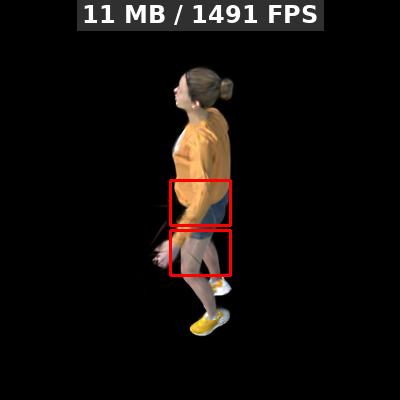}
    \\[2pt]
    
    \makebox[\imgwidth]{%
        \includegraphics[width=\patchwidth]{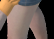}\hspace{\hwidth}
        \includegraphics[width=\patchwidth]{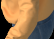}} 
    \makebox[\imgwidth]{%
        \includegraphics[width=\patchwidth]{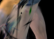}\hspace{\hwidth}
        \includegraphics[width=\patchwidth]{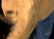}}
    \makebox[\imgwidth]{%
        \includegraphics[width=\patchwidth]{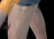}\hspace{\hwidth}
        \includegraphics[width=\patchwidth]{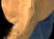}}
    \makebox[\imgwidth]{%
        \includegraphics[width=\patchwidth]{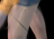}\hspace{\hwidth}
        \includegraphics[width=\patchwidth]{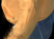}}
    \\[-0.1em]
    \makebox[\textwidth]{\small (a) Results on Jumpingjacks Scene.} 
    \\[0.5em]
    \includegraphics[width=\imgwidth]{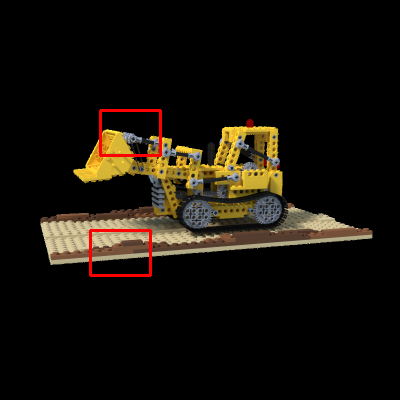}
    \includegraphics[width=\imgwidth]{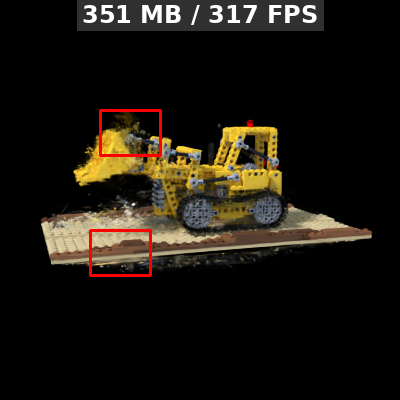}
    \includegraphics[width=\imgwidth]{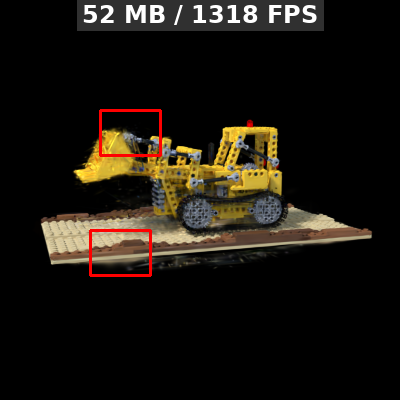}
    \includegraphics[width=\imgwidth]{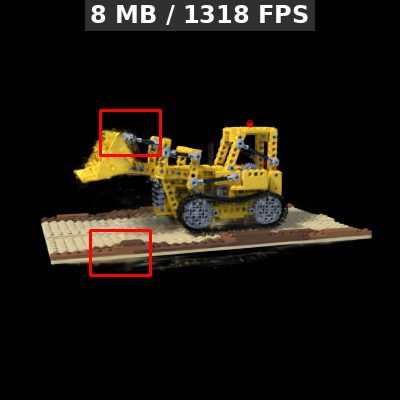}
    \\[2pt]
    
    \makebox[\imgwidth]{%
        \includegraphics[width=\patchwidth]{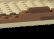}\hspace{\hwidth}
        \includegraphics[width=\patchwidth]{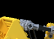}} 
    \makebox[\imgwidth]{%
        \includegraphics[width=\patchwidth]{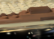}\hspace{\hwidth}
        \includegraphics[width=\patchwidth]{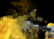}}
    \makebox[\imgwidth]{%
        \includegraphics[width=\patchwidth]{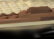}\hspace{\hwidth}
        \includegraphics[width=\patchwidth]{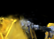}}
    \makebox[\imgwidth]{%
        \includegraphics[width=\patchwidth]{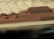}\hspace{\hwidth}
        \includegraphics[width=\patchwidth]{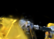}}
    \\[-0.1em]
    \makebox[\textwidth]{\small (b) Results on Lego Scene.} 
    \\[0.5em]
    \includegraphics[width=\imgwidth]{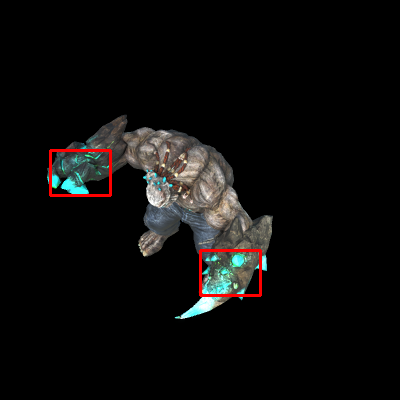}
    \includegraphics[width=\imgwidth]{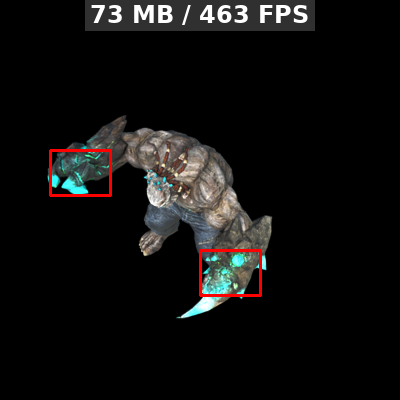}
    \includegraphics[width=\imgwidth]{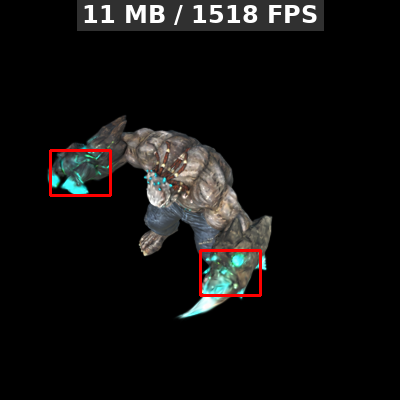}
    \includegraphics[width=\imgwidth]{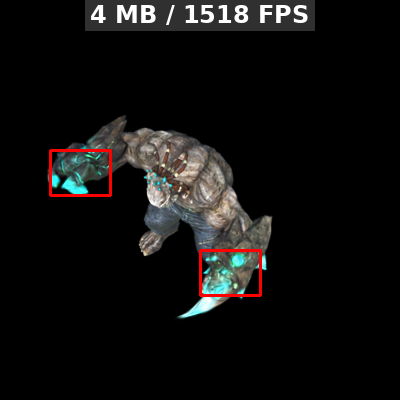}
    \\[2pt]
    
    \makebox[\imgwidth]{%
        \includegraphics[width=\patchwidth]{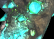}\hspace{\hwidth}
        \includegraphics[width=\patchwidth]{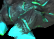}} 
    \makebox[\imgwidth]{%
        \includegraphics[width=\patchwidth]{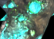}\hspace{\hwidth}
        \includegraphics[width=\patchwidth]{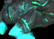}}
    \makebox[\imgwidth]{%
        \includegraphics[width=\patchwidth]{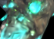}\hspace{\hwidth}
        \includegraphics[width=\patchwidth]{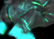}}
    \makebox[\imgwidth]{%
        \includegraphics[width=\patchwidth]{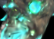}\hspace{\hwidth}
        \includegraphics[width=\patchwidth]{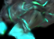}}
    \\[-0.1em]
    \makebox[\textwidth]{\small (c) Results on Mutant Scene.} 
    \\[0.5em]
    
    \caption{\textbf{Qualitative comparisons of 4DGS and our method on the D-nerf dataset.} To be continued in the next page.} 
    \label{fig:perdnerf2}
\end{figure*}

\begin{figure*}[t] \centering
    \newcommand{\hwidth}{1pt}
    \newcommand{\imgwidth}{0.24\textwidth}
    \newcommand{\patchwidth}{0.115\textwidth}
    \makebox[\imgwidth]{\small Ground Truth}
    \makebox[\imgwidth]{\small 4DGS}
    \makebox[\imgwidth]{\small Ours} 
    \makebox[\imgwidth]{\small Ours-PP} 
    \includegraphics[width=\imgwidth]{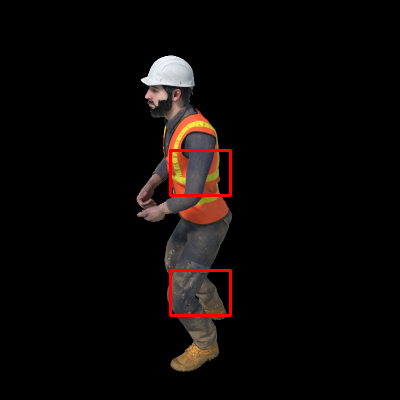}
    \includegraphics[width=\imgwidth]{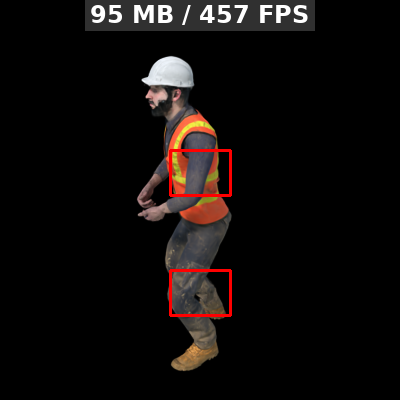}
    \includegraphics[width=\imgwidth]{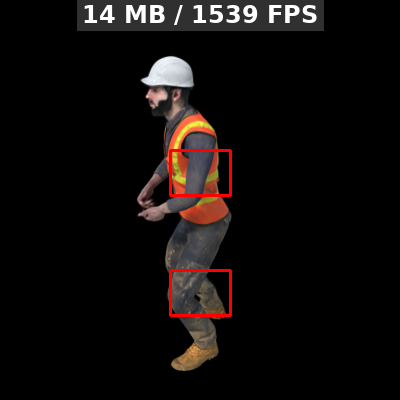}
    \includegraphics[width=\imgwidth]{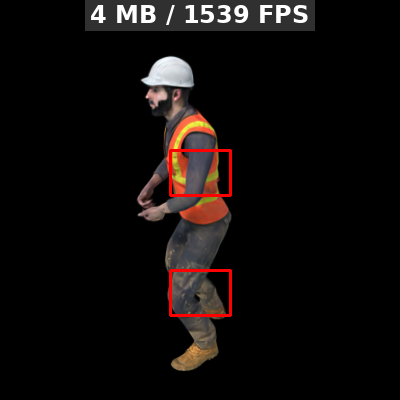}
    \\[2pt]
    
    \makebox[\imgwidth]{%
        \includegraphics[width=\patchwidth]{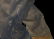}\hspace{\hwidth}
        \includegraphics[width=\patchwidth]{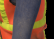}} 
    \makebox[\imgwidth]{%
        \includegraphics[width=\patchwidth]{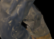}\hspace{\hwidth}
        \includegraphics[width=\patchwidth]{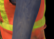}}
    \makebox[\imgwidth]{%
        \includegraphics[width=\patchwidth]{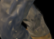}\hspace{\hwidth}
        \includegraphics[width=\patchwidth]{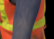}}
    \makebox[\imgwidth]{%
        \includegraphics[width=\patchwidth]{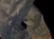}\hspace{\hwidth}
        \includegraphics[width=\patchwidth]{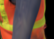}}
    \\[-0.1em]
    \makebox[\textwidth]{\small (a) Results on Standup Scene.} 
    \\[0.5em]
    \includegraphics[width=\imgwidth]{Fig/Qualitative_result/trex/boxed_image_gt.png}
    \includegraphics[width=\imgwidth]{Fig/Qualitative_result/trex/boxed_image_vanilla_label.png}
    \includegraphics[width=\imgwidth]{Fig/Qualitative_result/trex/boxed_image_filter_label.png}
    \includegraphics[width=\imgwidth]{Fig/Qualitative_result/trex/boxed_image_kmeans_label.png}
    \\[2pt]
    
    \makebox[\imgwidth]{%
        \includegraphics[width=\patchwidth]{Fig/Qualitative_result/trex/cropped_image_gt_1.png}\hspace{\hwidth}
        \includegraphics[width=\patchwidth]{Fig/Qualitative_result/trex/cropped_image_gt_2.png}} 
    \makebox[\imgwidth]{%
        \includegraphics[width=\patchwidth]{Fig/Qualitative_result/trex/cropped_image_vanilla_1.png}\hspace{\hwidth}
        \includegraphics[width=\patchwidth]{Fig/Qualitative_result/trex/cropped_image_vanilla_2.png}}
    \makebox[\imgwidth]{%
        \includegraphics[width=\patchwidth]{Fig/Qualitative_result/trex/cropped_image_filter_1.png}\hspace{\hwidth}
        \includegraphics[width=\patchwidth]{Fig/Qualitative_result/trex/cropped_image_filter_2.png}}
    \makebox[\imgwidth]{%
        \includegraphics[width=\patchwidth]{Fig/Qualitative_result/trex/cropped_image_kmeans_1.png}\hspace{\hwidth}
        \includegraphics[width=\patchwidth]{Fig/Qualitative_result/trex/cropped_image_kmeans_2.png}}
    \\[-0.1em]
    \makebox[\textwidth]{\small (b) Results on Trex Scene.} 
    
    \caption{\textbf{Qualitative comparisons of 4DGS and our method on the D-nerf dataset.} } 
    \label{fig:perdnerf3}
\end{figure*}

We provide per-scene quantitative comparisons on the N3V Dataset~\cite{li2022neural}(~\cref{tab:pern3v}) and D-NeRF Dataset~\cite{pumarola2021d}(~\cref{tab:perdnerf}). Compared to vanilla 4DGS~\cite{yang2023real}, our model significantly reduces the storage requirements and enhances rendering speed while maintaining high-quality reconstruction. ~\cref{fig:pern3v1} and ~\cref{fig:pern3v2} show more visual comparisons on the N3V Dataset. ~\cref{fig:perdnerf1}, ~\cref{fig:perdnerf2} and ~\cref{fig:perdnerf3} show visual comparisons on the D-NeRF Dataset.


\subsection{Resource consumption} We present the resource consumption metrics, including training time, GPU memory allocation and additional storage space. On the N3V dataset~\cite{li2022neural}, 4DGS-1K only takes approximately 30 minutes to fine-tune, with GPU memory allocation of 10.54GB. During rendering, it only consumes 1.62GB of GPU memory. For storage requirement, 4DGS-1K requires additional storage for the mask of filter and codebook; however, these occupy only a minimal portion of the total storage, approximately 1 MB per scene. These parts are also included in the final experiment results.

The above results demonstrate the potential of 4DGS-1K for deployment on low-performance hardware. Consequently, we further test 4DGS-1K on TITAN X GPU, where 4DGS-1K maintains 200+ FPS on the N3V dataset, still far outperforming vanilla 4DGS (20 FPS).



\subsection{Additional experiments for redundancy}
In this section, we provide additional experiments for redundancy study as a supplement to \cref{subsection:Delving}. It is composed of two parts: first, the visualization of the Gaussian with short lifespan distribution, and secondly, the relationship between FPS and the number of inactive Gaussians.

\noindent\textbf{Visualization of Gaussians with small lifespan.} 
In \cref{subsection:Delving}, we argue that in vanilla 4DGS, nearly all Gaussians have a
short lifespan, especially around the edge of fast-moving objects. Therefore, we visualize the spatial distribution of $\Sigma_t$ to better support our redundancy study in~\cref{subsection:Delving}.

\begin{figure*}[t] \centering
    \newcommand{\hwidth}{1pt}
    \newcommand{\imgwidth}{0.24\textwidth}
    \newcommand{\patchwidth}{0.115\textwidth}
    \includegraphics[width=\imgwidth]{Fig/temporal_study/00080.png}
    \includegraphics[width=\imgwidth]{Fig/temporal_study/00120.png}
    \includegraphics[width=\imgwidth]{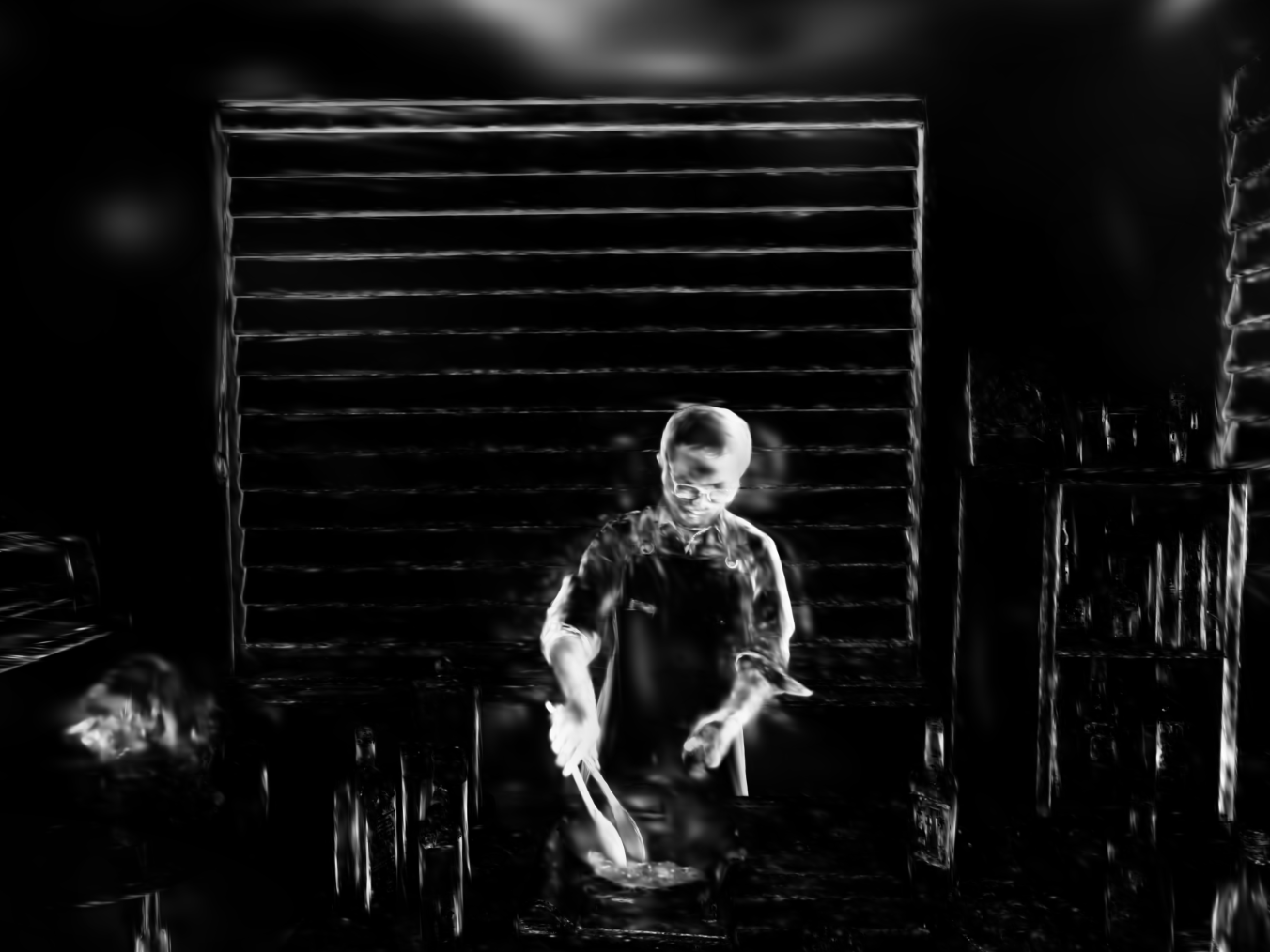}
    \includegraphics[width=\imgwidth]{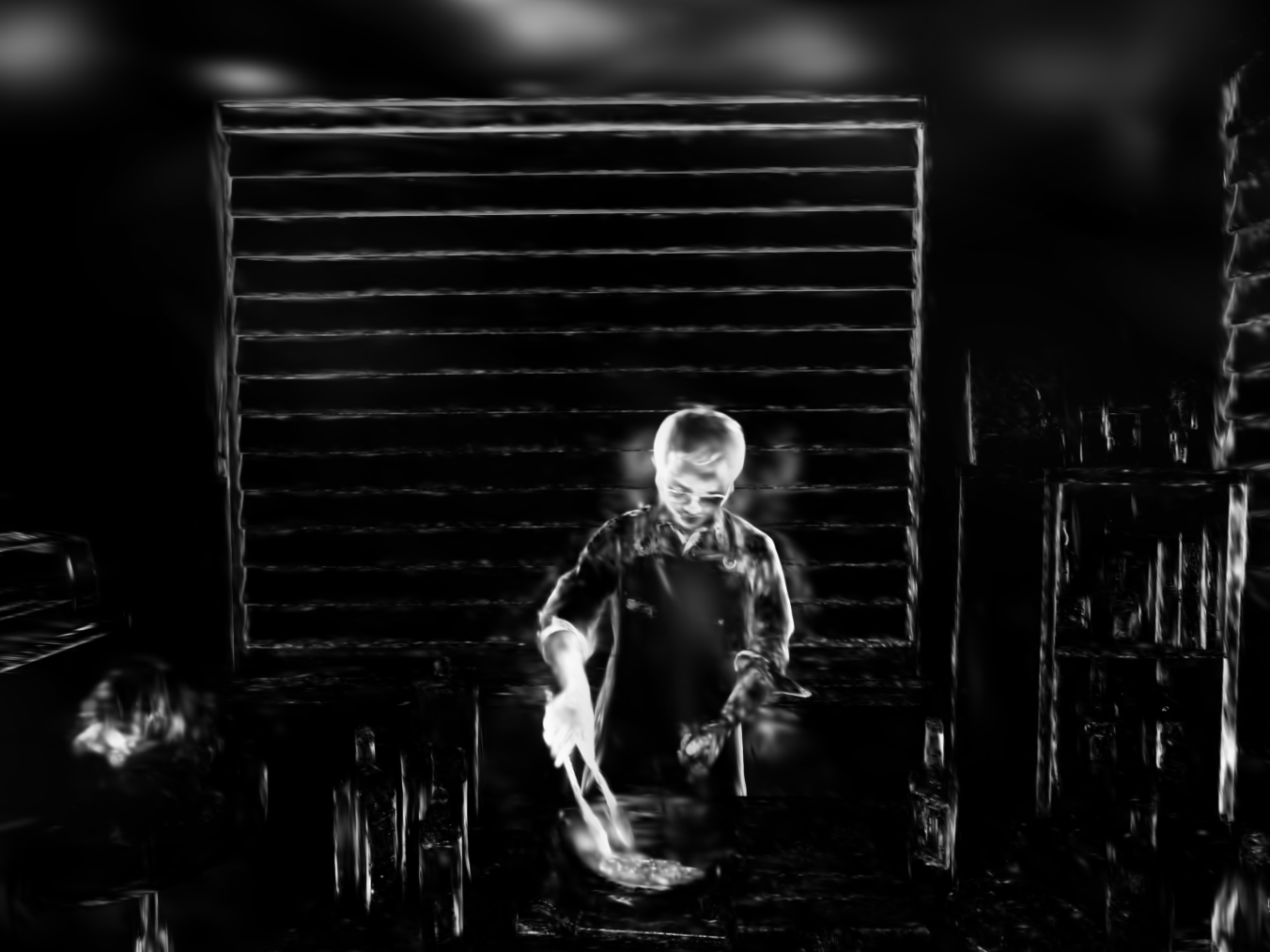}
\caption{\textbf{Visualizations of Distribution of $\Sigma_t$.} }
    \label{fig:temporal_vis}
    
\end{figure*}

Specifically, we visualize the distribution of $\Sigma_t$ at several timestamps in \textit{Sear Steak} Scene. The visualization results are shown in \cref{fig:temporal_vis}. For visualization, we take the reciprocal of $\Sigma_t$ during rendering and then normalize it. Therefore, brighter regions in the rendered image indicate smaller $\Sigma_t$.

As shown in \cref{fig:temporal_vis}, Gaussians with short lifespan are primarily concentrated in regions of object motion, such as the moving person and dog. Moreover, we observe that Gaussians with small $\Sigma_t$ also appear on the edges of some objects which exhibit significant color variation. This is because small Gaussians are preferred in these regions to capture the high-frequency details in the spatial dimension. As vanilla 4DGS~\cite{yang2023real} treats time and space dimensions equally, these Gaussians also have short lifespan in the temporal dimension.

\noindent\textbf{Relationship between FPS and the number of inactive Gaussians.} In \cref{subsection:Delving}, our primary prior assumption is that the number of inactive Gaussians affects the FPS. Therefore, we visualize the relationship between FPS and the number of inactive Gaussians.

However, only limiting the total number of Gaussians is incorrect in this task. As the total number increases, the number of active Gaussians and inactive Gaussians also increases, which cannot clarify whether the FPS variation is caused by active or inactive Gaussians. Consequently, we first identify the active Gaussians by rendering and then add a mount of inactive Gaussians among these Gaussians.

\begin{figure}[thb] \centering
    \includegraphics[width=0.45\textwidth]{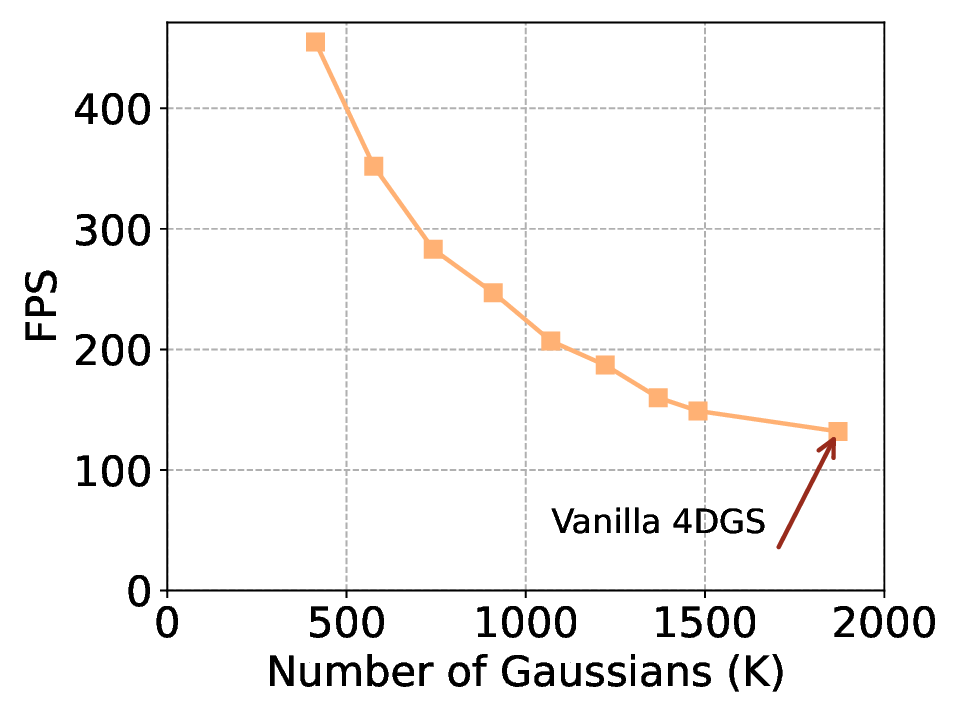}
    \caption{\textbf{Relationship between rendering speed and the number of inactive Gaussians.}} \label{fig:fps}
\end{figure}

We visualize the result in the \textit{Sear Steak} Scene(See \cref{fig:fps}). The FPS decreases as the number of inactive Gaussians increases. This phenomenon strongly supports our redundancy study in~\cref{subsection:Delving}.

\subsection{Visualizations of Pruned Gaussians}
\begin{figure*}[t] \centering
    \begin{subfigure}{0.23\linewidth}
    \includegraphics[width=.9\linewidth]{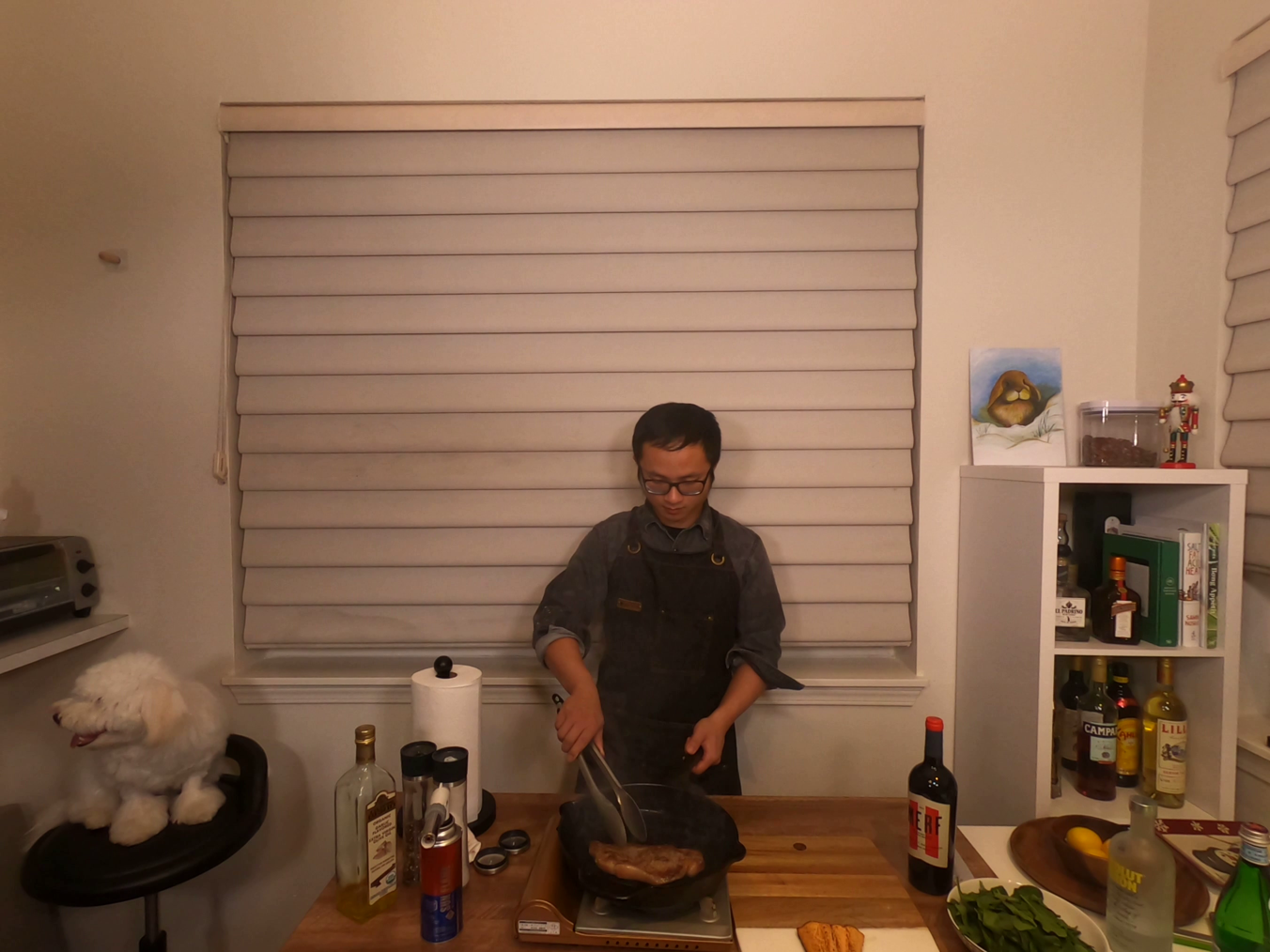}
    \caption{Ground Truth}

  \end{subfigure}
    \begin{subfigure}{0.23\linewidth}
    \includegraphics[width=.9\linewidth]{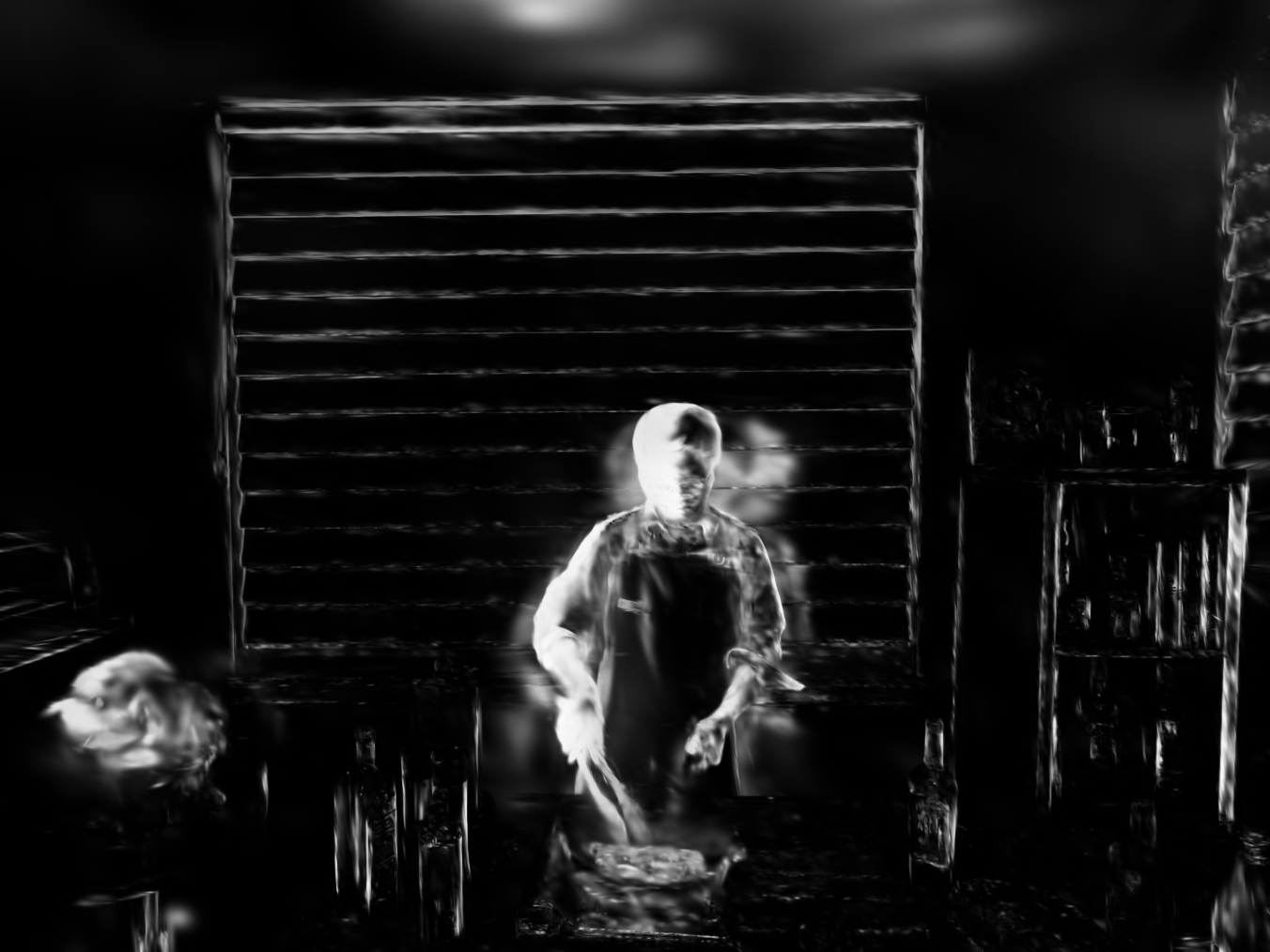}
    \caption{Distribution of $\Sigma_t$}

  \end{subfigure}
  \begin{subfigure}{0.23\linewidth}
    \includegraphics[width=.9\linewidth]{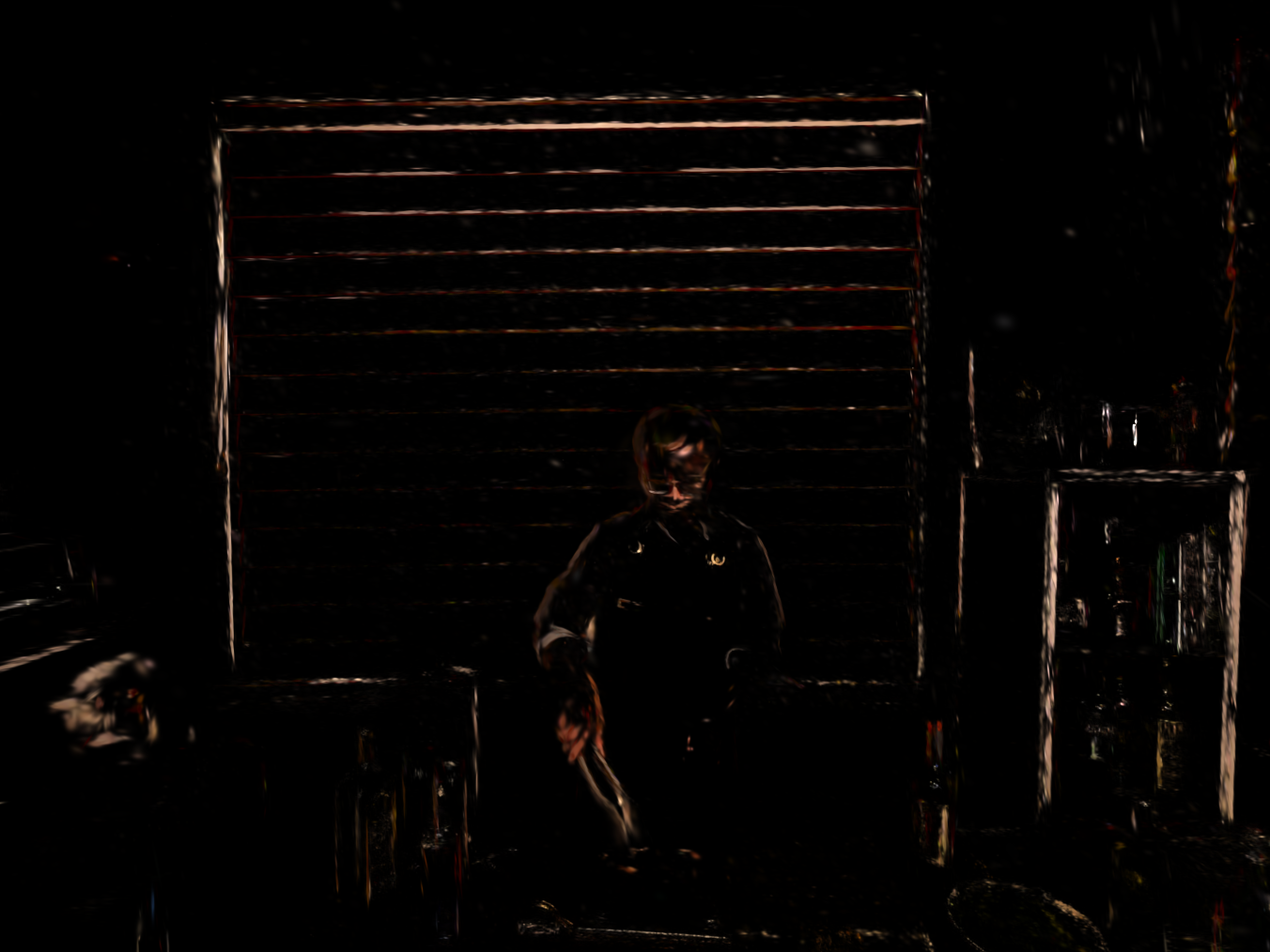}
    \caption{Pruned Gaussians}
    \label{fig:pruned_pruned}

  \end{subfigure}
  \begin{subfigure}{0.23\linewidth}
    \includegraphics[width=.9\linewidth]{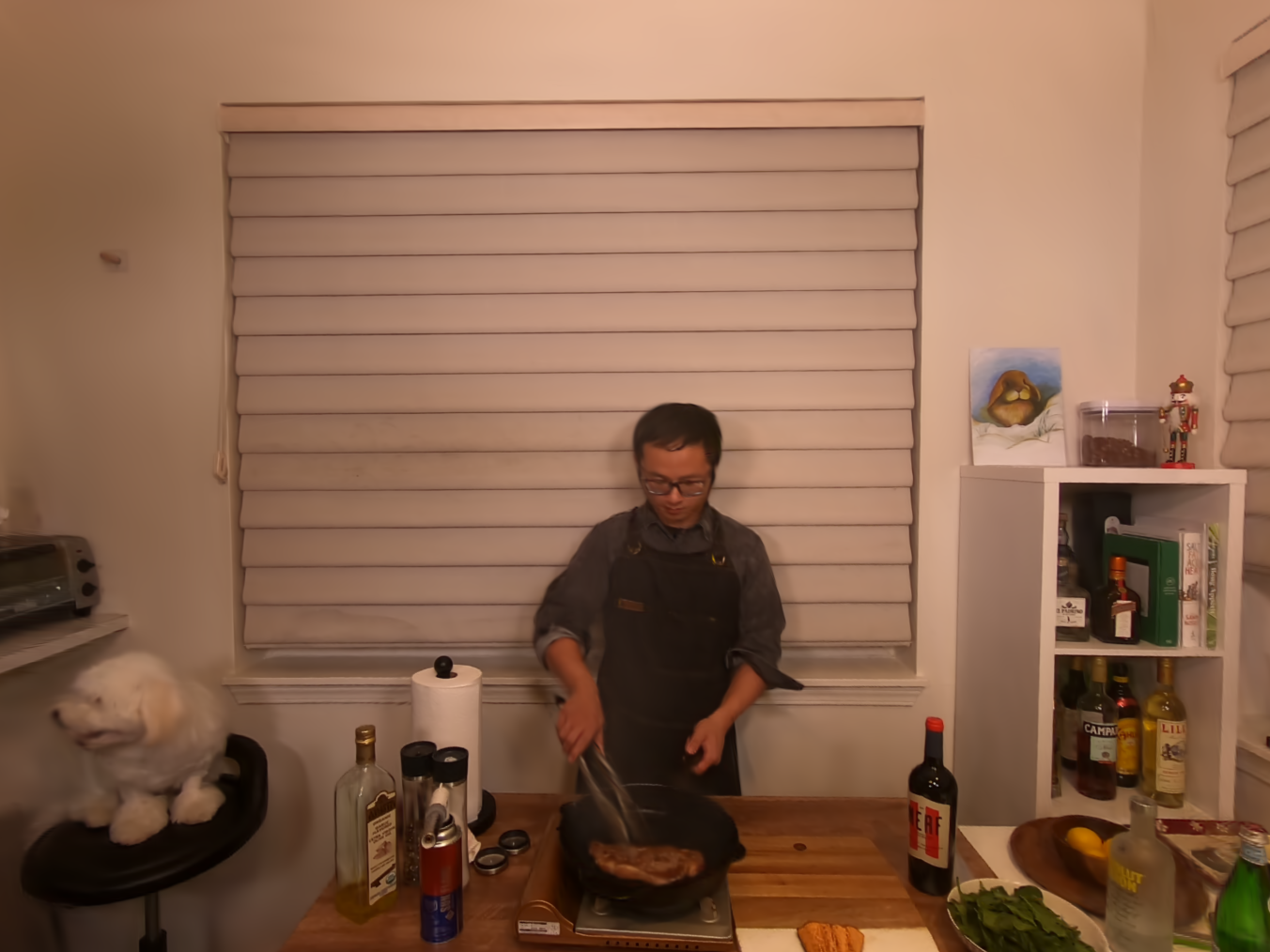}
    \caption{Ours}
    \label{fig:pruned_ours}

  \end{subfigure}
\caption{\textbf{Visualization of Pruned Gaussians.} }
\label{fig:pruned}
\end{figure*}
We provide the visualization of pruned Gaussians in the \textit{Sear Steak} Scene, as shown in~\cref{fig:pruned}. Our pruning strategy can accurately identify Gaussians with short lifespan(See \cref{fig:pruned_pruned}) while maintaining the high quality reconstruction(See \cref{fig:pruned_ours}). The quantized results after pruning are presented in~\cref{tab:ablation}. Our pruning technique achieves $5\times$ compression ratio and $5\times$ faster rasterization speed while slightly improving rendering quality.

\subsection{Video result}
In this work, we propose a novel framework for dynamic 3D reconstruction. Therefore, we provide several videos that are rendered from testing viewpoints on the N3V datasets and D-NeRF datasets to show the reconstruction quality and temporal consistency of 4DGS-1K. These videos are composed by concatenating each frame of 4DGS and our method.

\section{Additional ablation study}
\label{sec:add_abl}
\label{sec:moreablation}
In this section, we firstly provide the variant settings of~\cref{tab:ablation_score}. Furthermore, in addition to the ablation study in the main text, we also investigate the impact of the pruning ratio and different key-frames intervals on rendering quality. We select three distinct scenes, \textit{Cook Spinach}, \textit{Cut Roasted Beef}, and \textit{Sear Steak} on the N3V dataset~\cite{li2022neural} due to the varying performance across different scenes resulting from their unique characteristics. These results show that our default configuration is a well-rounded choice for a wide range of scenes. 

\noindent\textbf{Variant Settings.} As described in~\cref{subsection:pruning}, our Spatial-Temporal Variation Score is composed of two parts, \emph{spatial score} that measures the Gaussians contributions to the pixels in rendering, and \emph{temporal score} considering the lifespan of Gaussians. By aggregating both spatial and temporal score, our score $\mathcal{S}_i$ can be written as:
\begin{equation}
\mathcal{S}_i =\sum^{T}_{t=0}\mathcal{S}_i^T\mathcal{S}_i^S
\end{equation}

Therefore, the variant scores in~\cref{tab:ablation_score} can be written as follow.
\begin{itemize}
    \item (b) $\mathcal{S}^S_{i}$ Only: only considering the spatial part of our score. It can be written as:
    \begin{equation}
        \mathcal{S}_i =\sum^{T}_{t=0}\mathcal{S}_i^S
    \end{equation}
    
    \item (c) $\mathcal{S}^T_{i}$ Only: only considering the temporal contribution part of our score. It can be written as:
    \begin{equation}
        \mathcal{S}_i =\sum^{T}_{t=0}\mathcal{S}_i^T
    \end{equation}
    \item (b) $\mathcal{S}_{i}$ (w. $p^{(1)}_{i}(t)$): Replace the $p^{(2)}_{i}(t)$ with $p^{(1)}_{i}(t)$ in temporal score $\mathcal{S}_i^T$. It can be written as:
    \begin{equation}
    \begin{split}
        \mathcal{S}_i & = \sum^{T}_{t=0}\mathcal{S}_i^T\mathcal{S}_i^S \\
             &= \sum^{T}_{t=0}\mathcal{S}_i^{TV}\gamma(S^{4D}_i)\mathcal{S}_i^S \\
             &= \sum^{T}_{t=0}\frac{1}{0.5\cdot\tanh(
    \left|p^{(1)}_i(t)\right|)+0.5}\gamma(S^{4D}_i)\mathcal{S}_i^S .
    \end{split}
    \end{equation}
    
    \item (c) $\mathcal{S}_{i}$ (w. $\Sigma_t$) Replace the $\mathcal{S}_i^{TV}$ with $\Sigma_t$. It can be written as:
    \begin{equation}
    \begin{split}
        \mathcal{S}_i & = \sum^{T}_{t=0}\mathcal{S}_i^T\mathcal{S}_i^S \\
             &= \sum^{T}_{t=0}\Sigma_t\gamma(S^{4D}_i)\mathcal{S}_i^S \\
    \end{split}
    \end{equation}
\end{itemize}

\noindent\textbf{Performance change with pruning ratio.}  As illustrated in~\cref{fig:ratios}, we analyze the relationship between the pruning ratio and rendering quality. This reveals that our spatial-temporal variation score based pruning can even improve scene rendering quality when the pruning ratio is relatively low in the \textit{Cook Spinach} and \textit{Sear Steak} scenes. Moreover, at higher thresholds, it can maintain results comparable to the vanilla 4DGS~\cite{yang2023real}. Our default setting represents a balanced trade-off between rendering quality and storage size. This setting allows us to achieve a $5\times$ compression ratio while still maintaining high-quality reconstruction.

\begin{figure*}[hp] \centering
    \begin{subfigure}{0.33\linewidth}
    \includegraphics[width=.9\linewidth]{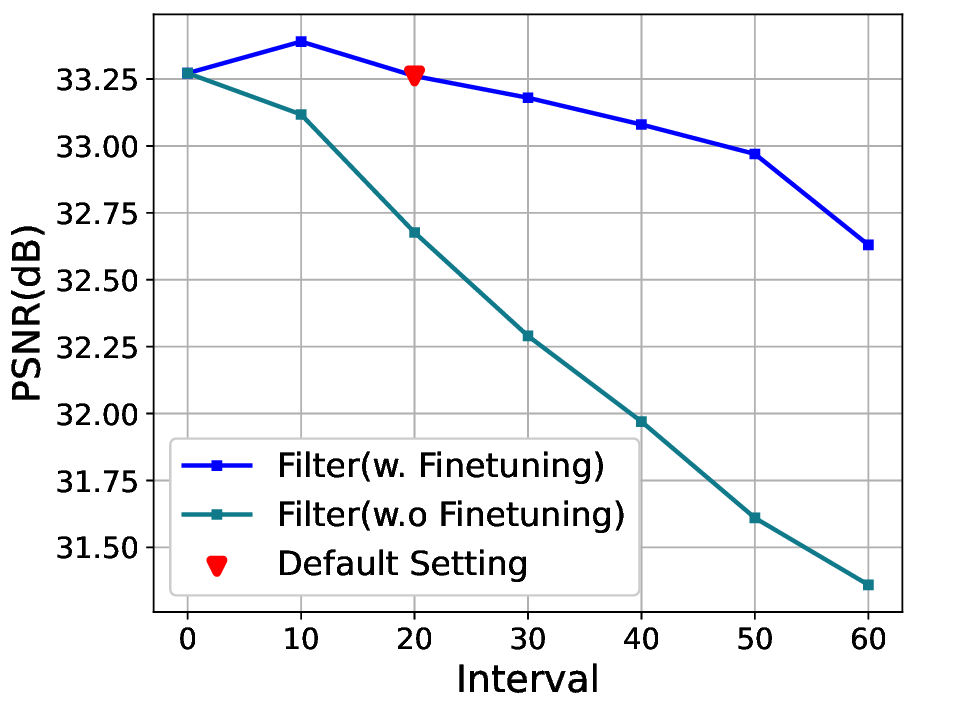}
    \caption{Cook Spinach}

  \end{subfigure}
    \begin{subfigure}{0.33\linewidth}
    \includegraphics[width=.9\linewidth]{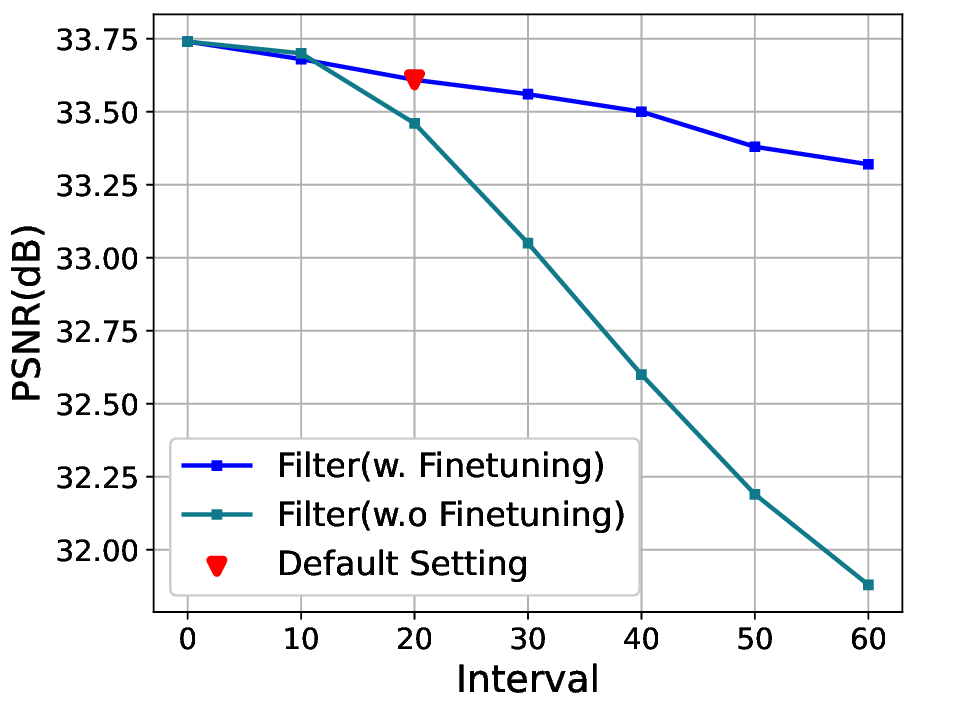}
    \caption{Cut Roasted Beef}

  \end{subfigure}
  \begin{subfigure}{0.33\linewidth}
    \includegraphics[width=.9\linewidth]{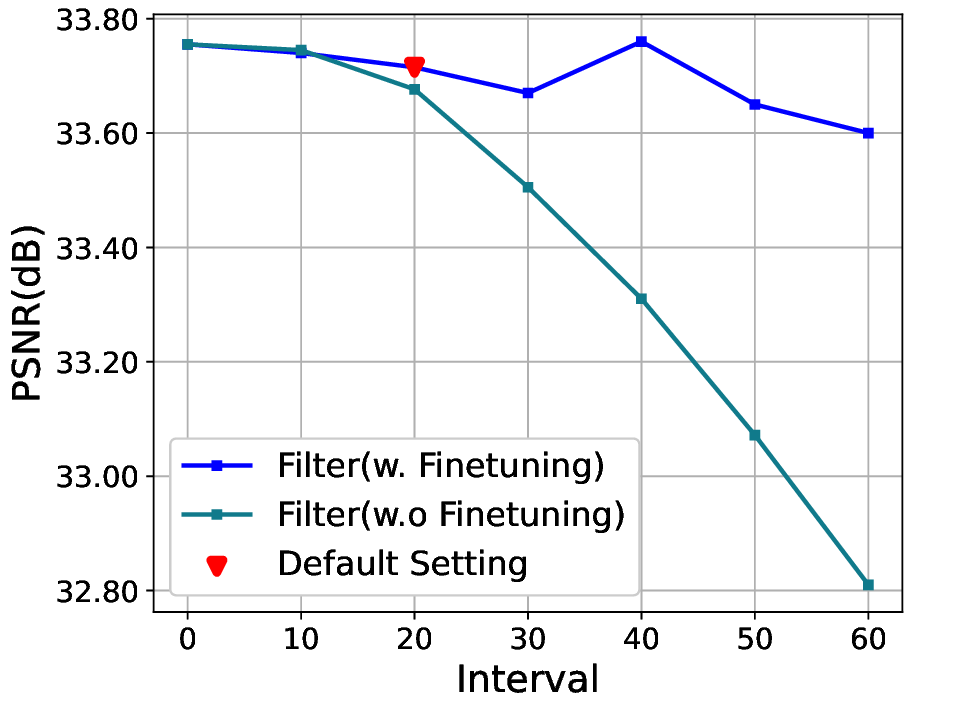}
    \caption{Sear Steak}

  \end{subfigure}
\caption{\textbf{Rate-distortion curves evaluated on diverse scenes with different key-frames interval.}} 
\label{fig:interval}
\end{figure*}

\noindent\textbf{Performance change with key-frames intervals.}
As shown in ~\cref{fig:interval}, although the temporal filter effectively improves rendering speed, its performance degrades significantly when the filter is with long-interval keyframes. However, by integrating the temporal filter into the fine-tuning process, this limitation can be mitigated. The fundamental reason is that some Gaussians which may carry critical scene information are being overlooked due to overly long intervals. However, the fine-tuning process effectively compensates for the loss of this portion of information. This allows us to utilize longer intervals to reduce the additional computational overhead caused by mask calculations.

\section{Discussion}
\label{sec:discussion}
\begin{figure}[h] \centering
    \makebox[0.15\textwidth]{\scriptsize (a) Ground Truth}
    \makebox[0.15\textwidth]{\scriptsize (b) 4DGS}
    \makebox[0.15\textwidth]{\scriptsize (c) Ours}
    \\
    \includegraphics[width=0.15\textwidth]{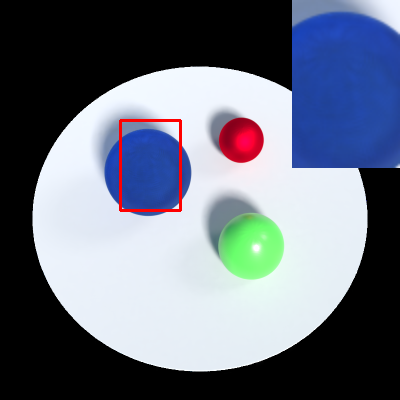}
    \includegraphics[width=0.15\textwidth]{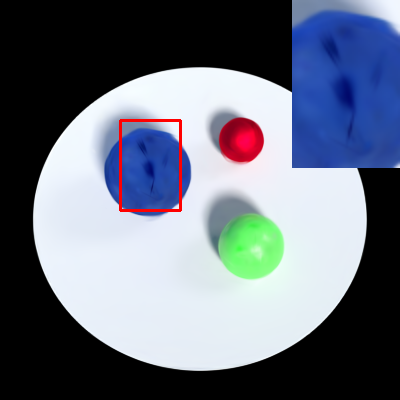}
    \includegraphics[width=0.15\textwidth]{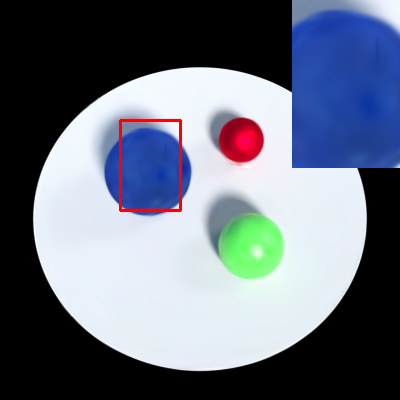}
    \\
    \includegraphics[width=0.15\textwidth]{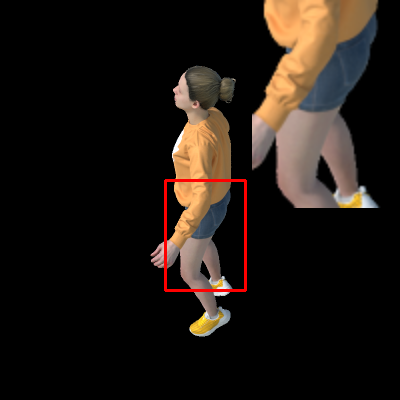}
    \includegraphics[width=0.15\textwidth]{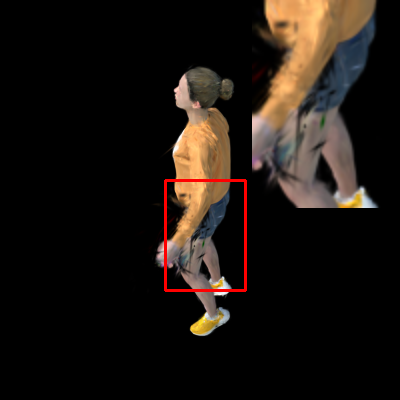}
    \includegraphics[width=0.15\textwidth]{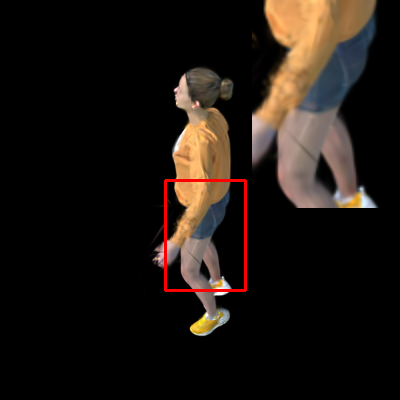}
    
    \caption{\textbf{Visualization of improved performance.}} 
    \label{fig:floaters}
\end{figure}
\noindent\textbf{Improved performance.} 
As shown in \cref{tab:dnerf}, our model achieves a slight PSNR improvement on the D-NeRF Dataset~\cite{pumarola2021d}. This is because vanilla 4DGS often suffers from floaters and artifacts, due to the limited training viewpoints on the D-NeRF Datasets. However, in our study, 4DGS-1K not only can prune the Gaussians with short lifespan, but also reduce the occurrence of floaters and artifacts, as shown in~\cref{fig:floaters}. We visualize two scenes, \textit{Bouncingballs} and \textit{Jumpingjacks}, on the D-NeRF Dataset. These two scenes exhibit floaters and artifacts issues due to limited training viewpoints, as shown in the red box. However, this issue does not appear in 4DGS-1K. Through pruning and filtering, 4DGS-1K successfully mitigates the occurrence of this phenomenon.

\textbf{}

\begin{figure*}[hp] \centering
    \begin{subfigure}{0.33\linewidth}
    \includegraphics[width=.9\linewidth]{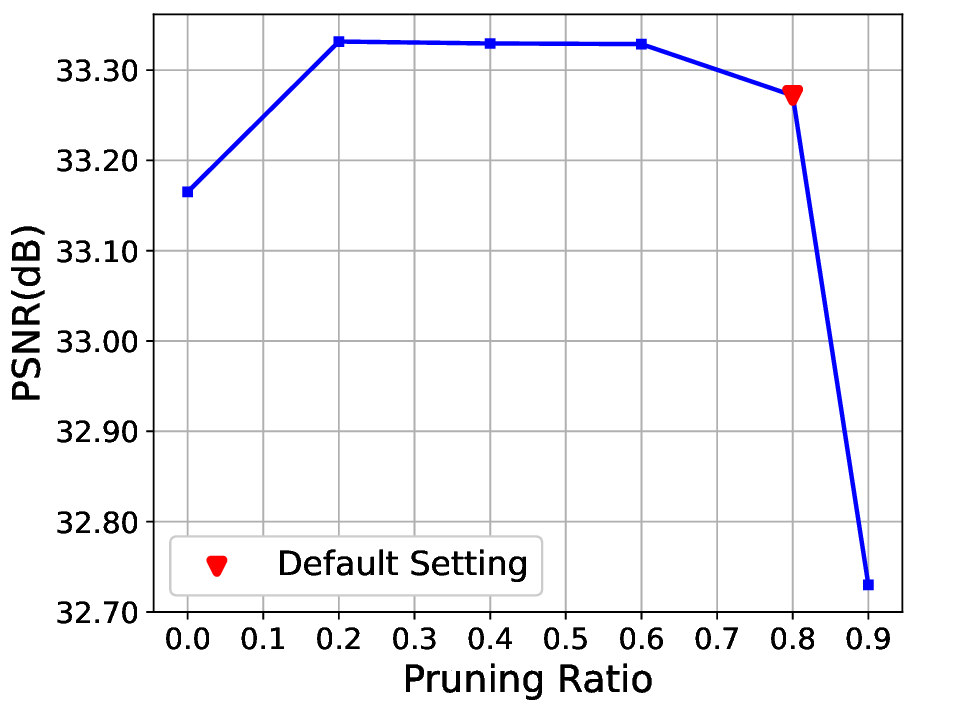}
    \caption{Cook Spinach}

  \end{subfigure}
    \begin{subfigure}{0.33\linewidth}
    \includegraphics[width=.9\linewidth]{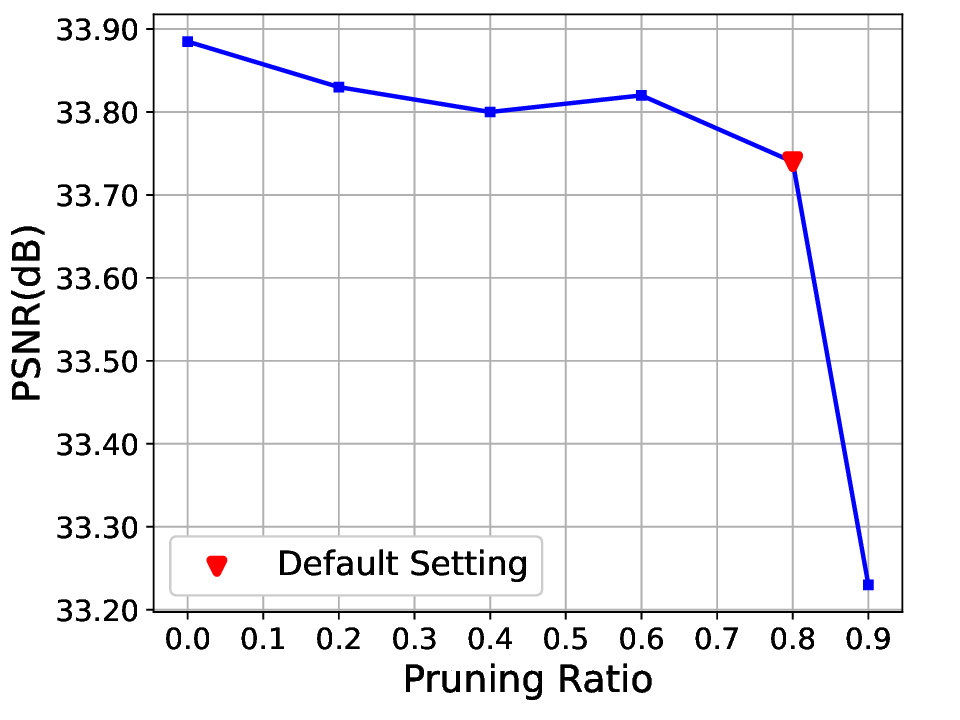}
    \caption{Cut Roasted Beef}

  \end{subfigure}
  \begin{subfigure}{0.33\linewidth}
    \includegraphics[width=.9\linewidth]{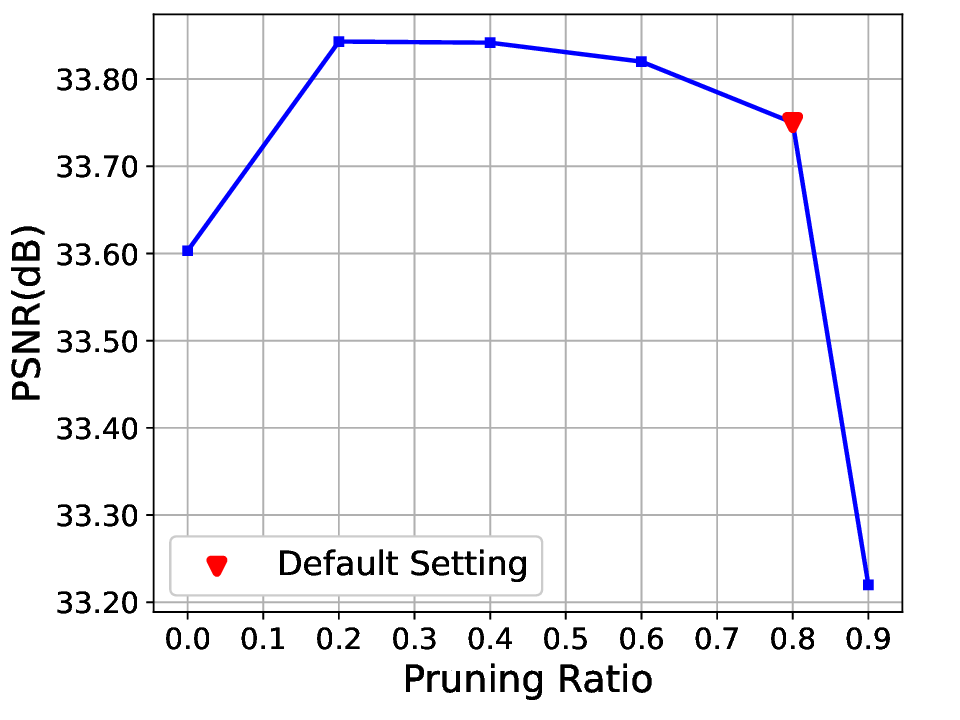}
    \caption{Sear Steak}

  \end{subfigure}
\caption{\textbf{Rate-distortion curves evaluated on diverse scenes with different pruning ratios.} }
\label{fig:ratios}
\end{figure*}

\noindent\textbf{Limitations and Future work.}
As shown in ~\cref{tab:pern3v} and ~\cref{tab:perdnerf}, due to the acceleration provided by the temporal filter, the proportion of time spent on the rasterization process sharply decreases relative to the total rendering time. Therefore, the time consumed by preliminary preparation stages has not gradually become negligible. We hope that future work will focus on optimizing these additional operations within the rendering module to improve its computational performance. 
Moreover, during the pruning process, we specified a predefined pruning ratio. This pruning ratio is influenced by the inherent characteristics of the scene. As shown in~\cref{fig:ratios}, an improper pruning ratio will cause a sharp drop in rendering quality. Therefore, identifying the minimal number of Gaussians required to maintain high-quality rendering across different scenes remains a challenge.
Lastly, there is a significant amount of existing work on Gaussian-based novel view synthesis for dynamic scenes, whereas our model is specifically tailored to a particular model, 4DGS~\cite{yang2023real}. Therefore, developing a universal compression method for these Gaussian-based models is a promising direction for subsequent research endeavors.

\begin{table*}[hp]
\caption{\textbf{Per-scene results of N3V datasets.}}
\centering
\footnotesize
\label{tab:pern3v}
\resizebox{0.9\linewidth}{!}{
\begin{tabular}{ccccccccc}
\hline
\multicolumn{2}{c}{Scene}           & Coffee Martini & Cook Spinach & Cut Roasted Beef & Flame Salmon & Flame Steak & Sear Steak & Average \\ \hline
\multirow{7}{*}{4DGS} & PSNR        &   27.9286       & 33.1651       &  33.8849          &     29.1009      &   33.7970       &   33.6031    &    31.9133   \\
                      & SSIM        &   0.9160        &  0.9545        &   0.9589         & 0.9236        &   0.9615    &   0.9607   & 0.9459  \\
                      & LPIPS       &    0.0759      &  0.0449      & 0.0408        &     0.0691     &      0.0383  &    0.0418  &    0.0518   \\
                      & Storage(MB) &    2764      & 2211       &    1863           &      2969        &         1536  &    1167        &  2085       \\
                      & FPS         &     43       &    89          &   103               &      31        &        122     &      152      &  90       \\
                      & Raster FPS  &    75        &  103        &      122            &       70       &       148      &        195    &   118      \\
                      & \#NUM       &  4441271       &    3530165          &      2979832            &      4719443        &     2457356        &   1870891         &    3333160     \\ \hline
\multirow{7}{*}{Ours} & PSNR        &     28.5780      &       33.2613       &  33.6092        &        28.8488      &    33.2804         &    33.7150        &     31.8821    \\
                      & SSIM        &   0.9185      &   0.9553       &  0.9570            &  0.9221         &  0.9598     &    0.9615  &    0.9457  \\
                      & LPIPS       &    0.0726       &    0.0459       &          0.0435        &      0.0707    &       0.0417      &      0.0401      &     0.0524  \\
                      & Storage(MB) &  557.4        &     443.11     &      374.05         &      592.4    &     308.4     &     234.8    &    418.36  \\
                      & FPS         &   696        &    803       &      853          &   680        &     864     &    935    &   805   \\
                      & Raster FPS  &   901       &     1088      &     1163          &     879      &     1189     &    1332     &  1092    \\
                      & \#NUM       &    888254     &     706033     &        595967       &     943889    &     491471    &    374178     &  666632   \\ \hline
\multirow{7}{*}{Ours-PP} & PSNR     &  28.5472        &    33.0641        &    33.7767         &    28.9878     &    33.2519      &   33.6053     &     31.8722    \\
                      & SSIM        &   0.9166        &    0.9540          &       0.9562           &   0.9209           &       0.9581      &     0.9604       &   0.9444      \\
                      & LPIPS       &   0.0744      &      0.0467        &    0.0445              &        0.0712      &     0.0421        &     0.0402       &     0.0532  \\
                      & Storage(MB) &      64.94    &    52.04          &     44.54             &        69.24      &       36.94      &         29.34   &     49.50    \\
                      & FPS         &   696        &    803       &      853          &   680        &     864     &    935    &   805   \\
                      & Raster FPS  &   901       &     1088      &     1163          &     879      &     1189     &    1332     &  1092    \\
                      & \#NUM       &    888254     &     706033     &        595967       &     943889    &     491471    &    374178     &  666632   \\ \hline
\end{tabular}}
\end{table*}
\begin{table*}[t]
\caption{\textbf{Per-scene results of D-NeRF datasets.}}
\centering
\footnotesize
\label{tab:perdnerf}
\resizebox{0.9\linewidth}{!}{
\begin{tabular}{ccccccccccc}
\hline
\multicolumn{2}{c}{Scene}           & Bouncingballs& Hellwarrior&  Hook  &Jumpingjacks&   Lego   &  Mutant  &Standup&Trex & Average        \\ \hline
\multirow{7}{*}{4DGS} & PSNR        &    33.3472      &    34.7296     &   31.9369     &  30.8247          &     25.3320     &    38.9257      &   39.0411    &  29.8542   &  32.9989              \\
                      & SSIM        &   0.9821      &     0.9516         &    0.9635              &     0.9684         &      0.9178       &    0.9903        &     0.9896&0.9795&0.9678    \\
                      & LPIPS       &    0.0252      &    0.0652          &      0.0385            &       0.0340       &      0.0819       &    0.0090        &      0.0094&0.0193&0.0353   \\
                      & Storage(MB) &       83.69         &     156.53         &   164.91               &    510.99          &     351.19        &     73.24       &   95.38&791.66&278.45      \\
                      & FPS         &    462        &    426          &   414               &     267         &  317           &     463       &    457&202&376     \\
                      & Raster FPS  &    1951       &      1433        &        1309          &   489           & 634            &    1861        &   1878&302&1232      \\
                      & \#NUM       &  133762        &   250201           &     263593             &      816773        &      561357       &    117062        &  152454&1265408&445076       \\ \hline
\multirow{7}{*}{Ours} & PSNR        &    33.4532       &      35.0316        &  32.5118                &    31.8045          &     26.8319        &   37.1916         &   39.3990 &30.4726&33.3370      \\
                      & SSIM        &   0.9826         &   0.9530           &  0.9653                &  0.9716            &    0.9280         &  0.9886          &  0.9896 &0.9811&0.9699       \\
                      & LPIPS       &    0.0248     &    0.0644          &   0.035               &   0.0322           &     0.0674        &   0.0124         &  0.0099 & 0.0180 &0.0330       \\
                      & Storage(MB) &     12.56     &    23.38          &    24.63              &      76.19        &    52.45         &     10.97       &    14.25 &118.24 &41.58     \\
                      & FPS         &     1509    &    1517          &      1444            &    1491          &        1318     &    1518        &   1539 &1361&1462     \\
                      & Raster FPS  &  2600              &     2665         &   2634               &   2476           &     2067        &         2598   &    2644&2174&2482     \\
                      & \#NUM       &      20065          &   37368           &      39360            &  121776            &    83837         &   17527         &   22768 &188986&66460      \\ \hline
\multirow{7}{*}{Ours-PP} & PSNR     &   33.4592       &     35.1570         &     32.5498             &    31.8467          & 27.2850            &     37.0218       &  39.0713&30.6063&33.3746     \\
                      & SSIM        & 0.9821        &  0.9537            &   0.9671               &    0.9728          &  0.9315           &    0.9883        &  0.9896 &0.9821&0.9709       \\
                      & LPIPS       &    0.0259       &    0.0629          &    0.0345              &   0.0309           &   0.0646          &   0.0139         &  0.0109&0.0173&0.0326       \\
                      & Storage(MB) & 4.12       &     5.29         &      5.39            &     11.04         &      8.48       &   3.56         &      3.88&16.11 &7.23  \\
                      & FPS         &     1509    &    1517          &      1444            &    1491          &        1318     &    1518        &   1539 &1361&1462     \\
                      & Raster FPS  &  2600              &     2665         &   2634               &   2476           &     2067        &         2598   &    2644&2174&2482     \\
                      & \#NUM       &      20065          &   37368           &      39360            &  121776            &    83837         &   17527         &   22768 &188986&66460      \\ \hline
\end{tabular}}
\end{table*}

\clearpage 
\newpage


\end{document}